\documentclass[10pt,twocolumn,letterpaper]{article}

\usepackage{wacv}
\usepackage{times}
\usepackage{epsfig}
\usepackage{graphicx}
\usepackage{amsmath}
\usepackage{amssymb}
\usepackage{subfig}
\usepackage{tabularx}

\def\wacvPaperID{3} 

\wacvfinalcopy 
\ifwacvfinal
\def\assignedStartPage{9876} 
\fi

\ifwacvfinal
\usepackage[breaklinks=true,bookmarks=false]{hyperref}
\else
\usepackage[pagebackref=true,breaklinks=true,colorlinks,bookmarks=false]{hyperref}
\fi

\ifwacvfinal
\else
\fi

\begin{document}

\title{Deep Learning based Novel View Synthesis}

\author{Amit More \hspace{2cm} Subhasis Chaudhuri \\
Indian Institute of Technology Bombay\\
{\tt\small amitmore@ee.iitb.ac.in}
}


\maketitle
\ifwacvfinal\thispagestyle{empty}\fi

\begin{abstract}
Predicting novel views of a scene from real-world images has always been a challenging task. 
In this work, we propose a deep convolutional neural network (CNN) which learns to predict novel views of a scene from given collection of images. 
In comparison to prior deep learning based approaches, which can handle only a fixed number of input images to predict novel view, proposed approach works with different numbers of input images. 
The proposed model explicitly performs feature extraction and matching from a given pair of input images and estimates, at each pixel, the probability distribution (pdf) over possible depth levels in the scene. 
This pdf is then used for estimating the novel view. 
The model estimates multiple predictions of novel view, one estimate per input image pair, from given image collection. 
The model also estimates an occlusion mask and combines multiple novel view estimates in to a single optimal prediction. 
The finite number of depth levels used in the analysis may cause occasional blurriness in the estimated view. 
We mitigate this issue with simple multi-resolution analysis which improves the quality of the estimates. 
We substantiate the performance on different datasets and show competitive performance. 
\end{abstract}


\begin{figure}[h]
\centering
{\includegraphics[scale=.4,trim={0 0 0 0},clip]{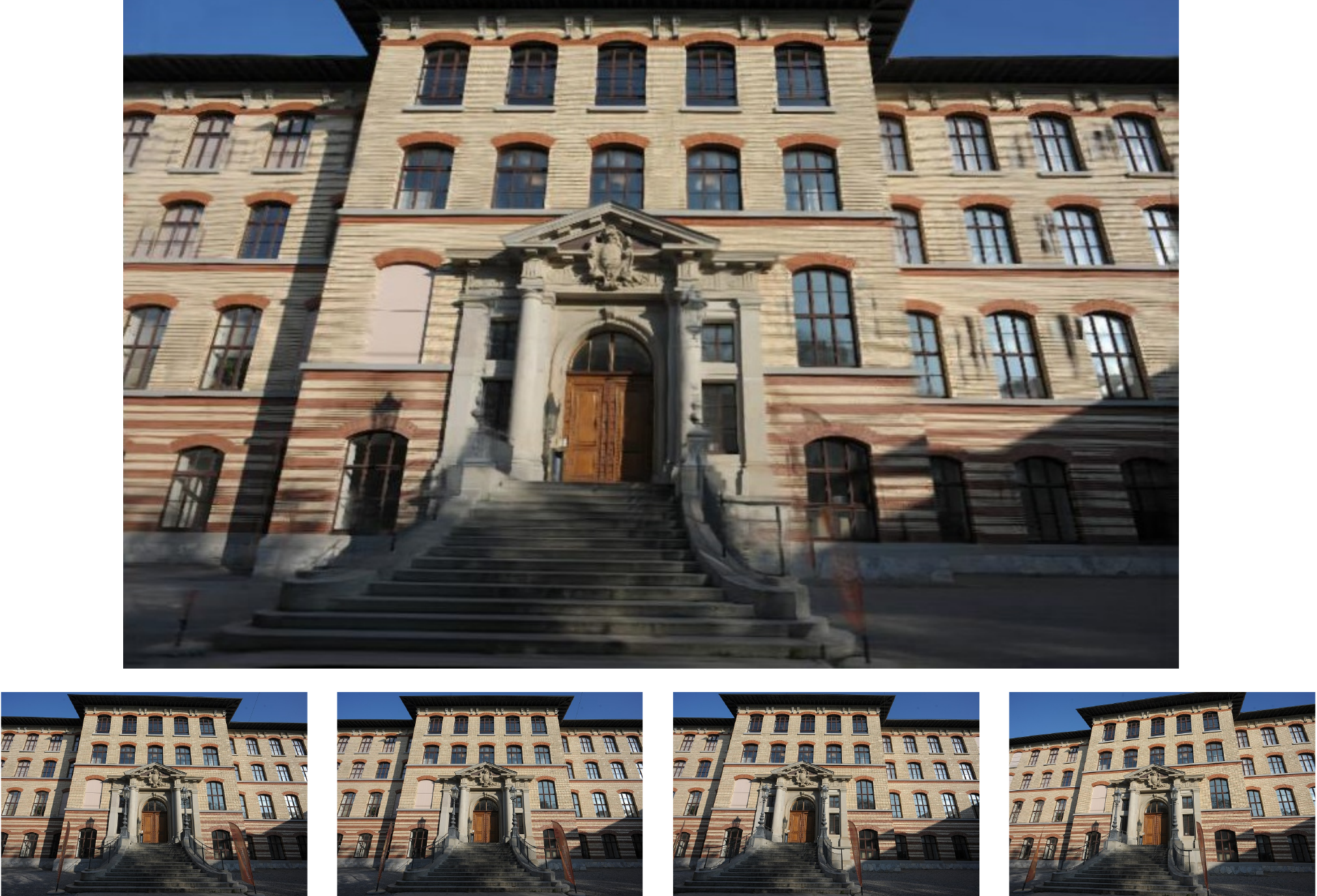}}
\caption{ We show novel view of a scene (top) generated from other images (bottom) using proposed approach. }
\label{fig:ViewSynth}
\end{figure}

\section{Introduction}

Structure from motion (sfm) and synthesis of a novel view have been very popular and challenging problems in computer vision. 
There are several methods in the literature attempting to solve these problems such as~\cite{snavely2008modeling,strecha2006combined,woodford2007new,chaurasia2013depth,cremers2011multiview}. 
Typically, these approaches consist of several stages of a carefully designed pipeline. 
In general, these methods perform well only under limited settings such as indoor scenes and man made objects, like buildings and roads, where one can impose strong priors.
Such methods fail to perform well in more general situations, for example, in case of outdoor scenes. 
This is mainly due to the ill posed nature of the problem under such situations and failure of the priors or assumptions imposed by the method itself. 
Learning based methods show great potential under such challenging conditions. 
Such methods are relatively easy to design, as it needs a limited set of assumptions on the data and generalizes well given a rich training dateset. 

In this work, we present a Deep Learning (DL) based approach which aims at solving the novel view synthesis problem. 
We aim to synthesize new view of a given scene when images of a scene from different view points are available as shown in Fig.~\ref{fig:ViewSynth}. 
However, view synthesis is inherently an ill posed problem due to several challenges involved. 
For example, the object present in the scene may appear totally different in different images due to changes in view point and occlusions.

Our work closely relates to the work of Flynn \textit{et al.} \cite{flynn2016deepstereo} and Habtegebrial \textit{et al.} \cite{habtegebrial2018fast} on novel view synthesis. 
Flynn was the first to use DL for view synthesis for images captured from unstructured camera positions with challenging scene structure such as vegetation. 
They use $4$ input views of a scene to predict target image from a novel camera view point. 
This limits the applicability of their method to more general settings where there are either more or less than $4$ input images available. 
\cite{habtegebrial2018fast} uses a similar approach and achieves better performance. 
However, their method relies on a collection of stereo pairs which limits their applicability, as it is difficult to obtain stereo pairs in general. 
In contrast, the proposed method can handle any number of input images, unlike \cite{flynn2016deepstereo}. 
Further, the approach works with images captured from arbitrary camera positions, unlike stereo pairs needed in \cite{habtegebrial2018fast}. 
In particular, we divide the collection of input images into several pairs. 
We sample epipolar lines from inputs at different depth levels to create plane sweep volumes (PSV). 
We use CNN for processing these PSVs and estimate probabilities (pdf) over depth levels in the scene at each pixel. 
This way our model can handle different numbers of input images, one pair at a time, giving multiple estimates of the pdf. 
These estimates are used to warp corresponding input images to target camera view point giving multiple estimates of the novel view. 
In general, these estimates may have different distortions due to occlusions present in input images. 
Further, errors in the estimated pdfs can cause certain artifacts and blurring effects. 
Proposed model predicts an occlusion mask to combine the estimated views such that only accurately predicted regions are retained in the final output and distortions are minimized. 

Using finite number of depth levels may cause inaccuracies in the estimates of depth probabilities which can cause blurriness in the predictions \cite{srinivasan2019pushing}. 
We mitigate this issue by adaptive depth sampling to increase the sampling frequency. 
In particular, we propose to resample epipolar lines with the help estimated pdfs, i.e. to discretize depth levels in a smaller range where the estimated pdf have high probability. 
We recompute PSVs this way which results in more accurate feature correspondence and improves the quality of the predictions.  

\section{Related Work}

View synthesis have been used as a proxy for unsupervised depth estimation in \cite{Zhou_2017_CVPR, Godard_2017_CVPR, zhan2018unsupervised,yin2018geonet,mahjourian2018unsupervised}. 
The estimated depth map is used for warping input images to synthesize a view from different camera position and view prediction error is used as a supervision. 

Given a collection of images, one can estimate global scene geometry using structure-from-motion techniques \cite{hedman2017casual,debevec1996modeling,goesele2010ambient,hedman2018deep,hedman2016scalable,arikan2014large}. 
The estimated geometry can be used to render input images into novel camera viewpoint. 
However, it is generally difficult to get a single consistent global model of entire scene geometry and often the estimated point cloud is sparse in nature which can result in artifacts when generating the novel view. 

View synthesis using per view local geometry is presented in \cite{chaurasia2013depth,nie2017homography,cayon2015bayesian,penner2017soft}. 
The planar scene approximation at super pixel level is presented in \cite{chaurasia2013depth,nie2017homography}. 
Different image based rendering algorithms are analyzed in Bayesian approach in \cite{cayon2015bayesian}. 
Accuracy of the estimated depth map for each input image is improved in \cite{penner2017soft} by analyzing multiple depth maps for occlusions. 

Recently view synthesis is posed as a supervised learning problem.  
Scene information is represented in the form of Multi Plane Images (MPIs) in \cite{zhou2018stereo,srinivasan2019pushing,mildenhall2019local} which results in fast rendering of novel views for view interpolation and extrapolation. 
Similarly, view extrapolation to much larger baselines have been shown in \cite{choi2019extreme}. 
In \cite{flynn2016deepstereo} feature correspondence is established along epipolar lines for $4$ input images using convolutional network. 
The model estimates probability distribution for depths levels in scene which is used for view synthesis. 
In \cite{habtegebrial2018fast} $4$ stereo pairs are used for depth estimation which is then used for view synthesis. 
An iterative approach to view synthesis is presented in \cite{flynn2019deepview} where an optimization problem to estimate an MPI is posed as a learning problem and solved using learned gradient descent approach \cite{adler2017solving}. 
View synthesis for light field cameras is presented in \cite{kalantari2016learning,srinivasan2017learning}. 
In \cite{kalantari2016learning} novel view is estimated from four input images. 
The authors model possible disparities and generates multiple view estimates, which are used by deep network to estimate final refined disparity map for view prediction. 
In \cite{srinivasan2017learning} depth map from a single image is predicted and used to synthesize light field images. 

View synthesis can also be modeled simply as a pixel flow without need to estimate an explicit scene geometry. 
Video frame interpolation has been posed as a learning problem in \cite{jiang2018super,niklaus2018context,liu2017video,niklaus2017video,Niklaus_2017_CVPR,de2016dynamic,mahajan2009moving,zitnick2004high,fitzgibbon2005image} to model the pixel motion between input images for view interpolation.  


\begin{figure*}
\centering
\subfloat[]{\includegraphics[scale=0.225,trim={0cm 0cm 0cm 0cm},clip]{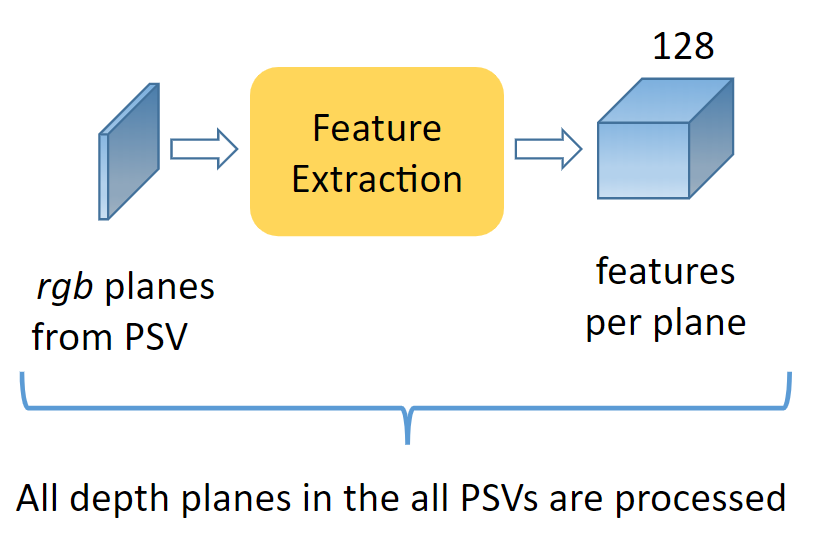}{\label{subfig:fe}}}~
\subfloat[]{\includegraphics[scale=0.225,trim={0cm 0cm 0cm 0cm},clip]{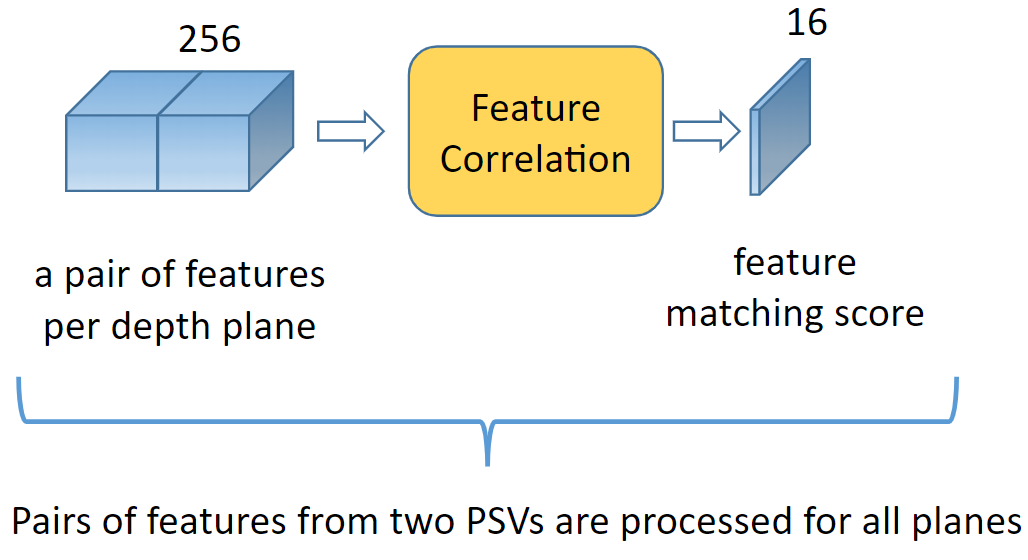}{\label{subfig:fc}}}~
\subfloat[]{\includegraphics[scale=0.225,trim={0cm 0cm 0cm 0cm},clip]{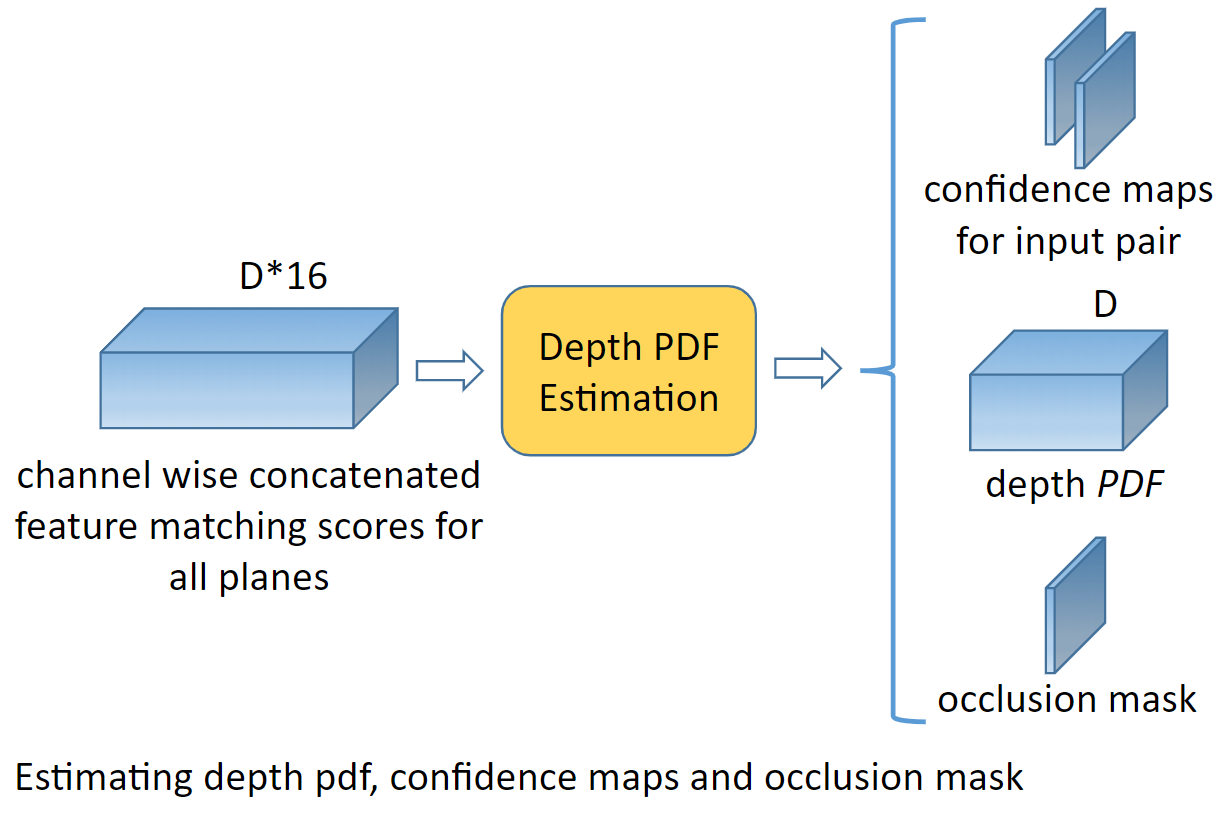}{\label{subfig:de}}}~
\subfloat[]{\includegraphics[scale=0.225,trim={0cm 0cm 0cm 0cm},clip]{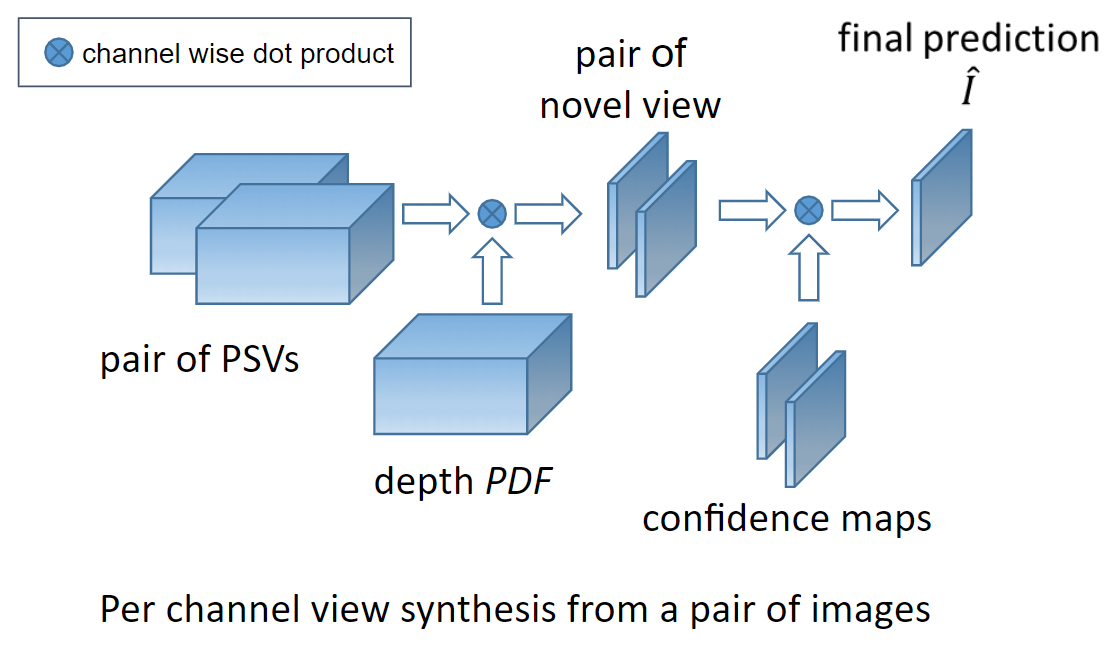}{\label{subfig:ve}}}


\caption{The overall model for novel view synthesis is shown in the figure. 
(a): Each depth plane from PSVs are processed independently and $128$ dimensional feature vector is extracted. 
(b): Features from two PSVs for each depth plane are processed by feature correlation module. The module generates $16$ dimensional feature matching score per depth plane. 
(c): Feature matching scores for all depth planes are compared by depth estimation module and the pdf is estimated. The model also estimates the confidence maps $C$ and $C^{'}$ and occlusion mask $O$. 
(d): The novel view is generated using estimated pdf and confidence maps from given pair of input PSVs as shown.}
\label{fig:cnn}
\end{figure*}

\section{Proposed Method}
Given a collection of images, our model predicts a novel view from the perspective of virtual camera. 
We assume that the intrinsic and extrinsic parameters of all cameras are known. 
Given these parameters, we construct PSVs from input images similar to \cite{flynn2016deepstereo}. 
The PSVs contains all epipolar lines extracted from input views. 
Thus, estimating the structure of a scene can be done efficiently by performing feature matching across input views along epipolar lines in PSVs. 
We show our proposed model in Fig.~\ref{fig:cnn}. 
The model has four submodules as shown below. 

\subsection{Feature extraction} 
We process two PSVs corresponding to given image pair for feature extraction as shown in Fig.~\ref{subfig:fe}. 
The input to this module is a single \textit{rgb} image plane extracted from given PSVs. 
We compute features for each plane in PSVs separately. 
Model takes $3$ channel image and extracts $128$ dimensional feature vector for each plane in both PSVs. 

\subsection{Feature matching} 
The inputs to this module are features computed by previous module. 
Features from specific depth plane from both PSVs are concatenated and processed together as shown in Fig.~\ref{subfig:fc}. 
The module may then learn to compare these features for a potential match. 
We process all planes in PSVs this way. 
The model output is a $16$ dimensional feature matching score. 

\subsection{Depth PDF estimation} 
This module processes the feature matching scores corresponding to all depth planes together as shown in Fig.~\ref{subfig:de}. 
Each plane in PSVs corresponds to specific depth level in a scene. 
By processing all planes together, this module learns to compare feature matching scores across all depth levels and generates the probability distribution. 
The model also predicts a pair of confidence maps corresponding to input image pair and an occlusion mask. 
We explain the utility of these additional estimates below. 

\subsection{Novel view estimation} 
\begin{figure}
\center

{\includegraphics[scale=.085,trim={550 370 0 0},clip]{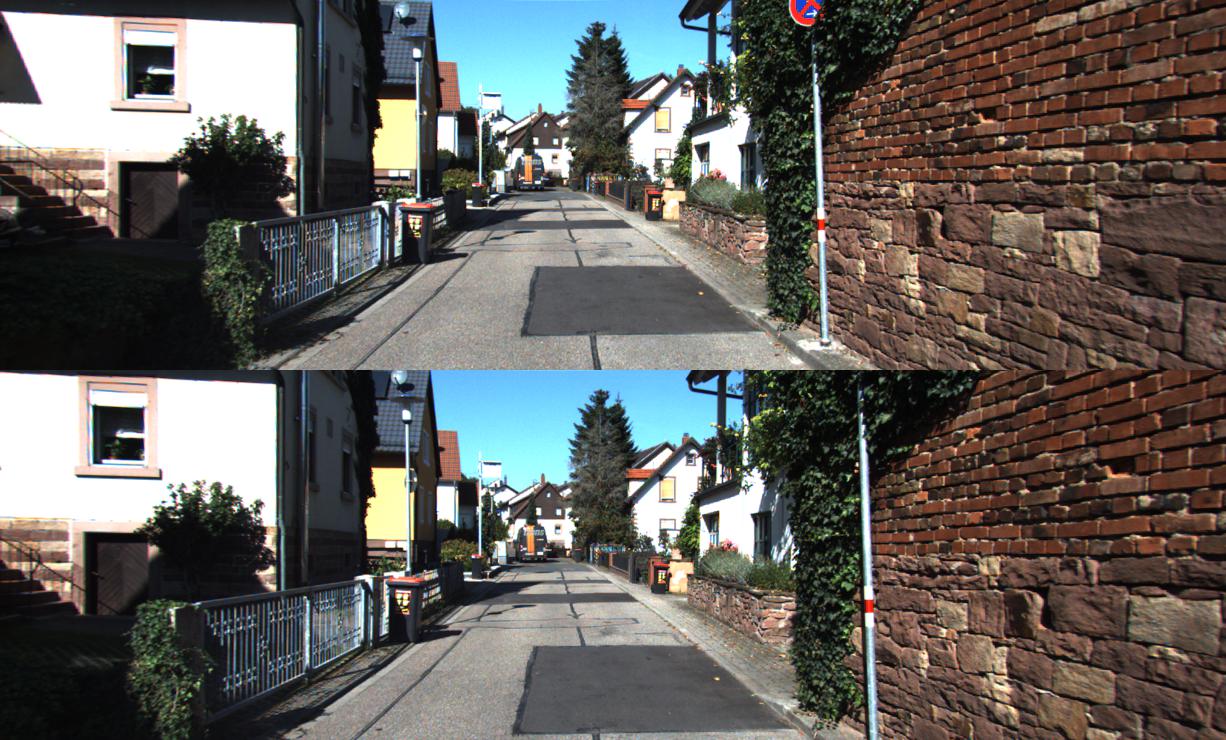}}~
{\includegraphics[scale=.085,trim={550 370 0 0},clip]{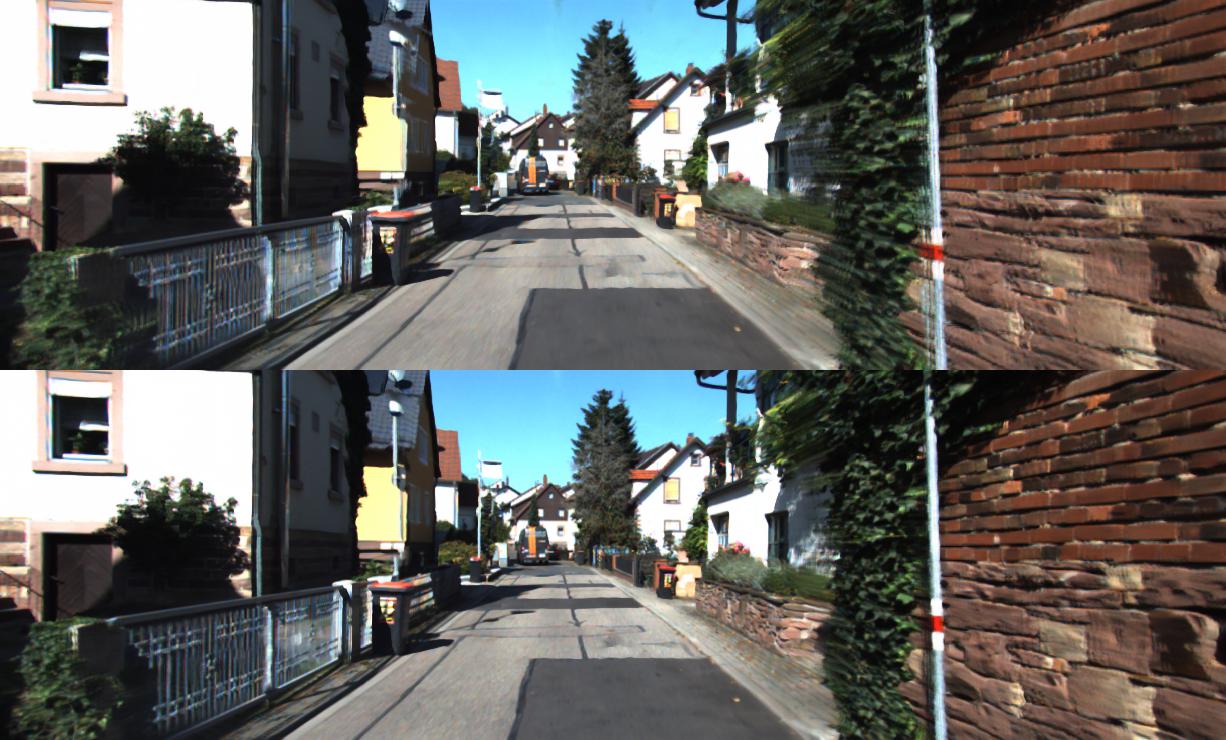}}~
{\includegraphics[scale=.085,trim={550 370 0 0},clip]{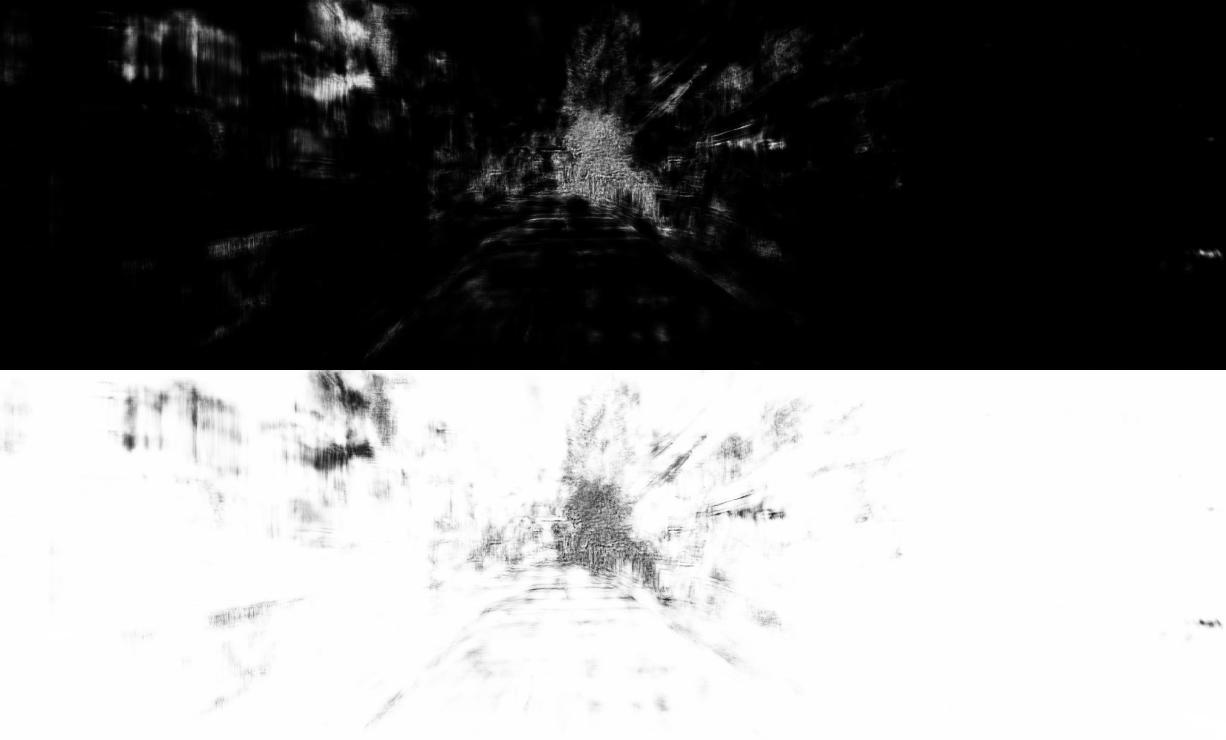}}~
{\includegraphics[scale=.085,trim={550 370 0 0},clip]{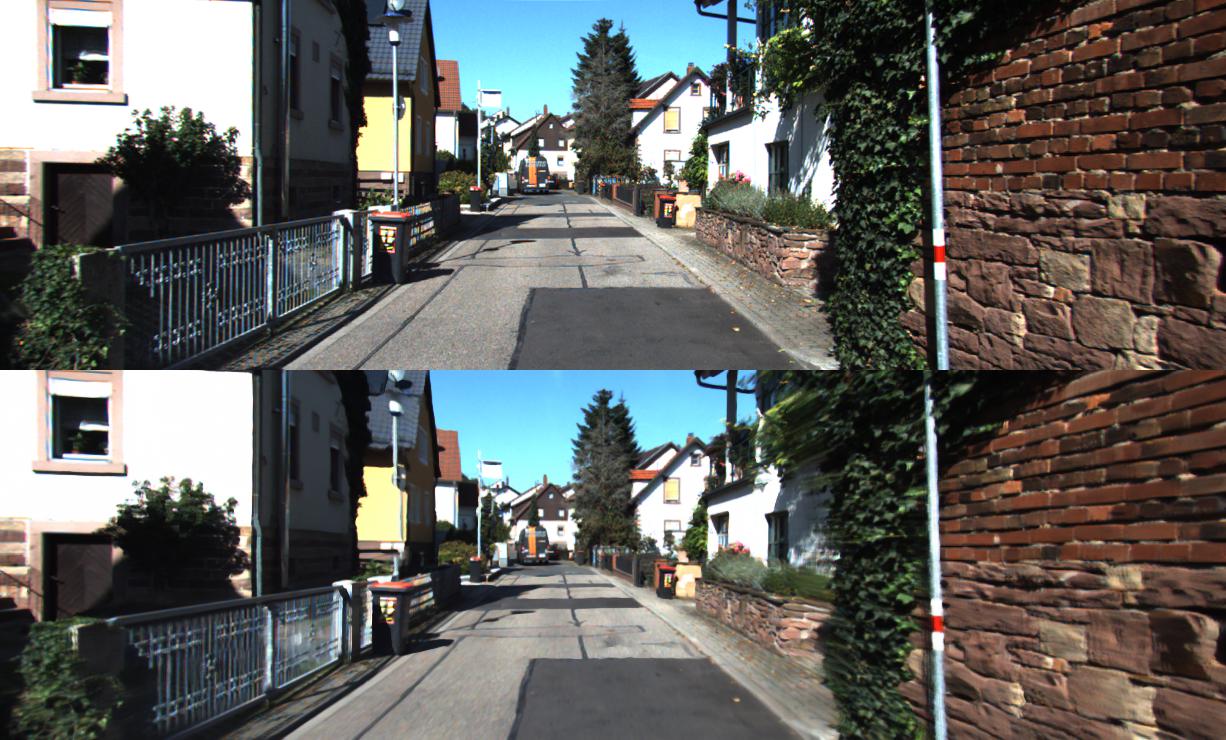}}\\

\subfloat[]{\includegraphics[scale=.085,trim={550 0 0 370},clip]{./Results/8/Testing_926_01_Inputs.jpg}\label{subfig:iv}}~
\subfloat[]{\includegraphics[scale=.085,trim={550 0 0 370},clip]{./Results/8/Testing_926_01_Preds.jpg}\label{subfig:nvs}}~
\subfloat[]{\includegraphics[scale=.085,trim={550 0 0 372},clip]{./Results/8/Testing_926_01_ViewConf.jpg}\label{subfig:conf_maps}}~
\subfloat[]{\includegraphics[scale=.085,trim={550 0 0 370},clip]{./Results/8/Testing_926_01_Targets.jpg}\label{subfig:nv}}

\caption{ View synthesis from a given image pair is shown. (a): Input image pair $I$ and $I^{'}$. (b): Novel views $\bar I$ and $\bar I{'}$ synthesized using Eq.~\ref{eq:pdf}. (c): The estimated confidence maps $C$ and $C^{'}$. (d): The ground truth view (top) and final estimate of the novel view (bottom) estimated using Eq.~\ref{eq:conf}. The model predicted high confidence on the second image (bottom) for pixels in the right side region and for pixels on the left side region model predicted equal confidence. }
\label{fig:confidence_maps}
\end{figure}

This module is a computational block without any parameters as shown in Fig.~\ref{subfig:ve}. 
Let the given input image pair be denoted by $I$ and $I^{'}$ and corresponding PSVs by $V$ and $V^{'}$. 
Let $P$ denote the estimated pdf, then estimated target view is given by 
\vspace{-.5cm}
\begin{align}
\bar I(x,y) &= \sum_{d=1}^{D} V(x,y,d) \  P(x,y,d), \nonumber \\
\bar I^{'}(x,y) &= \sum_{d=1}^{D} V^{'}(x,y,d) \  P(x,y,d).
\label{eq:pdf}
\end{align}

where $x,y$ spans the spatial dimension, $d$ is specific depth plane and $D$ denotes the total number of depth levels used for sampling epipolar lines. 
Since we get two estimates, one per input image, we further combine these estimates as shown below,

\begin{equation}
\hat I(x,y) = \bar I(x,y)*C(x,y) + \bar I^{'}(x,y)*C^{'}(x,y).
\label{eq:conf}
\end{equation}

where $C$ and $C^{'}$ denote pixel-wise confidence maps (softmax probabilities). 
The model may predict one of the estimates from $\bar I$ and $\bar I^{'}$ to be better than the other depending on the scene visibility in input images. 
For example, in Fig.~\ref{fig:confidence_maps} we show estimated novel views $\bar I$ and $\bar I{'}$ (Fig.~\ref{subfig:nvs}) and corresponding confidence maps $C$ and $C^{'}$ (Fig.~\ref{subfig:conf_maps}). 
One may note that for pixels in the right hand side region of images model predicts high confidence on $\bar I{'}$ (Fig.~\ref{subfig:nvs} bottom) as corresponding scene is better visualized in image $I{'}$ (Fig.~\ref{subfig:iv} bottom) as can be seen by observing input images (Fig.~\ref{subfig:iv}) and the target view (Fig.~\ref{subfig:nv} top).

\subsection{Combining predictions from multiple pairs}
So far we have considered only a pair of input images and reconstructed the novel view. 
The information of a scene provided by a single image pair may not always be complete. 
For example, there might be regions in target view which are not covered by given image pair or a particular region might be occluded in input images. 
We mitigate these effects by utilizing multiple image pairs to generate multiple predictions of the target view. 
These predictions may, in general, be supplementary to each other. 
Thus we also learn to combine these predictions using an occlusion mask predicted by the model. 

Let $\hat I_i$ denote the estimated novel view using $i^{th}$ image pair and corresponding occlusion mask be denoted by $O_i$. 
In order to combine multiple estimates, we normalize the occlusion masks across different predictions using softmax function. 
The final predicted novel view is given by
\begin{equation}
I_{final}(x,y) = \sum\limits_{\forall i} \hat I_i(x,y)*O_i(x,y)
\label{eq:ocl}
\end{equation}
where $I_{final}$ denote the final novel view estimate. 
We compute the prediction error (loss function) as $L_1$ norm between the ground truth $I_{gt}$ and the final estimate as
\begin{equation}
Loss = \sum\limits_{x,y} |I_{gt}(x,y) - I_{final}(x,y)|
\label{eq:loss}
\end{equation}

Our idea of self asserting the quality of the estimates using occlusion masks is similar to \cite{zhou2016view,sun2018multi}. 
The model does not need explicit supervision for occlusion masks but learns to estimate it using view synthesis criteria.

\begin{figure}
\center
\subfloat[]{\includegraphics[scale=.125,trim={613 0 0 370}, clip]{./Results/8/Testing_926_01_Inputs.jpg}{\label{subfig:0_in}}}~
\subfloat[]{\includegraphics[scale=.125,trim={613 370 0 0}, clip]{./Results/8/Testing_926_01_Inputs.jpg}{\label{subfig:1_in}}}~
\subfloat[]{\includegraphics[scale=.125,trim={0 0 0 0}, clip]{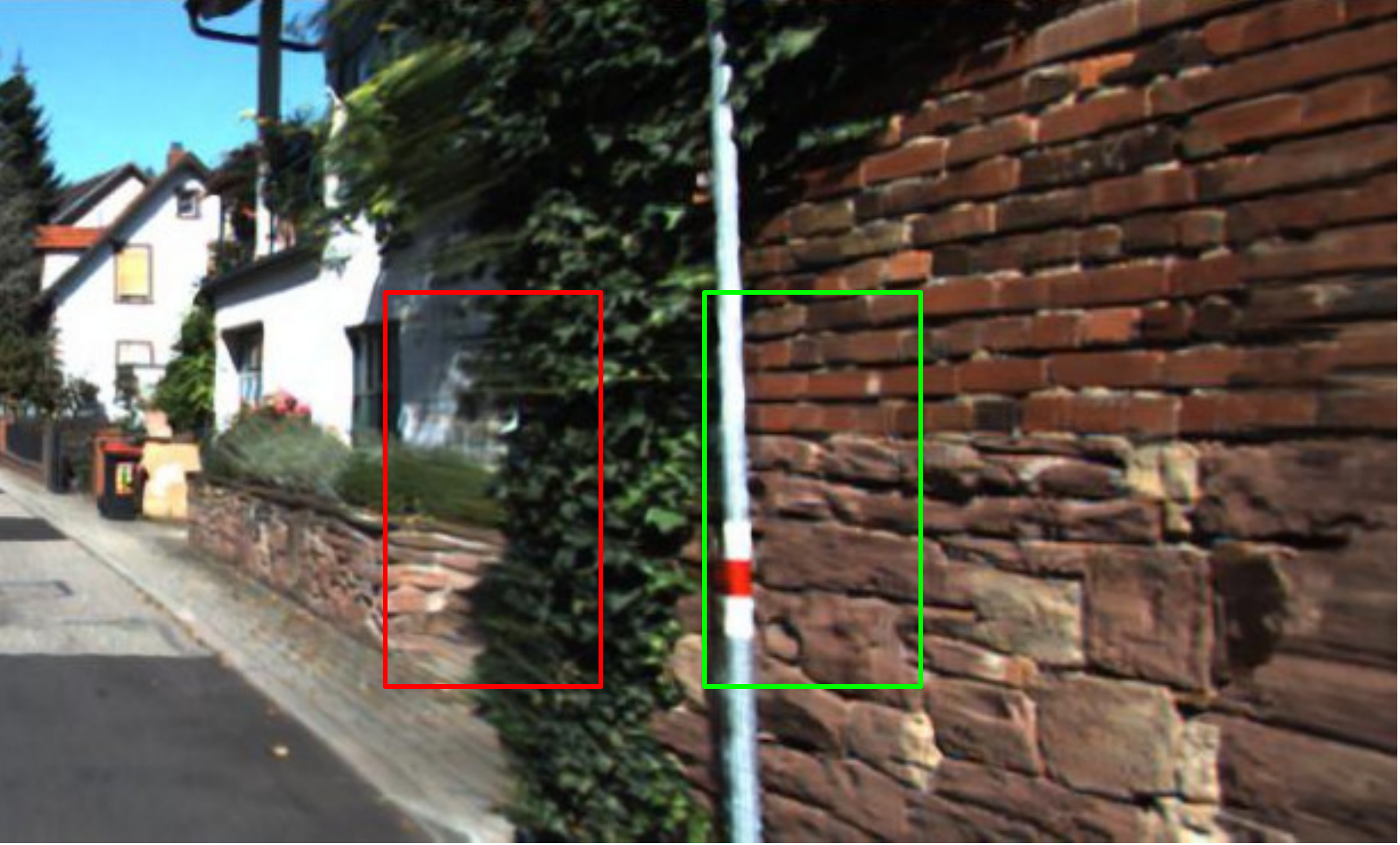}{\label{subfig:01_t}}}\\
\vspace{-.5\baselineskip}
\subfloat[]{\includegraphics[scale=.125,trim={613 0 0 370}, clip]{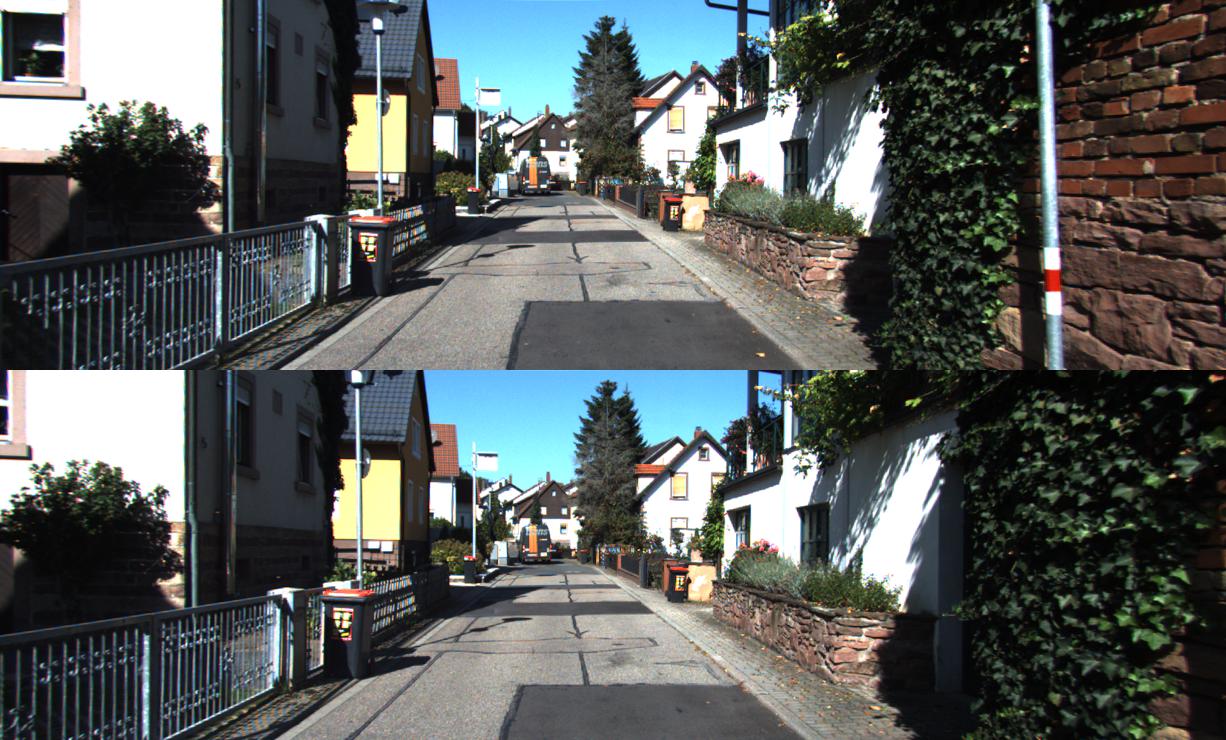}{\label{subfig:2_in}}}~
\subfloat[]{\includegraphics[scale=.125,trim={613 370 0 0}, clip]{./Results/8/Testing_926_23_Inputs.jpg}{\label{subfig:3_in}}}~
\subfloat[]{\includegraphics[scale=.125,trim={0 0 0 0}, clip]{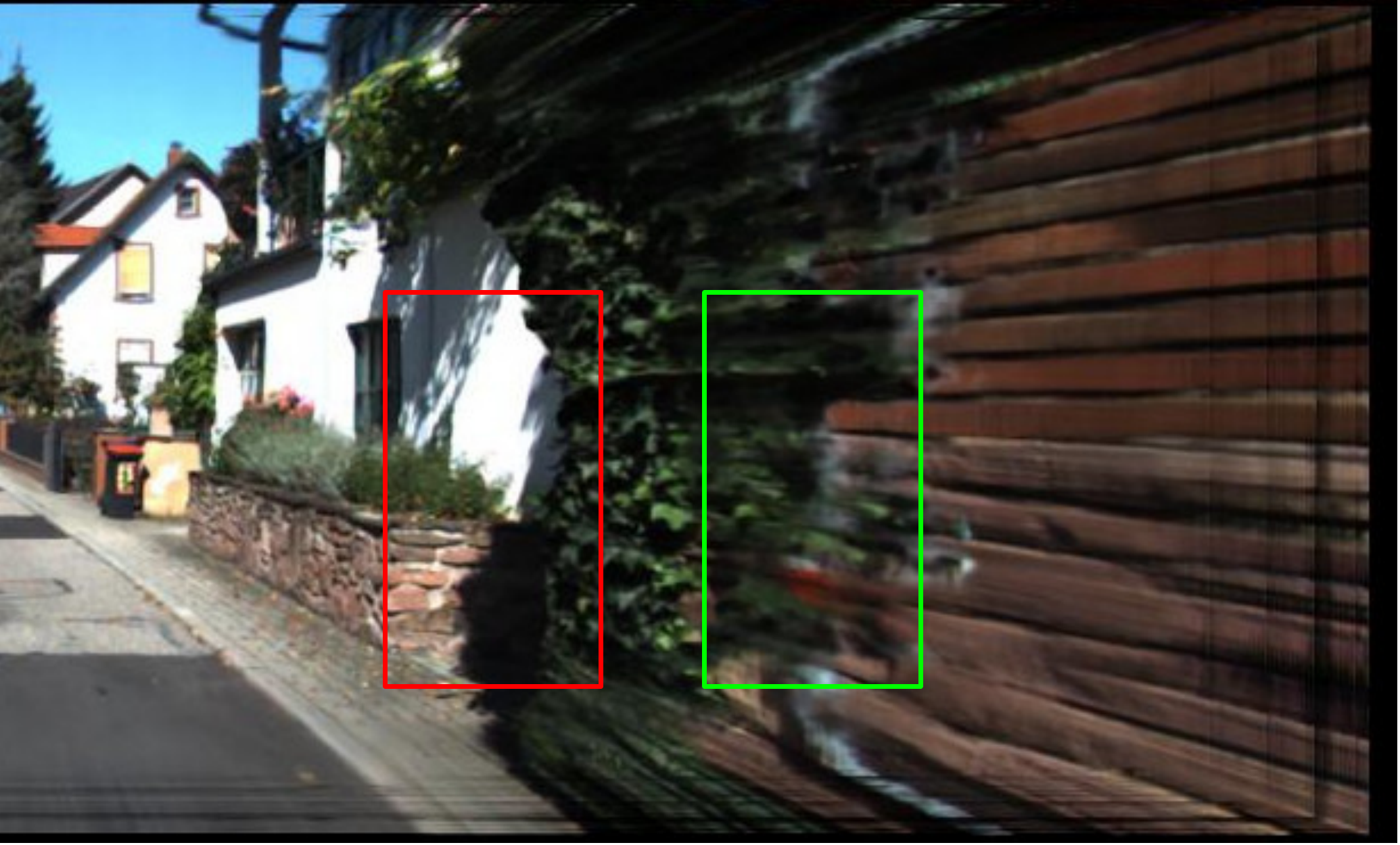}{\label{subfig:23_t}}}\\
\vspace{-.5\baselineskip}
\subfloat[]{\includegraphics[scale=.125,trim={613 370 1226 0}, clip]{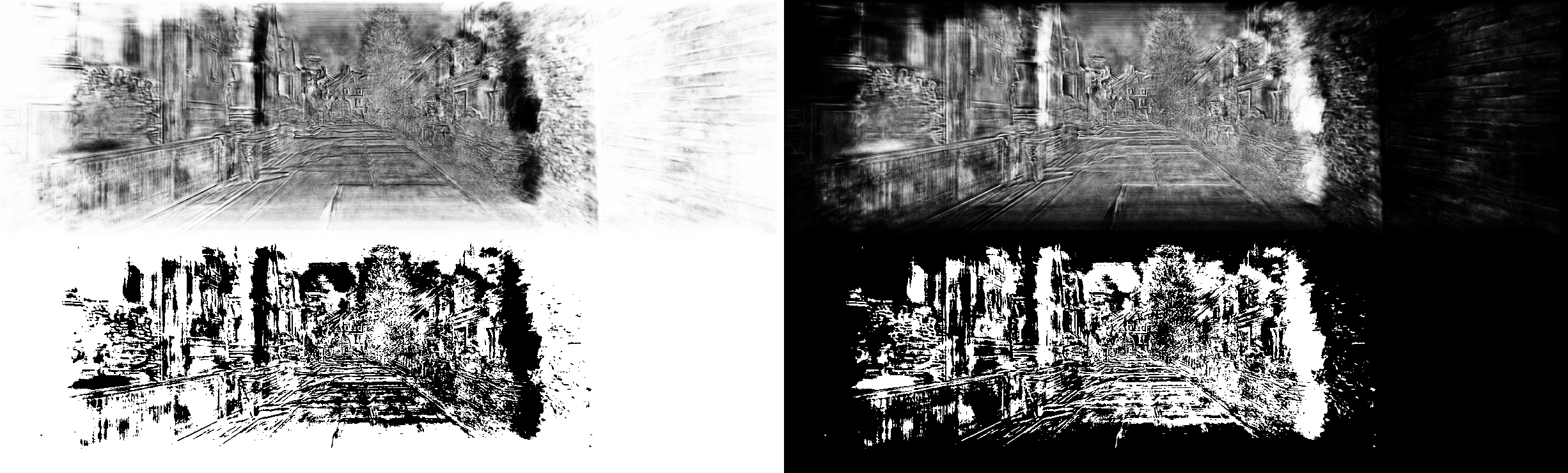}{\label{subfig:oclMask_1}}}~
\subfloat[]{\includegraphics[scale=.125,trim={1839 370 0 0}, clip]{./Results/8/01_23/conf_all_926_all.jpg}{\label{subfig:oclMask_2}}}~
\subfloat[]{\includegraphics[scale=.125,trim={1839 370 0 370}, clip]{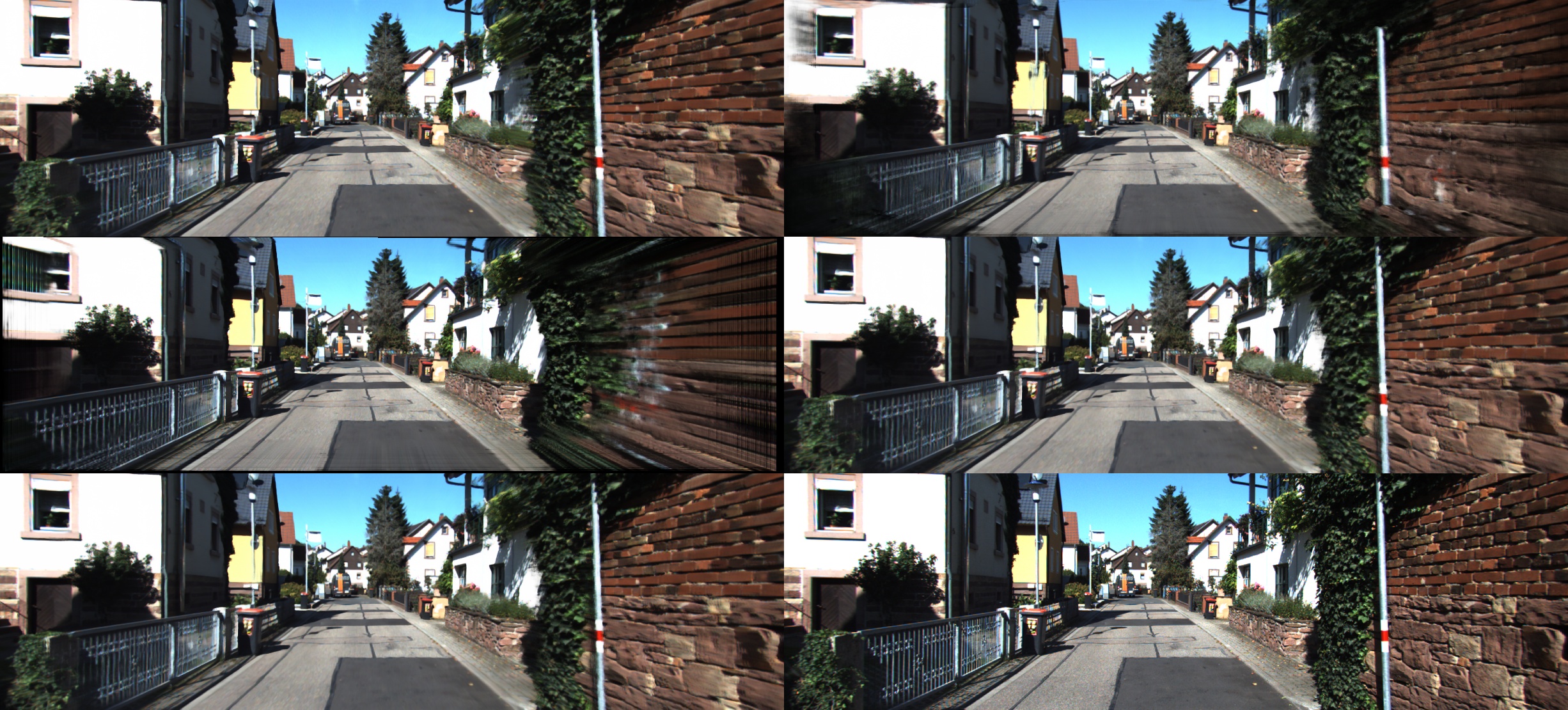}{\label{subfig:final_t}}}~

\caption{The figure above shows an example of occlusion and partial scene visibility. (a,b): An input image pair. (c): Novel view synthesized using images in (a,b). Predictions inside red box are poor because of occlusions. (d,e): Another input image pair. (f): Novel view estimated using image pair (d,e). Region within green box is poorly estimated because of partial scene visibility. (g,h): Occlusion masks for estimated novel views in (c) and (f). (i): The final view estimated from (c) and (f) using occlusion masks (g,h) is free from artifacts.}
\label{fig:occl_eg}
\end{figure}

Fig.~\ref{fig:occl_eg} shows a typical example of an occlusion and partial scene visibility. 
Fig.~\ref{subfig:01_t} and Fig.~\ref{subfig:23_t} show two estimates of the same scene using different image pairs. 
Due to occlusions present in image pair (Fig.~\ref{subfig:0_in},~\ref{subfig:1_in}) the region shown in the red box in Fig.~\ref{subfig:01_t} is poorly synthesized, however, the same region is synthesized properly in Fig.~\ref{subfig:23_t} due to better visibility (Fig.~\ref{subfig:2_in},~\ref{subfig:3_in}). 
On the other hand, image pairs in Fig.~\ref{subfig:2_in} and Fig.~\ref{subfig:3_in} have partial scene visibility and hence can not generate most of the scene as observed in the green box in Fig.~\ref{subfig:23_t}. 
We show the estimated occlusion masks for these predictions in Fig.~\ref{subfig:oclMask_1} and Fig.~\ref{subfig:oclMask_2}. 
The final novel view computed using occlusion masks, shown in Fig.~\ref{subfig:final_t}, is free from the artifacts. 
Thus we note that the proposed model learns to combine multiple estimates to improve the overall prediction quality. 

\subsection{Multi-resolution analysis (mr)}
Due to finite quantization of depth levels, the estimated pdf may have poor accuracy. 
The effect is more prominent when camera baseline is high \cite{srinivasan2019pushing}. 
In order to mitigate this issue, we resample depth levels in a much shorter range. 
We take a small patch from the pdf and estimate the average depth range for which the probabilities are higher than the predefined threshold. 
We resample depth levels in this small range giving dense depth sampling. 
Due to higher sampling frequency, model is able to establish more accurate correspondence between PSVs. 
In Fig.~\ref{fig:mr} (left) we show the novel view estimated using original depth range showing blurring artifacts. 
In the reestimated novel view after depth resampling, shown in Fig.~\ref{fig:mr} (middle), these artifacts have been reduced significantly. 


\begin{figure}
\center
{\includegraphics[scale=.8,clip]{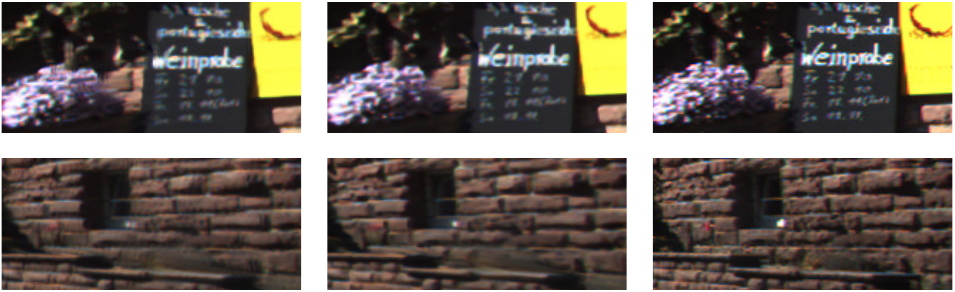}}
\caption{Left: Novel view synthesized using original depth levels have blurriness. Middle: After adaptive depth resampling in a shorter range estimate of the novel view improves. Right: Ground truth view.}
\label{fig:mr}
\end{figure}

\begin{figure*}
\centering
\raisebox{0.0in}{\rotatebox[origin=b]{90}{\ \  \textbf{0.8m} \ \ \ \ \ \ \ \ \ \ \ \ \  \textbf{0.8m} \ \ \ \ \ \ \ \ \ \ \ \  \textbf{1.6m} \ \ \ \ \ \ \ \ \ \ \ \  \textbf{0.8m}}}   
\begin{minipage}{0.18\textwidth}
\centering
\textbf{\scriptsize Ours, 2 i/p views}\par\medskip
{\includegraphics[scale=0.15,trim={0 0 600 0},clip]{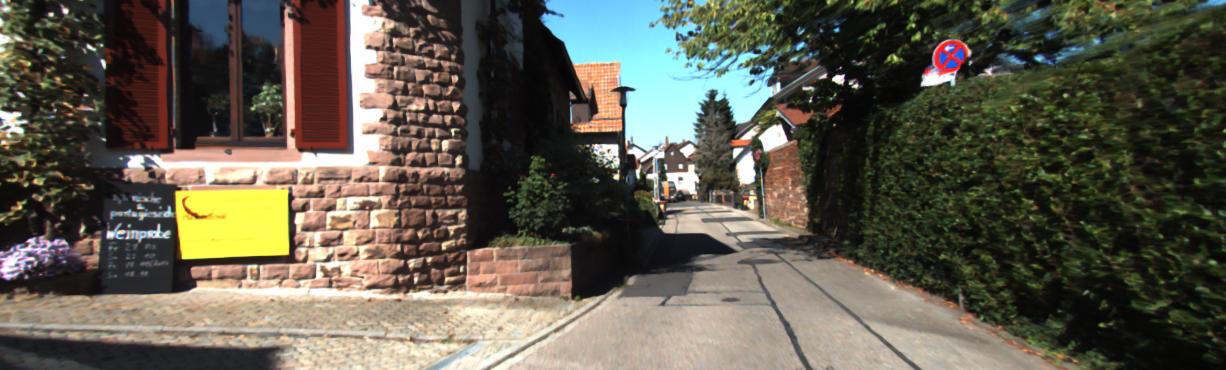}}
{\includegraphics[scale=0.15,trim={0 0 600 0},clip]{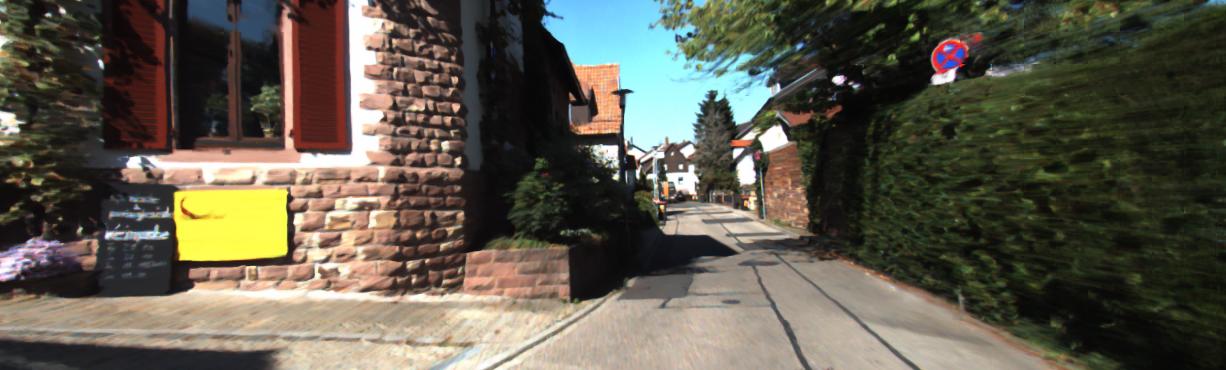}}
{\includegraphics[scale=0.15,trim={600 0 0 0},clip]{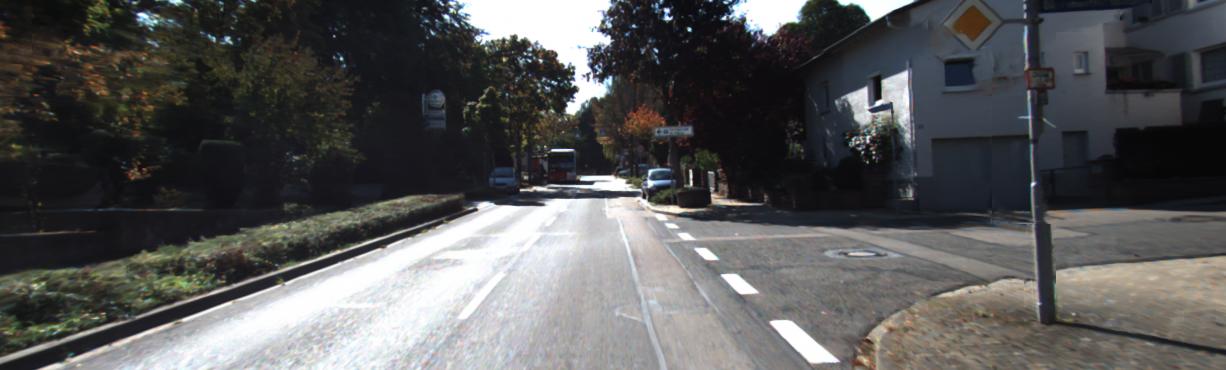}}
{\includegraphics[scale=0.15,trim={0 0 600 0},clip]{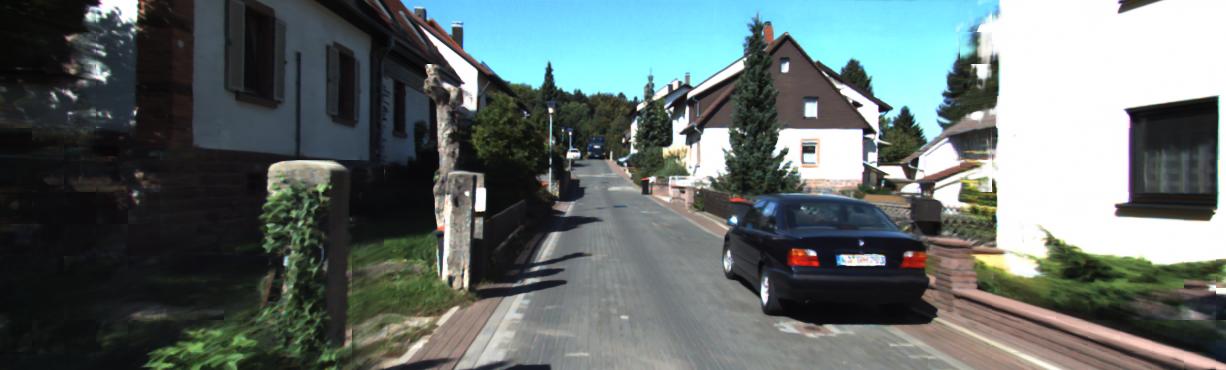}}
\end{minipage}\hfill
\begin{minipage}{0.18\textwidth}
\centering
\textbf{\scriptsize Ours, 3 i/p views}\par\medskip
{\includegraphics[scale=0.15,trim={0 0 600 0},clip]{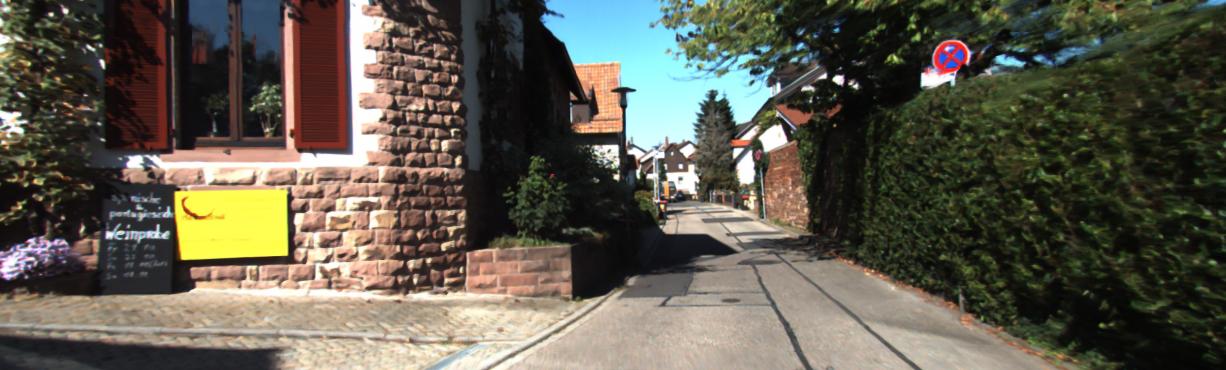}}
{\includegraphics[scale=0.15,trim={0 0 600 0},clip]{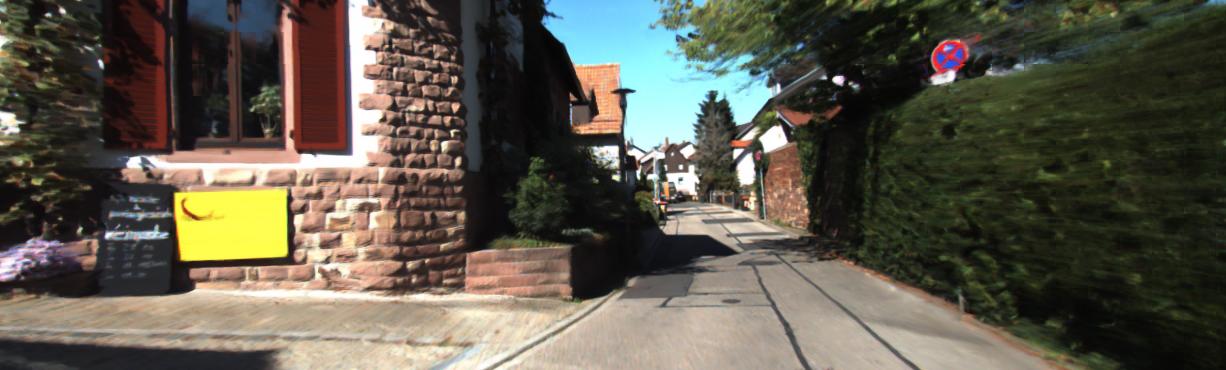}}
{\includegraphics[scale=0.15,trim={600 0 0 0},clip]{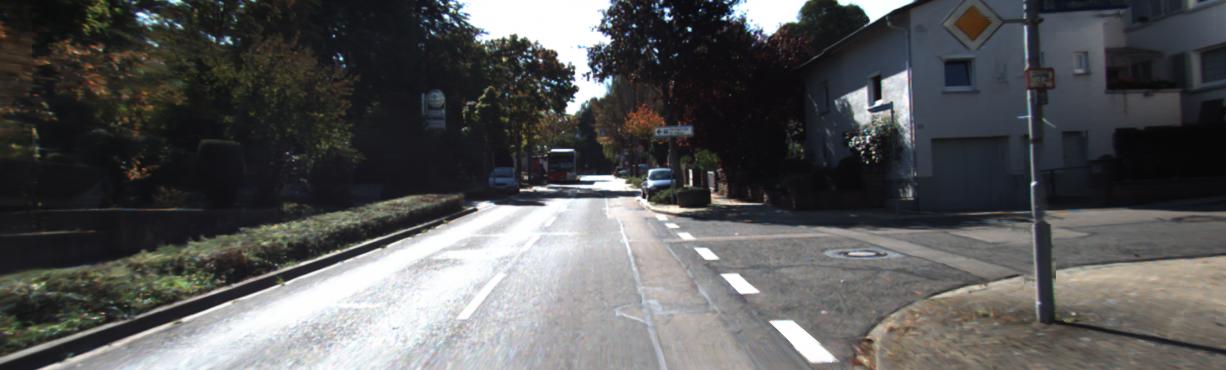}}
{\includegraphics[scale=0.15,trim={0 0 600 0},clip]{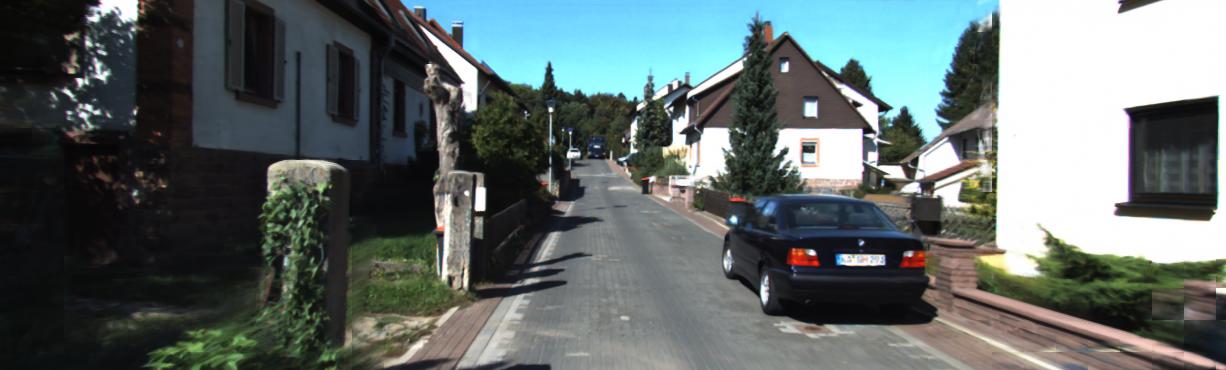}}
\end{minipage}\hfill
\begin{minipage}{0.18\textwidth}
\centering
\textbf{\scriptsize Ours, 4 i/p views}\par\medskip
{\includegraphics[scale=0.2,trim={0 0 0 0},clip]{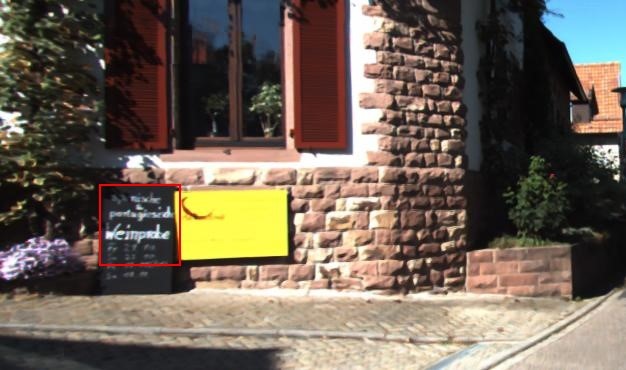}}
{\includegraphics[scale=0.2,trim={0 0 0 0},clip]{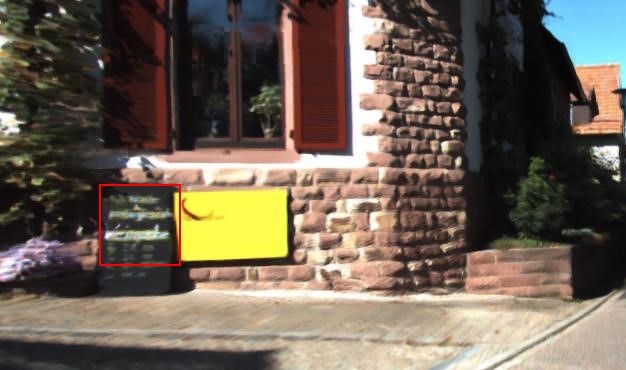}}
{\includegraphics[scale=0.2,trim={0 0 0 0},clip]{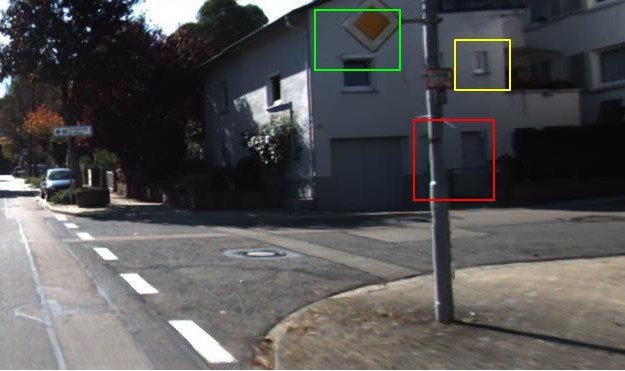}}
{\includegraphics[scale=0.2,trim={0 0 0 0},clip]{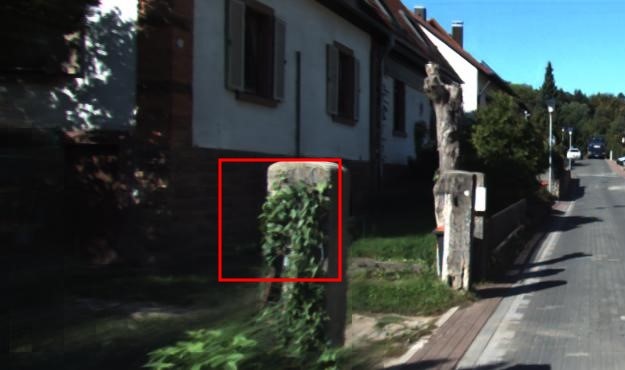}}
\end{minipage}\hfill
\begin{minipage}{0.18\textwidth}
\centering
\textbf{\scriptsize DeepStereo}\par\medskip
{\includegraphics[scale=0.15,trim={0 0 0 0},clip]{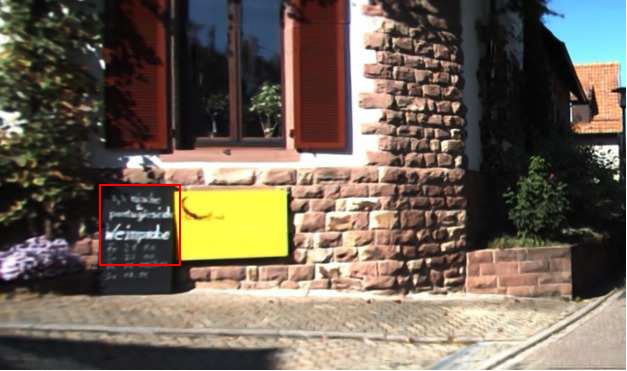}}
{\includegraphics[scale=0.15,trim={0 0 0 0},clip]{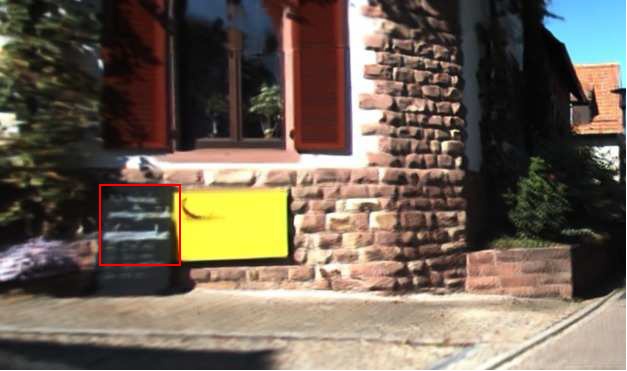}}
{\includegraphics[scale=0.15,trim={0 0 0 0},clip]{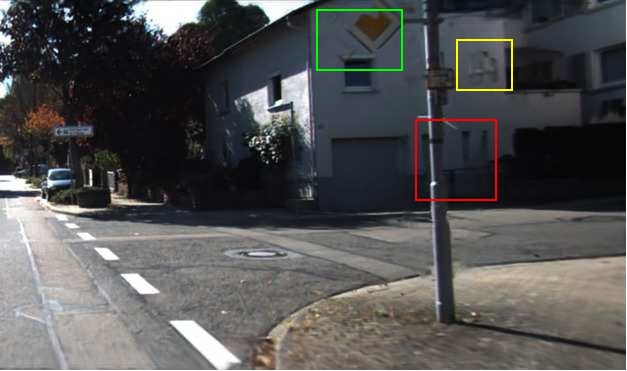}}
{\includegraphics[scale=0.15,trim={0 0 0 0},clip]{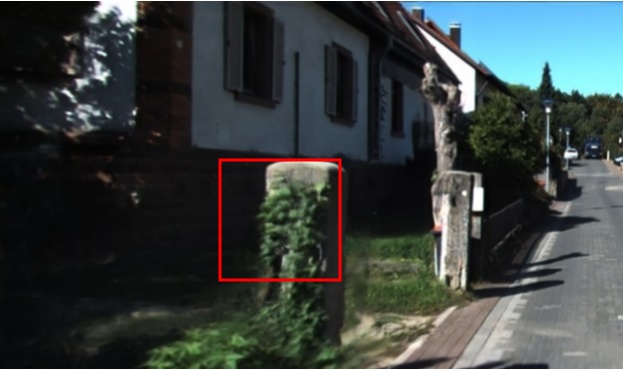}}
\end{minipage}\hfill
\begin{minipage}{0.18\textwidth}
\centering
\textbf{\scriptsize Ground Truth}\par\medskip
{\includegraphics[scale=0.20,trim={0 0 0 0},clip]{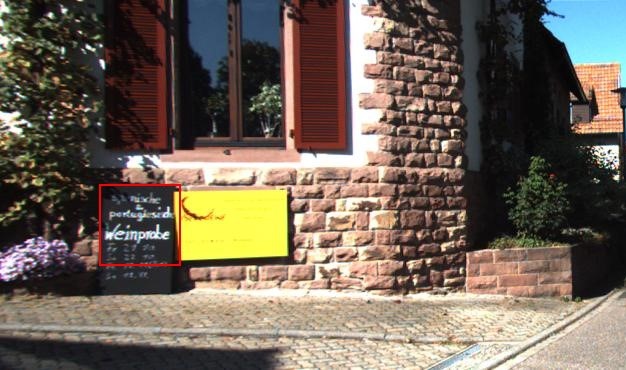}}
{\includegraphics[scale=0.20,trim={0 0 0 0},clip]{./kitti/crops/done/mr64dd_8_gt_894.jpg}}
{\includegraphics[scale=0.20,trim={0 0 0 0},clip]{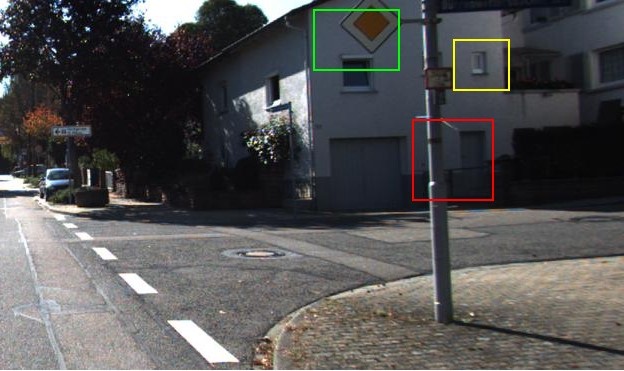}}
{\includegraphics[scale=0.20,trim={0 0 0 0},clip]{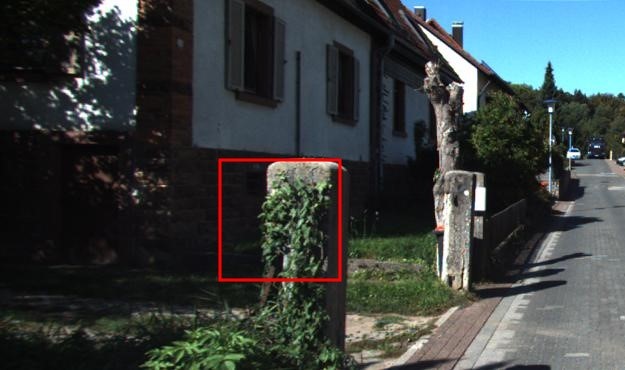}}
\end{minipage}\hfill
\caption{ The figure above show view synthesis results for proposed method and DeepStereo~\cite{flynn2016deepstereo} on KITTI dataset. From left to right, each column show novel views estimated using proposed method (columns 1-3), DeepStereo and original image. The median baseline between input cameras is shown on left. The figure shows that estimates from the proposed method have better perceptual quality and is free from artifacts in comparison to the baseline (refer the highlighted regions).}
\label{fig:DeepStereo}
\end{figure*}

\begin{table*}[t]
\centering
\begin{tabular}{c c c c c c c c c}
\hline
  &  & 	 & \multicolumn{6}{c}{Ours} \\
  &  & 	 & \multicolumn{2}{c}{4 i/p views} & \multicolumn{2}{c}{3 i/p views} & \multicolumn{2}{c}{2 i/p views}\\
baselines & Habtegebrial \textit{et al.}  & DeepStereo & w/ mr & w/o mr & w/ mr & w/o mr & w/ mr & w/o mr \\
\hline
0.8m & 6.66 & 7.49 & 8.68 & 8.90 & 8.98 & 9.17 & 10.32 & 10.46 \\
1.6m & 8.90 & 10.41 & 11.67 & 12.11 & 12.12 & 12.54 & 13.79 & 14.12\\
\hline
\end{tabular}
\caption{The table show performance comparison of proposed method with DeepStereo~\cite{flynn2016deepstereo} and Habtegebrial \textit{et al.} \cite{habtegebrial2018fast} in terms of L1 prediction error (lower the better) for different baselines on KITTI dataset. We show the performance of our approach using different numbers of input images. The table shows that performance improves with the number of input images and proposed multi-resolution analysis improves it further.}
\label{tab:l1_err}
\end{table*}


\begin{figure*}[t]
\centering
{\includegraphics[scale=0.25,trim={0 265 0 0},clip]{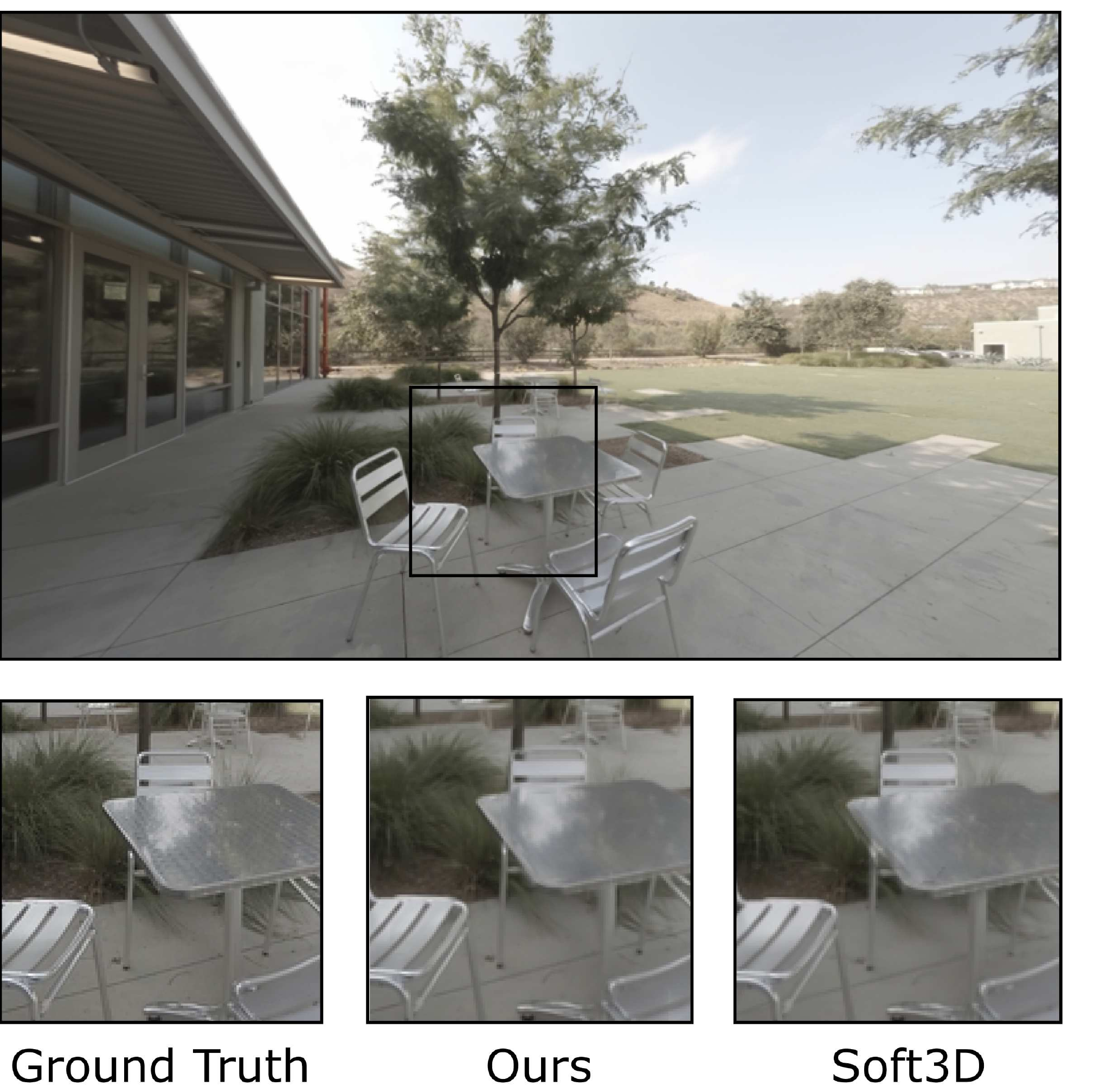}}~
{\includegraphics[scale=0.25,trim={0 265 0 0},clip]{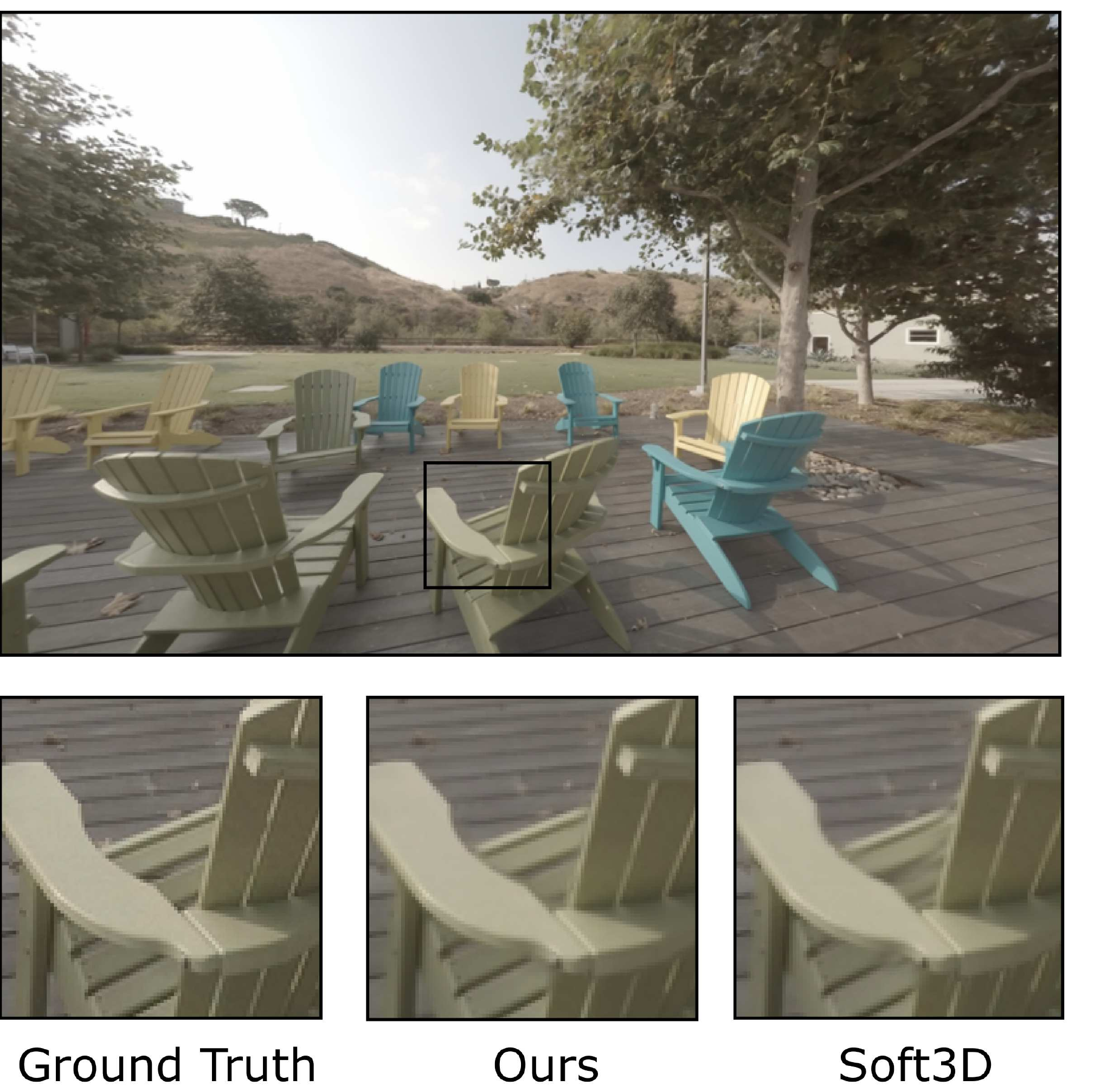}}~
{\includegraphics[scale=0.25,trim={0 265 0 0},clip]{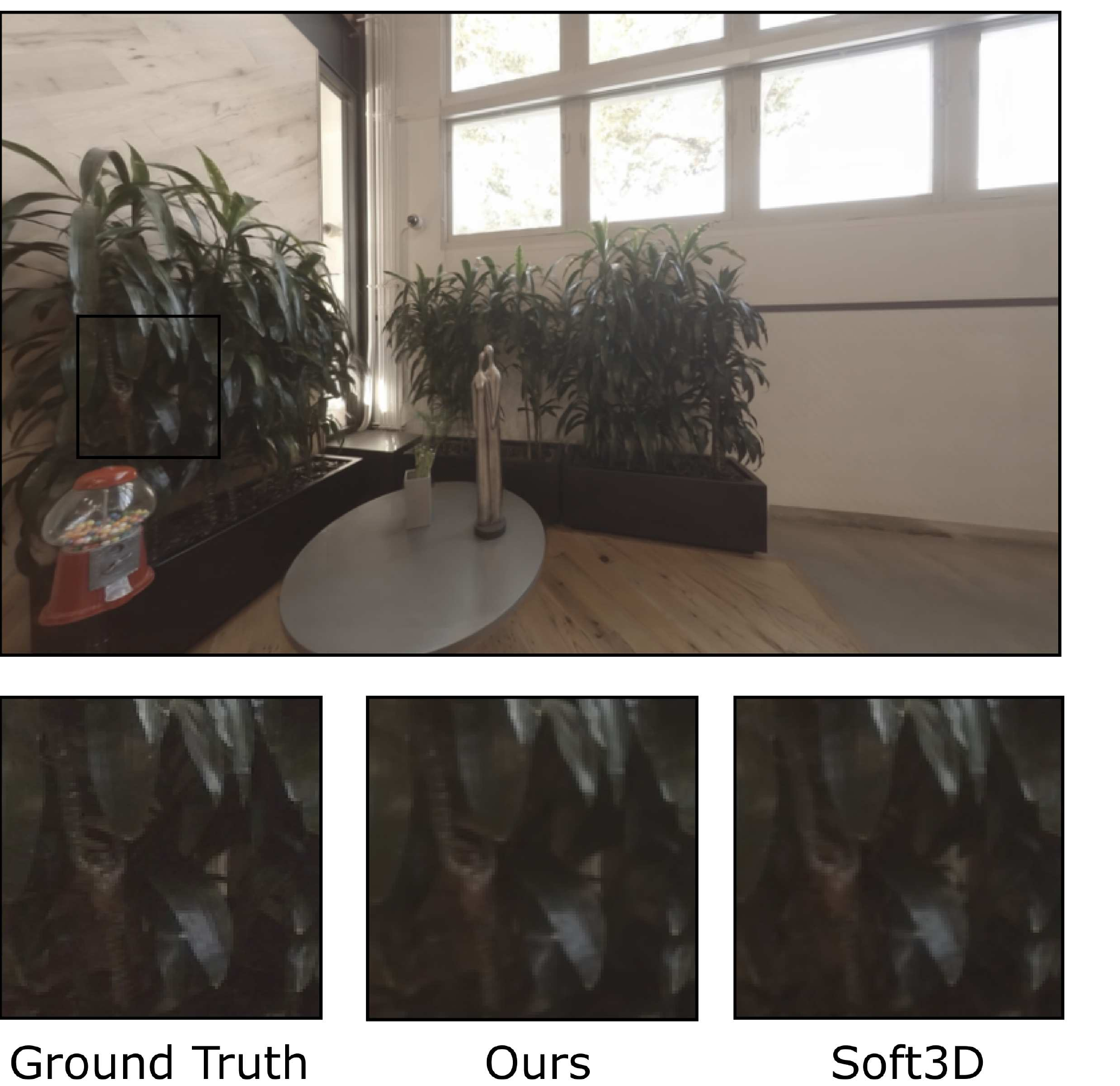}}\\
\caption{ We show novel view estimates using $64$ depth levels and using multi-resolution analysis on Spaces dataset. We also show cropped regions from the estimated novel views for comparative analysis with Soft3D~\cite{penner2017soft}. }
\label{fig:spaces}
\end{figure*}

\begin{figure*}
\begin{minipage}{0.2\textwidth}
\centering
\subfloat[]{\includegraphics[scale=1.50,trim={0 0 0 0},clip]{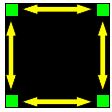}{\label{fig:4nbr}}}~
\subfloat[]{\includegraphics[scale=1.50,trim={0 0 0 0},clip]{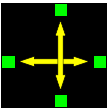}{\label{fig:8nbr}}}\\
\subfloat[]{\includegraphics[scale=1.50,trim={0 0 0 0},clip]{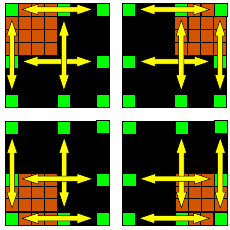}{\label{fig:8_6nbr}}}
\end{minipage}\hfill
\begin{minipage}{0.8\textwidth}
\centering
{\includegraphics[scale=1.0,trim={0 0 0 0},clip]{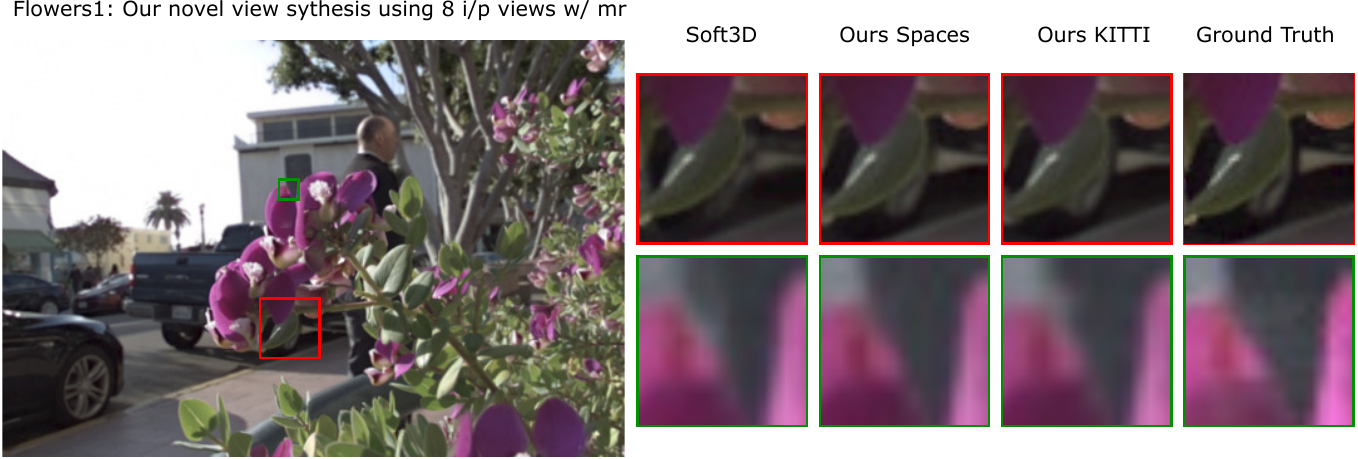}}\\
\subfloat[]{\includegraphics[scale=1.0,trim={0 0 0 0},clip]{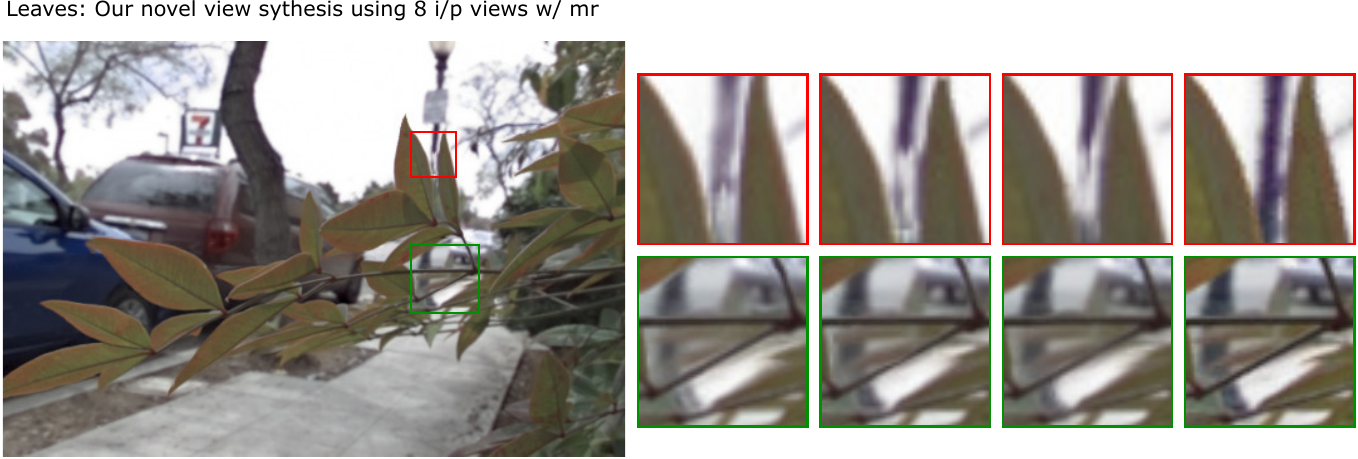}{\label{fig:lytro}}}\\
\end{minipage}\hfill
\caption{ (a) We create $4$ image pairs using $4$ images at the corner and synthesize entire $8$x$8$ light field. 
(b) We consider $2$ additional image pairs as inputs shown here making total $8$ images and $6$ pairs (including pairs in a \& b together). 
(c) We choose the subset of $4$ image pairs closest to the target view. Different positions of the target view (shown in orange) and corresponding input image pairs are shown. 
(d) The figure show view synthesis results for Spaces trained model on Flowers1 and Leaves scene from light field dataset of Kalantari \textit{et al.} \cite{kalantari2016learning}. 
}
\end{figure*}

\begin{table}[t]
\centering
\begin{tabular}{c c c c c}
\hline
 &  &  & \multicolumn{2}{c}{Ours} \\
Inputs/Baselines & \cite{zhou2018stereo} & Soft3D & w/ mr & w/o mr\\
\hline

4-view (small) & 0.8884 & 0.9260  & 0.9417 & 0.9372 \\
4-view (medium) & 0.8874 & 0.9300 & 0.9423 & 0.9379 \\
4-view (large) & 0.8673 & 0.9315 & 0.9379 & 0.9327 \\
12-view 					& -- & 0.9402 & \textbf{0.9553} & 0.9521 \\
\hline
\end{tabular}
\caption{ We show the performance in terms of SSIM (higher the better) on Spaces dataset. }
\label{tab:spaces}
\end{table}

\begin{table*}[t]
\centering
\begin{tabular}{c c c c c c c c c}
\hline
 & & & \multicolumn{3}{c}{Ours KITTI Trained} & \multicolumn{3}{c}{Ours Spaces Trained}\\
Scene & Kalantari \textit{et al.}  & Soft3D & 4 views & 8 views & 8 views* & 4 views & 8 views & 8 views* \\
\hline
Flowers1 & 0.9480 & 0.9581 & 0.9516 & 0.9567 & 0.9621 & 0.9596 & 0.9650 & 0.9671 \\
Flowers2 & 0.9507 & 0.9616 & 0.9523 & 0.9579 & 0.9632 & 0.9630 & 0.9683 & 0.9702 \\
Cars 	 & 0.9700 & 0.9705 & 0.9678 & 0.9719 & 0.9751 & 0.9725 & 0.9764 & 0.9776 \\
Rock 	 & 0.9555 & 0.9595 & 0.9559 & 0.9609 & 0.9649 & 0.9628 & 0.9671 & 0.9687 \\
Leaves 	 & 0.9282 & 0.9525 & 0.9325 & 0.9406 & 0.9494 & 0.9481 & 0.9568 & 0.9605 \\
\hline
Average  & 0.9505 & 0.9604 & 0.9520 & 0.9576 & 0.9629 & 0.9612 & 0.9667 & \textbf{0.9688} \\
\hline
\end{tabular}
\caption{ 
We show the performance on light field dataset of Kalantari \textit{et al.} \cite{kalantari2016learning} in the table. We show SSIM (higher the better) on different scenes in the dataset. We note that our model trained on KITTI and Spaces datasets generalizes very well on this dataset. Please refer the text for more details.
}
\label{tab:lytro}
\end{table*}

\section{Experiments}
We substantiate the performance of the model using several datasets. 
We use KITTI~\cite{Geiger2012CVPR} and Spaces~\cite{flynn2019deepview} datasets for training. 
Further, in order to show robustness of the proposed model we also validate the performance on light field dateset of Kalantari \textit{et al.} \cite{kalantari2016learning}. 

\subsection{Results on KITTI dataset}
We use settings similar to that of \cite{flynn2016deepstereo} for analysis and show the performance on KITTI test split. 
We consider sequence of $5$ consecutive images from KITTI dataset as one training example. 
The middle image is used as a target and remaining four images in the sequence are used as inputs. 
We create $3$ image pairs by grouping adjacent images. 
Thus inputs to our model are $3$ image pairs which generates $3$ novel view estimates, one at a time, which are then combined using predicted occlusion masks. 
The inputs to the model are patches of size $112$x$112$ from PSVs and output is a patch of size $32$x$32$. 
We use minibatch size of $48$ image pairs for training which contains $16$ unique novel views ($3$ predictions per novel view) to be synthesized. 
During each iteration we use half the samples from previous minibatch for multi-resolution analysis using pdfs estimated in that iteration. 
Thus our minibatch consists of $8$ views estimated with original depth levels and $8$ view estimated with resampled depth levels. 

We train the model in two phases. 
First, we use only $16$ discrete depth levels. 
For this phase the Depth PDF Estimation module is replaced with a smaller model to predict $16$ class probability distribution. 
We train this model for $1$mn iterations. 
We then replace this module by a module with $64$ depth levels. 
We freeze the weights of Feature Extraction and Feature Correlation module and train only the Depth PDF Estimation module for $250$k iterations. 
We use SELU activation, Adam optimizer with learning rate of $1$e$-5$ and gradient clipping of $1.0$. 
We provide more details about the model in supplementary material. 

We show comparative analysis of proposed method with DeepStereo~\cite{flynn2016deepstereo} in Fig.~\ref{fig:DeepStereo}. 
We show the estimated novel views using proposed approach when using different numbers of input images. 
We use $4$, $3$ and $2$ inputs to create $3$, $2$ and $1$ image pairs, respectively. 
We show estimated novel views when we perform multi-resolution (mr) analysis and re-estimate the novel view. 
The median camera baseline for respective examples is shown on the left side in the figure. 
From the figure one may notice that the qualitative performance of the proposed method is comparable to DeepStereo for $0.8$m baseline. 
For larger baselines both methods start to shown some distortions in the estimates, however, proposed method show relatively less distortions as can be seen in the highlighted regions in the figure. 

We show quantitative performance evaluation for all the methods in Table~\ref{tab:l1_err} in terms of  L1 prediction error. 
All the models shown in the table have been trained with input images with median baseline of $0.8$m. 
One may notice that as the number of input views increases, corresponding accuracy for the proposed method improves. 
It should be noted that the baseline methods do not have the flexibility and fail to work when lower number of  input views are available. 
For example, Habtegebrial \textit{et al.} needs $4$ stereo pairs and DeepStereo needs $4$ images of the scenes. 
On the other hand proposed method can work with less number of images and shows competitive performance. 
Further the performance also improves when we re-estimate the view using proposed multi-resolution analysis as shown in the table. 
One may notice from the table that the quantitative performance of proposed method (with mr and using 4 i/p views) is comparable to other methods, although slightly lower. 
One of the reason, in our opinion, for the reduced performance is that proposed method estimates novel view and occlusion mask independently using only given image pair at any given time.
Thus the model have to learn to merge different estimates of the novel view by qualitatively asserting the estimates using predicted occlusion masks which is generally a difficult task. 
For example, while establishing the stereo correspondence there could be many false positives which may reduce the performance. 
On the other hand, the baseline methods uses all images simultaneously for the prediction and as a result performs better. 
Notwithstanding the above, one may notice from the table and from the figure that the performance of the proposed method is comparable while being flexible in its use case. 

\subsection{Results on Spaces dataset}
We also train our model on recently proposed Spaces~\cite{flynn2019deepview} dataset. 
The dataset captures light field images of $100$ different scenes using array of 16 cameras. 
We use $90$ scenes for training and $10$ for testing. 
We randomly sample $4$ pairs of images from camera array as inputs and sample the target image from remaining cameras. 
We use the same training procedure as used for training on KITTI dataset. 
We show the results in Table~\ref{tab:spaces}. 
We validate the performance with $3$ different baselines using $4$ input views (to create $6$ pairs). 
Further, we also show results when using $12$ images in the camera grid as inputs to synthesize remaining images. 
We combine neighboring input images depending on the position in the camera grid to create $18$ image pairs. 
The last row of the table show that the performance improves significantly when we use large number of images for analysis. 
We note that the model is trained using only $4$ image pairs and generalizes well when using large numbers of image pairs.  
We show in Fig.~\ref{fig:spaces} novel view synthesis using $12$ input images for qualitative analysis. 

\subsection{Results on Light Field dataset}
In this section we validate the performance on light field dataset of Kalantari \textit{et al.} \cite{kalantari2016learning}. 
We show the performance in Table~\ref{tab:lytro} for KITTI trained model (columns 4-6) and Spaces trained model (columns 7-9). 
We compare the SSIM performance with Kalantari \textit{et al.} \cite{kalantari2016learning} and Soft3D \cite{penner2017soft}. 
We show our results when using $4$ images at the corner of $8$x$8$ grid as inputs in the table (colums 4,7).
We create $4$ image pairs as shown in Fig.~\ref{fig:4nbr}. 
We consider using $2$ additional image pairs ($6$ image pairs in total) as shown in Fig.~\ref{fig:8nbr} making total number of inputs $8$. 
The table show that using additional images for analyses improves the performance (colum 5,8). 
We also consider an experiment where we dynamically choose input image pairs based on the location of target cameras. 
We consider using $4$ nearest image pairs as shown in Fig.~\ref{fig:8_6nbr} depending on the target camera position in the quadrant and show the performance (colum 6,9). 
As we ignore images farther from the target camera and only consider images which are closer accuracy of view estimates improves further. 
We show the visual comparison of the estimated images for qualitative analysis in Fig.~\ref{fig:lytro}. 

\section{Ablation Studies}
\subsection{Effect of different numbers of depth levels}
In this section we validate the effect of using different numbers of depth levels on the quality of estimated novel view. 
We show the performance when using $64$ and $16$ depth levels in Table~\ref{tab:ablation_mr_kitti} for KITTI dataset and in Table~\ref{tab:ablation_mr_lytro} for light field dataset of Kalantari \textit{et al.} \cite{kalantari2016learning}. 
We note from the table that on KITTI dataset using $16$ depth levels with multi-resolution analysis improves the performance significantly (comparable to using $64$ depth levels). 
On contrary, we do not see such significant improvement on lytro dataset shown in Table~\ref{tab:ablation_mr_lytro}. 
This is due to, in our opinion, the fact that lytro dataset presents scenes with high depth complexity. 
In such cases the initial pdf estimated using only $16$ depth levels is quite noisy to correctly estimate the depth probabilities and hence depth resampling  does not improve the performance. 

\begin{table}[t]
\centering
\begin{tabular}{c c c c c}
\hline
  &  \multicolumn{2}{c}{64 depth levels} & \multicolumn{2}{c}{16 depth levels} \\
baselines & w/ mr & w/o mr & w/ mr & w/o mr \\
\hline
0.8m & 8.68 & 8.90 & 8.99 & 10.16 \\
1.6m & 11.67 & 12.11 & 11.96 & 13.71 \\
\hline
\end{tabular}
\caption{ We show the performance (L1 prediction error) on KITTI dataset when using different depth levels. 
}
\label{tab:ablation_mr_kitti}
\end{table}

\begin{table}[t]
\centering
\begin{tabular}{c c c c c}
\hline
  &  \multicolumn{2}{c}{64 depth levels} & \multicolumn{2}{c}{16 depth levels} \\
 & w/ mr & w/o mr & w/ mr & w/o mr \\
\hline
4 views & 0.9520 & 0.9470 & 0.9377 & 0.9327 \\
8 views & 0.9576 & 0.9542 & 0.9427 & 0.9389 \\
\hline
\end{tabular}
\caption{The performance (SSIM) analysis using different depth levels on the light field dataset of Kalantari \textit{et al.} \cite{kalantari2016learning} is shown. 
}
\label{tab:ablation_mr_lytro}
\end{table}

\subsection{Validating the occlusion masks}
\begin{figure}
\center
{\includegraphics[scale=.225,trim={0 0 0 0}, clip]{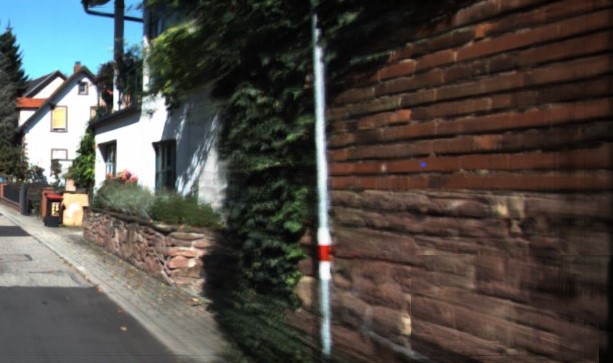}}~
{\includegraphics[scale=.225,trim={0 0 0 0}, clip]{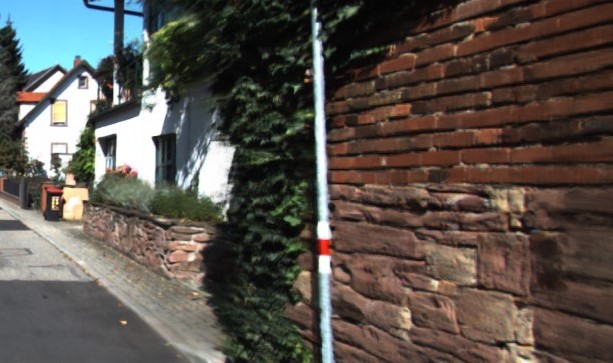}}
\caption{The left figure show results when using our trained version of DeblurGAN to combine multiple novel views. Result using occlusion mask are shown on right. }
\label{fig:ablation_gan}
\end{figure}

One may train a CNN model to learn and combine multiple novel view estimates into single optimal prediction instead of using occlusion masks. 
We performed this ablation study to validate the usefulness of occlusion masks learned by the model. 
For this task, we modified the DeblurGAN \cite{kupyn2017deblurgan} network to accept two novel view estimates as input and predict the view which is free from the artifacts of the input views. 
We trained the model using novel view patches synthesized using our model with $64$ depth levels without multi-resolution analysis. 
We show corresponding results in Fig.~\ref{fig:ablation_gan}. 
The figure show that novel view prediction using occlusion masks (shown on right) have better quality. 

\section{Discussion and Limitations}
As proposed method estimates each novel view from given image pair independently it can work with different numbers of input images. 
This may lead to slightly lower performance in some cases in comparison to the baseline methods which uses all input images simultaneously for analysis (Table~\ref{tab:l1_err}). 
For example, in KITTI dataset images have significant amount of occlusions and hence final performance is slightly lower than baseline methods. 
Analyzing the information from all inputs while being flexible to handle different numbers of inputs should improve the overall performance. 
We notice that thin foreground objects with large disparities are often poorly reconstructed and even disappear sometime. 
When using multi-resolution analysis, we notice blocking artifacts around the regions near depth discontinuity. 
Further, when using images farther from target camera the accuracy of occlusion masks reduces due to many false positives. 
As a result including many images (with large baseline) for analyses may reduce the performance. 
We notice this effect in Table.~\ref{tab:lytro} (columns 5-6 and 8-9) where performance improves when we consider only a subset of $4$ closer image pairs for analyses. 

\section{Conclusions}
In this work, we have presented an approach for novel view synthesis which can use different numbers of images of the scene. 
The presented approach estimates novel view from a pair of images independently and improves the estimate further by combining multiple estimates of the view. 
The proposed approach is more flexible in comparison to the prior works and hence is suitable to more general settings when number of input images can vary arbitrarily. 
The model is able to handle occlusions and partial scene visibility in input images gracefully by estimating occlusion masks. 
The results show that the model have competitive performance on several datasets and generalizes on unseen datasets as well. 

{\small
\bibliographystyle{ieee_fullname}
\bibliography{egbib}
}

\appendix
\section{Supplementary Material}
\subsection{Model Architecture}
Our CNN models are fully convolutional. 
Table~\ref{tab:fe} show details of the Feature Extraction model shown in Fig.~2a from main article. 
This is a $10$ layer standard CNN model. 
The input to this model is a 3-channel rgb image plane from given PSV. 
The kernel size and output channels of each layer are shown in the table. 
Similarly, we show the details of Feature Correlation and Depth PDF Estimation model (shown in Fig.~2b and 2c in main article) in Table~\ref{tab:fc} and Table~\ref{tab:de}, respectively. 
These are $8$ and $4$ layer models, respectively. 
The input to the Feature Correlation model is a $256$ channel feature from specific depth plane. 
The input to the Depth PDF Estimation model is $1024$ channel feature matching score corresponding to all $64$ depth planes. 
All convolution layers are followed by SELU non-linearity except for the last layer of the Depth PDF Estimation model. 
Depth PDF Estimation model with $16$ depth level that we use during first phase of the training is shown in Table.~\ref{tab:de_16}. 

\subsection{Multi-resolution analysis}
During multi-resolution analysis we average the estimated depth pdf over a spatial region of $32$x$32$ size. 
We threshold the averaged pdf to estimate small range of depth levels with significant probabilities. 
Since the estimated pdf can be noisy at times, using very high threshold results in distortions and reduces the accuracy of the estimates. 
We empirically found that the threshold range between \( \frac{1}{200} \) to \( \frac{1}{30} \) is suitable for better performance. 
We select threshold of \( \frac{1}{100} \) for all the experiments shown in the paper. 

\begin{table*}[t]
\begin{minipage}{\linewidth}
\centering
\begin{tabular}{|| c | c c c c c c c c c c || }
\hline
kernel size & 13 & 13 & 13 & 13 & 9 & 9 & 5 & 5 & 5 & 5 \\
\hline
output channels & 8 & 8 & 16 & 16 & 32 & 32 & 64 & 64 & 128 & 128 \\
\hline
\end{tabular}
\caption{The table show CNN filter parameters for Feature Extraction model. Input to the model is a single rgb plane from PSVs. }
\label{tab:fe}
\end{minipage}

\begin{minipage}{\linewidth}
\centering
\begin{tabular}{|| c | c c c c c c c c || }
\hline
kernel size & 3 & 3 & 3 & 3 & 3 & 3 & 3 & 3 \\
\hline
output channels & 128 & 128 & 64 & 64 & 32 & 32 & 16 & 16 \\
\hline
\end{tabular}
\caption{The table show CNN filter parameters for Feature Correlation model. The input to the model is 256 channel feature from a pair of planes in given PSVs estimated by Feature Extraction model. }
\label{tab:fc}
\end{minipage}

\begin{minipage}{\linewidth}
\centering
\begin{tabular}{|| c | c c c c || }
\hline
kernel size & 3 & 3 & 3 & 3 \\
\hline
output channels & 512 & 256 & 128 & 67 \\
\hline
\end{tabular}
\caption{The CNN filter parameters for Depth PDF Estimation model are shown in the table above. The input to the model is 1024 channel feature matching score corresponding to all planes in PSV estimated by Feature Correlation model. }
\label{tab:de}
\end{minipage}

\begin{minipage}{\linewidth}
\centering
\begin{tabular}{|| c | c c c c || }
\hline
kernel size & 3 & 3 & 3 & 3 \\
\hline
output channels & 128 & 64 & 32 & 19 \\
\hline
\end{tabular}
\caption{The CNN filter parameters for Depth PDF Estimation model with $16$ depth levels are shown in the table above. The input to the model is 256 channel feature matching score corresponding to all planes in the PSV. }
\label{tab:de_16}
\end{minipage}%
\end{table*}

\subsection{Additional Results on KITTI dataset}
\textbf{Effect of different numbers of depth levels:}
We show estimated novel views using $64$ and $16$ depth levels, with and without using multi-resolution analysis in Fig.~\ref{fig:mr}. 
One may note from the figure that using multi-resolution analysis reduces blurring artifacts significantly. 
Further, the analysis significantly improves the performance even when using $16$ depth levels as can be observed by comparing second ($64$ depth levels w/ mr) and forth ($16$ depth levels w/ mr) columns. 

\begin{figure*}[h]
\centering
\textbf{\scriptsize Ground Truth \hspace{1.5cm} \scriptsize 64 depths w/ mr \hspace{1.5cm} \scriptsize 64 depths w/o mr \hspace{1.5cm} \scriptsize 16 depths w/ mr \hspace{1.5cm} \scriptsize 16 depths w/o mr}\par\medskip
{\includegraphics[scale=.08,trim={0 0 0 0},clip]{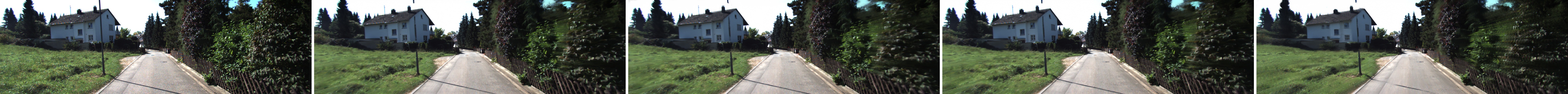}}\\
{\includegraphics[scale=.08,trim={0 0 0 0},clip]{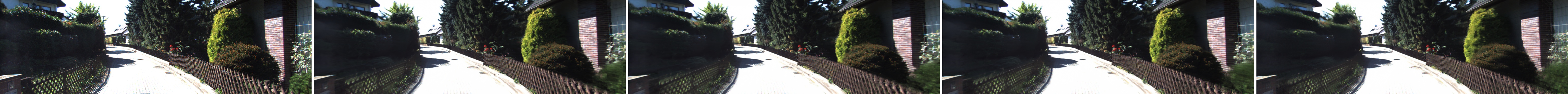}}\\
{\includegraphics[scale=.08,trim={0 0 0 0},clip]{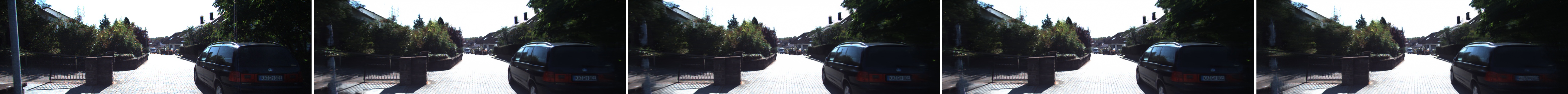}}\\
{\includegraphics[scale=.08,trim={0 0 0 0},clip]{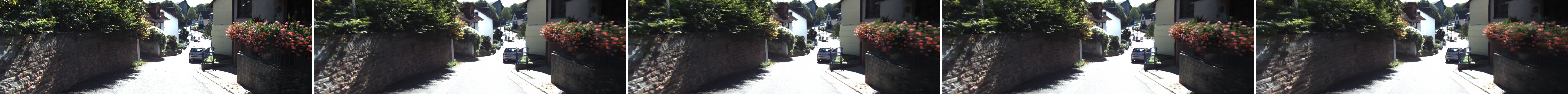}}\\
{\includegraphics[scale=.08,trim={0 0 0 0},clip]{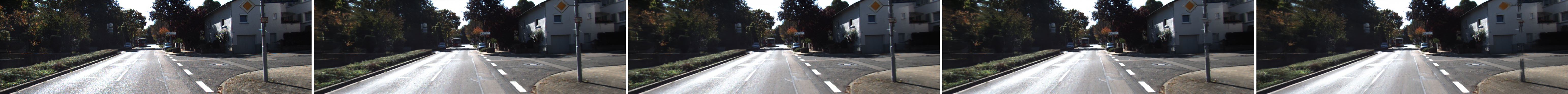}}\\
{\includegraphics[scale=.08,trim={0 0 0 0},clip]{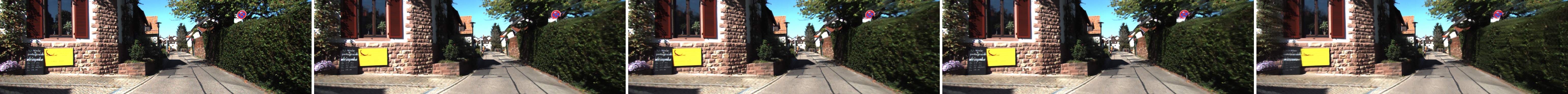}}\\
{\includegraphics[scale=.08,trim={0 0 0 0},clip]{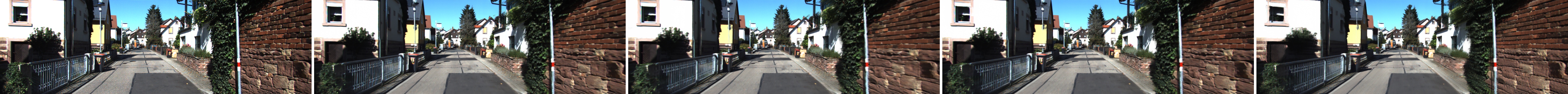}}\\
{\includegraphics[scale=.08,trim={0 0 0 0},clip]{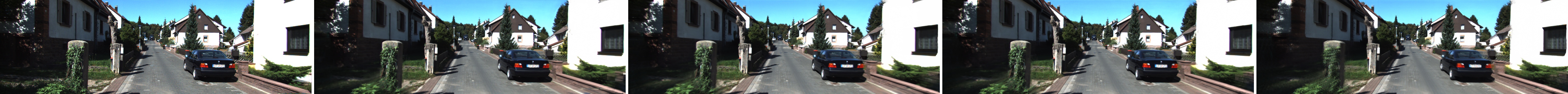}}
\caption{ We show effect of using different numbers of depth levels and multi-resolution analysis on view estimates in this figure. }
\label{fig:mr}
\end{figure*}

\begin{figure*}[h]
\centering
\textbf{\scriptsize 0.8m \hspace{5cm} \scriptsize 1.6m}\par\medskip
{\includegraphics[scale=.165,trim={0 0 0 0},clip]{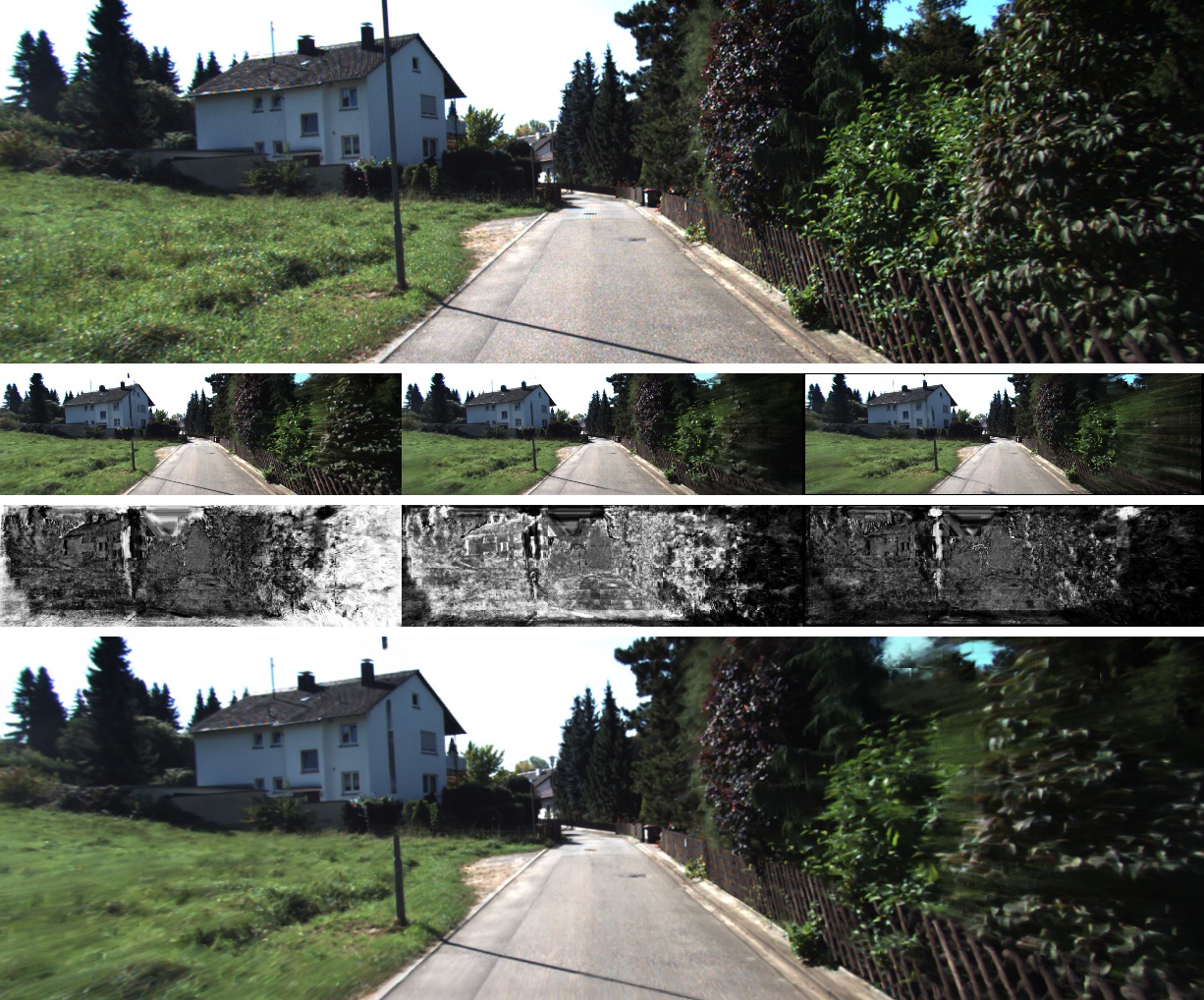}}~
{\includegraphics[scale=.165,trim={0 0 0 0},clip]{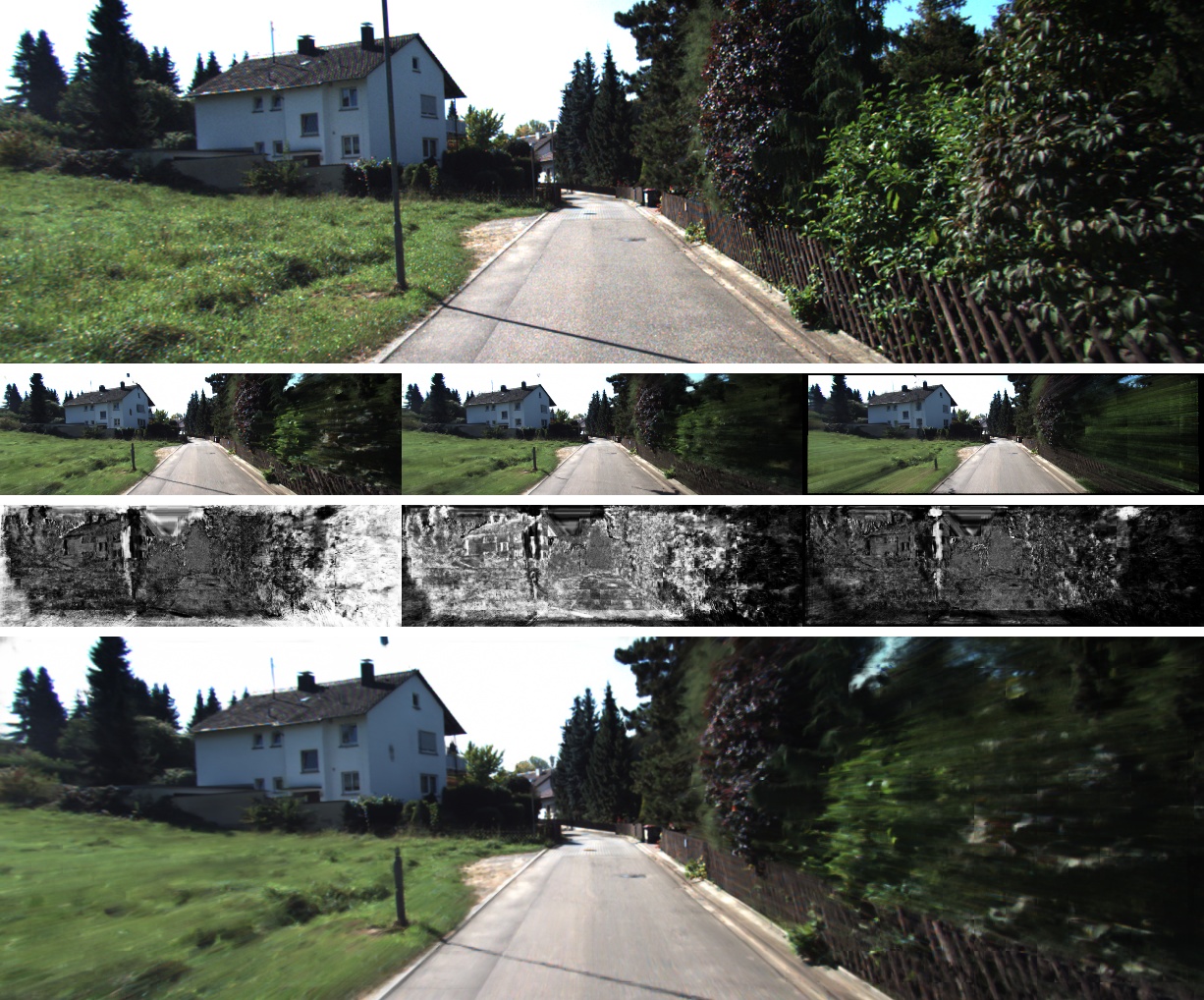}}\\
{\includegraphics[scale=.165,trim={0 0 0 0},clip]{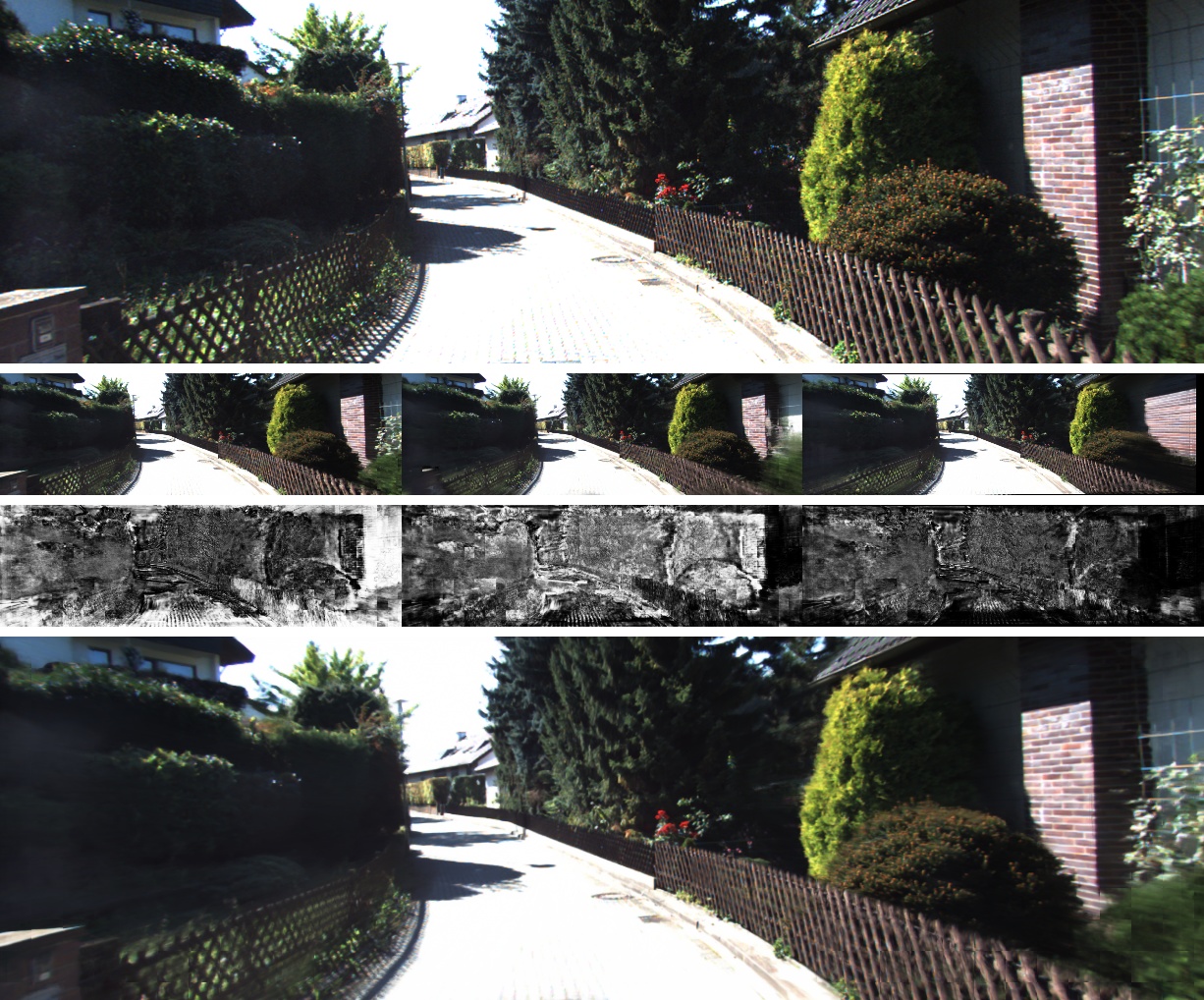}}~
{\includegraphics[scale=.165,trim={0 0 0 0},clip]{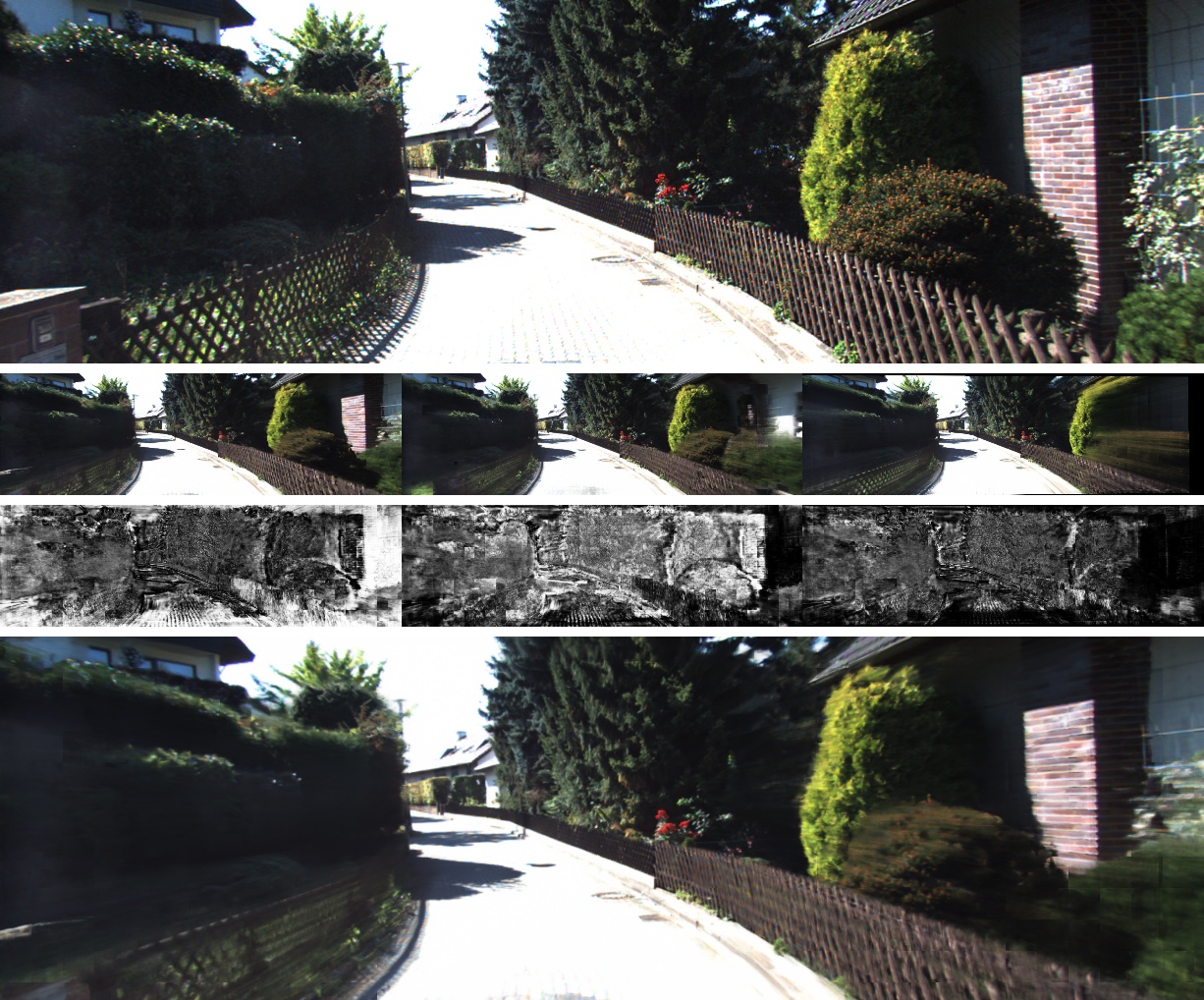}}
\caption{ The figure above shows novel view estimates using proposed method with different baselines.  
The ground truth view is shown in the first row, novel view estimates using $3$ different image pairs are shown in second row. 
We show estimated occlusion masks for each view in third row and final estimate is shown in the last row. 
One may now analyze the second and third row to compare usefulness of estimated occlusion masks. }
\label{fig:diff_baselines_1}
\end{figure*}

\begin{figure*}[h]
\centering
\textbf{\scriptsize 0.8m \hspace{5cm} \scriptsize 1.6m}\par\medskip
{\includegraphics[scale=.165,trim={0 0 0 0},clip]{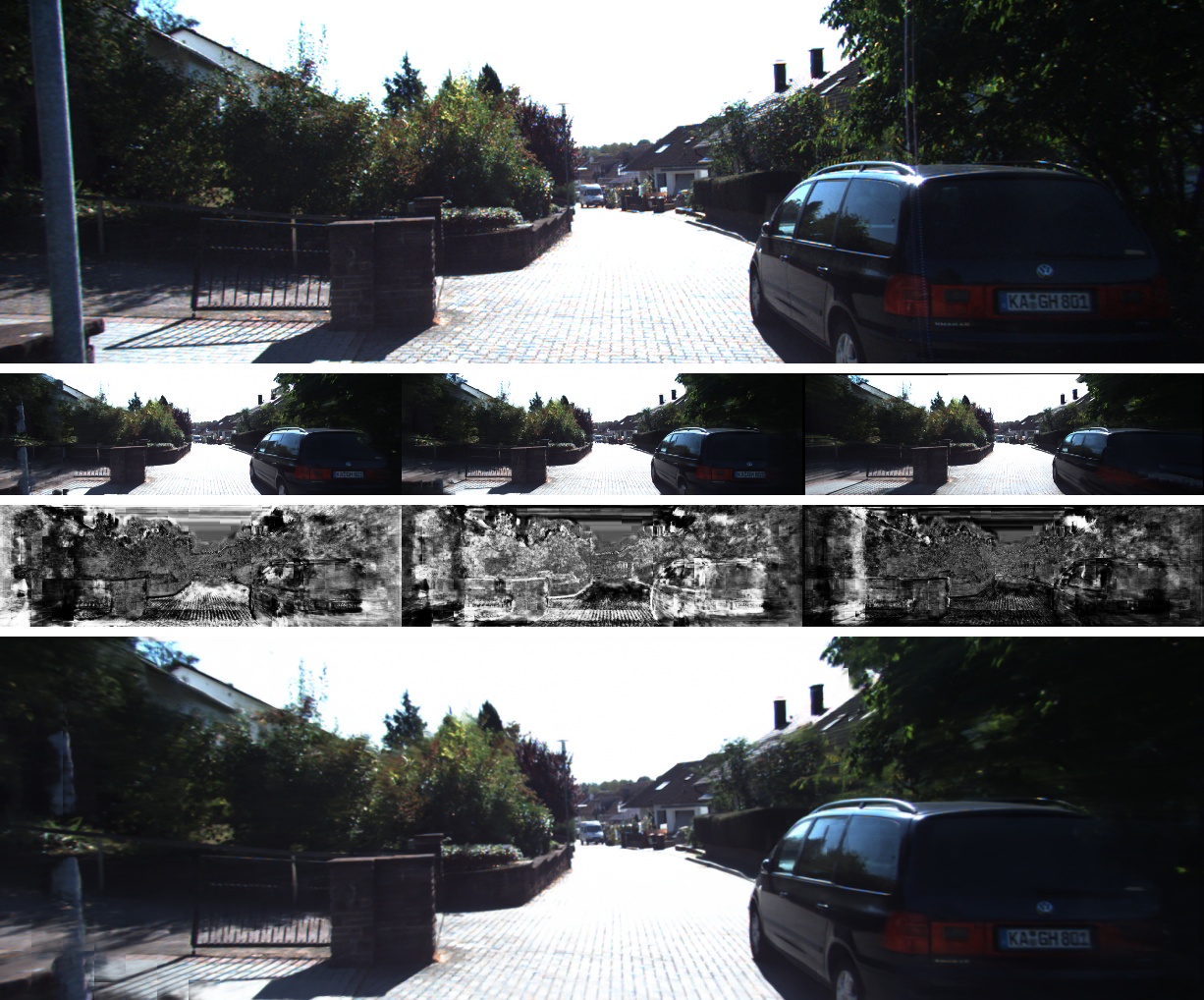}}~
{\includegraphics[scale=.165,trim={0 0 0 0},clip]{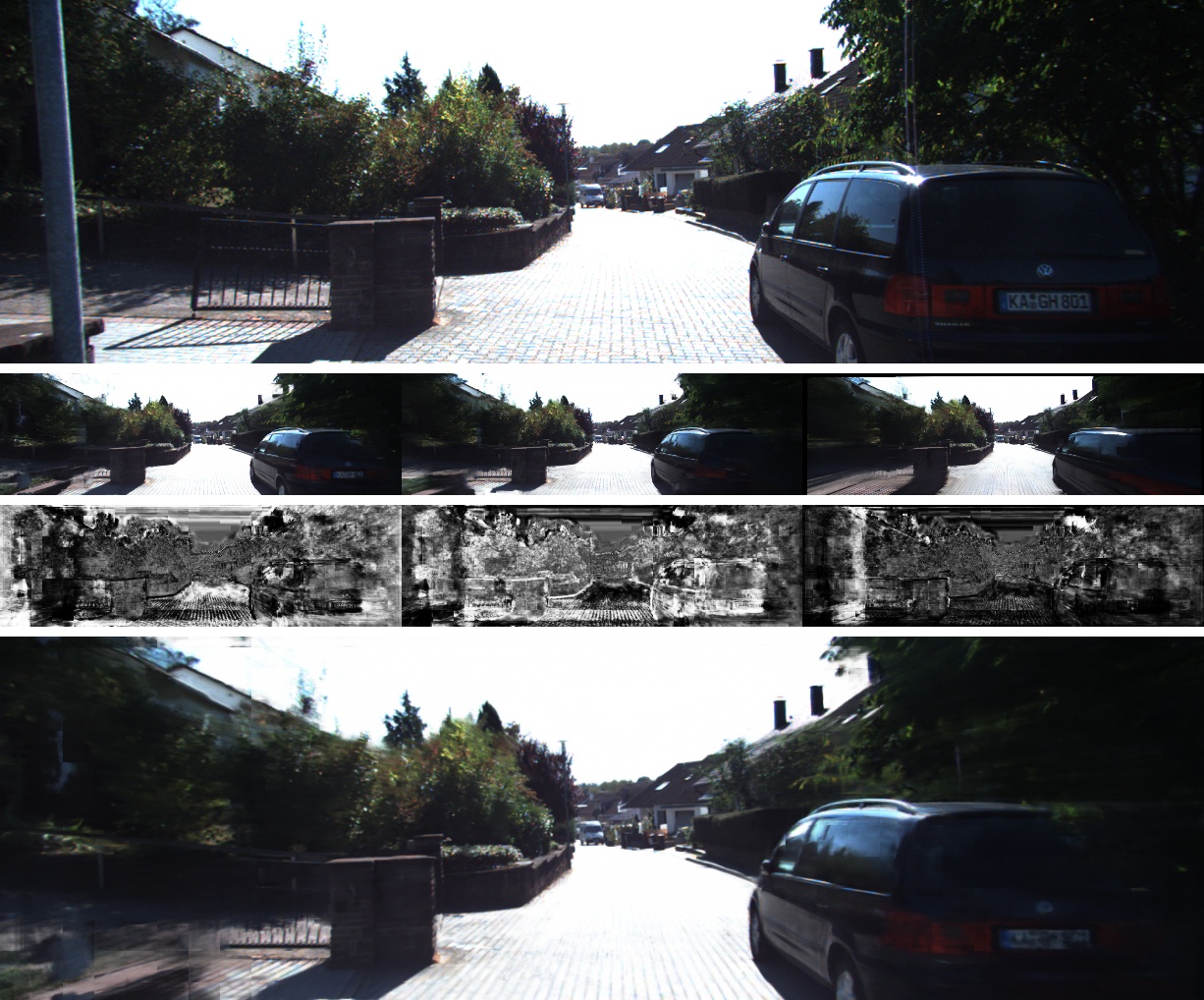}}\\
{\includegraphics[scale=.165,trim={0 0 0 0},clip]{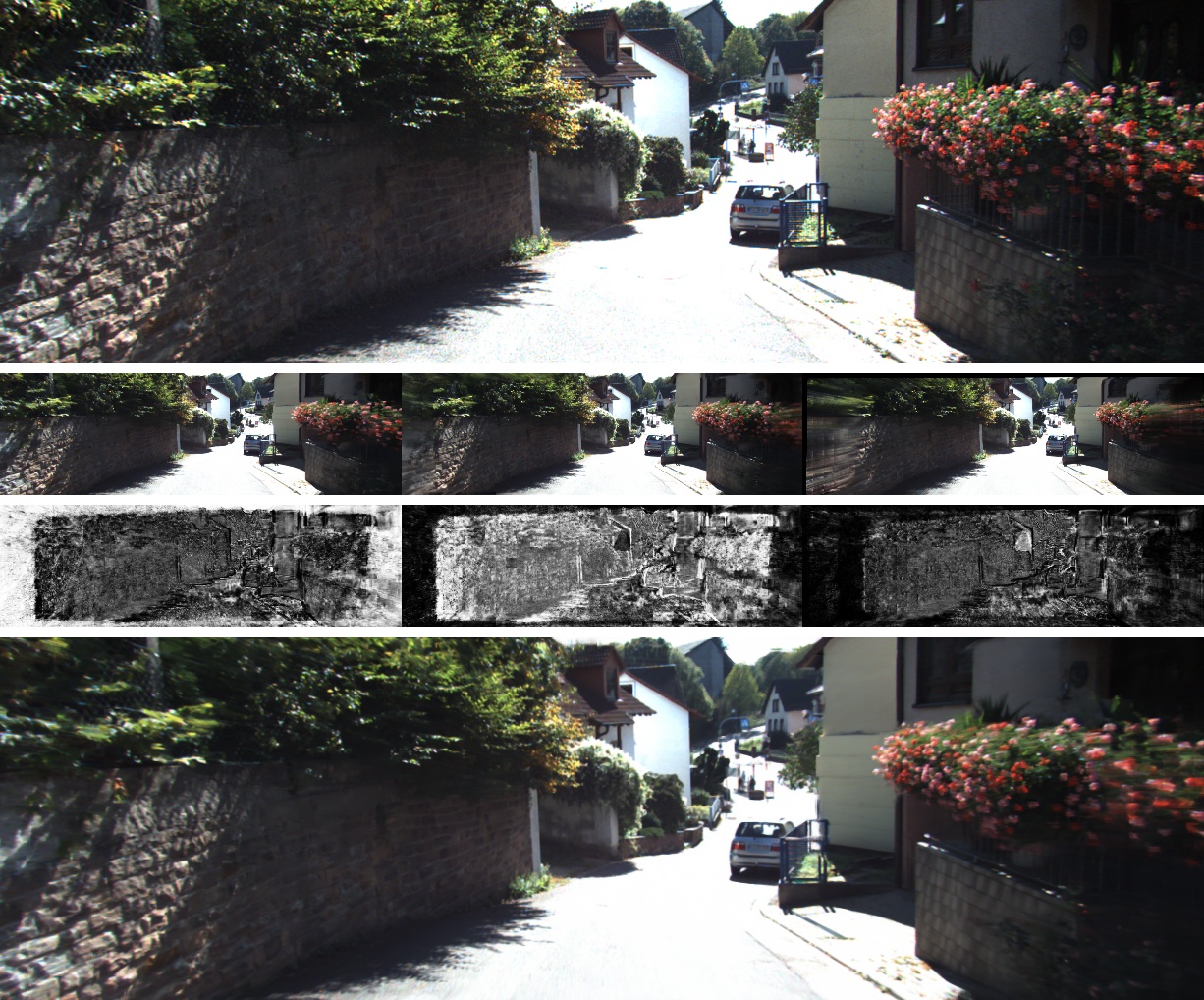}}~
{\includegraphics[scale=.165,trim={0 0 0 0},clip]{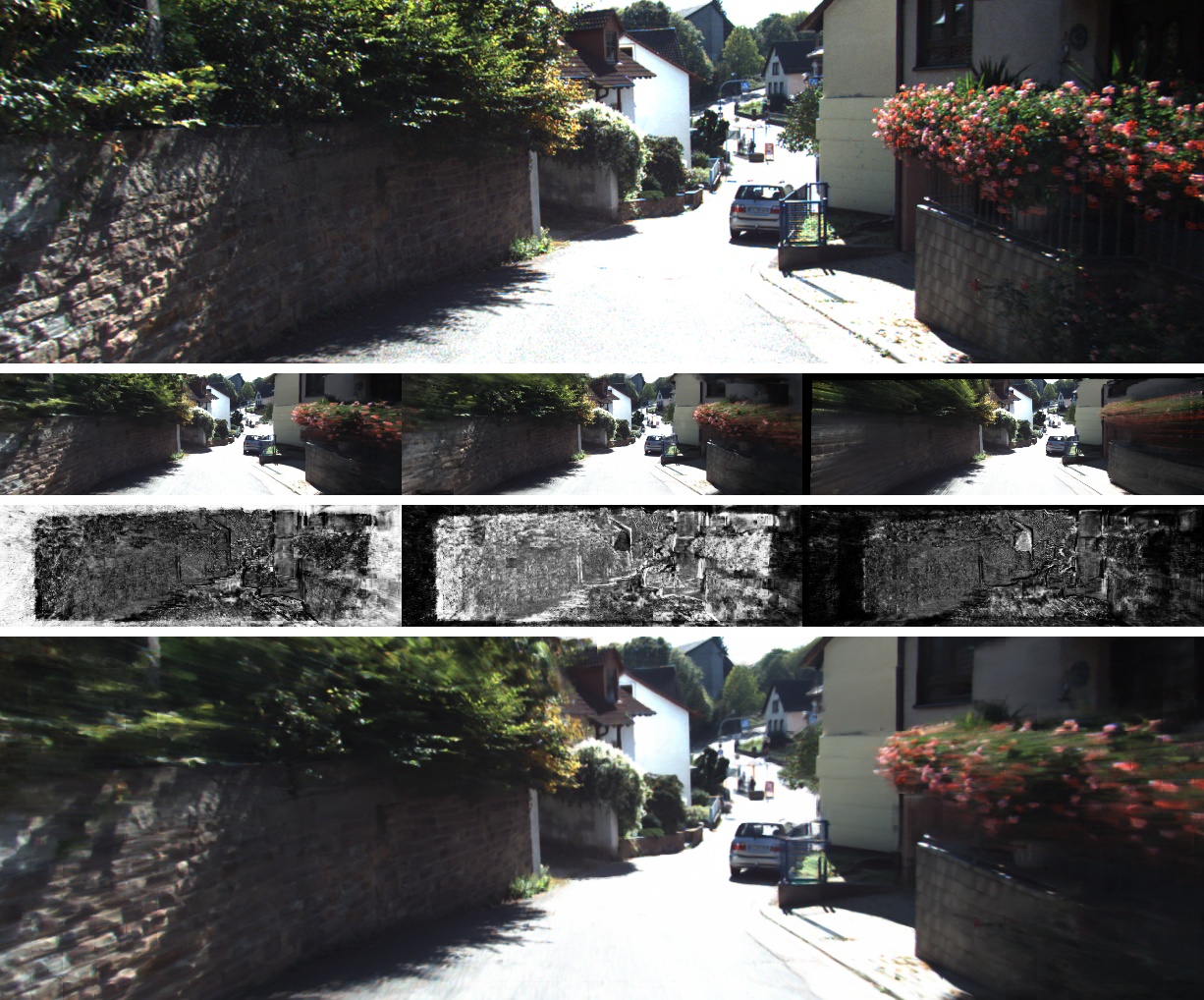}}\\
{\includegraphics[scale=.165,trim={0 0 0 0},clip]{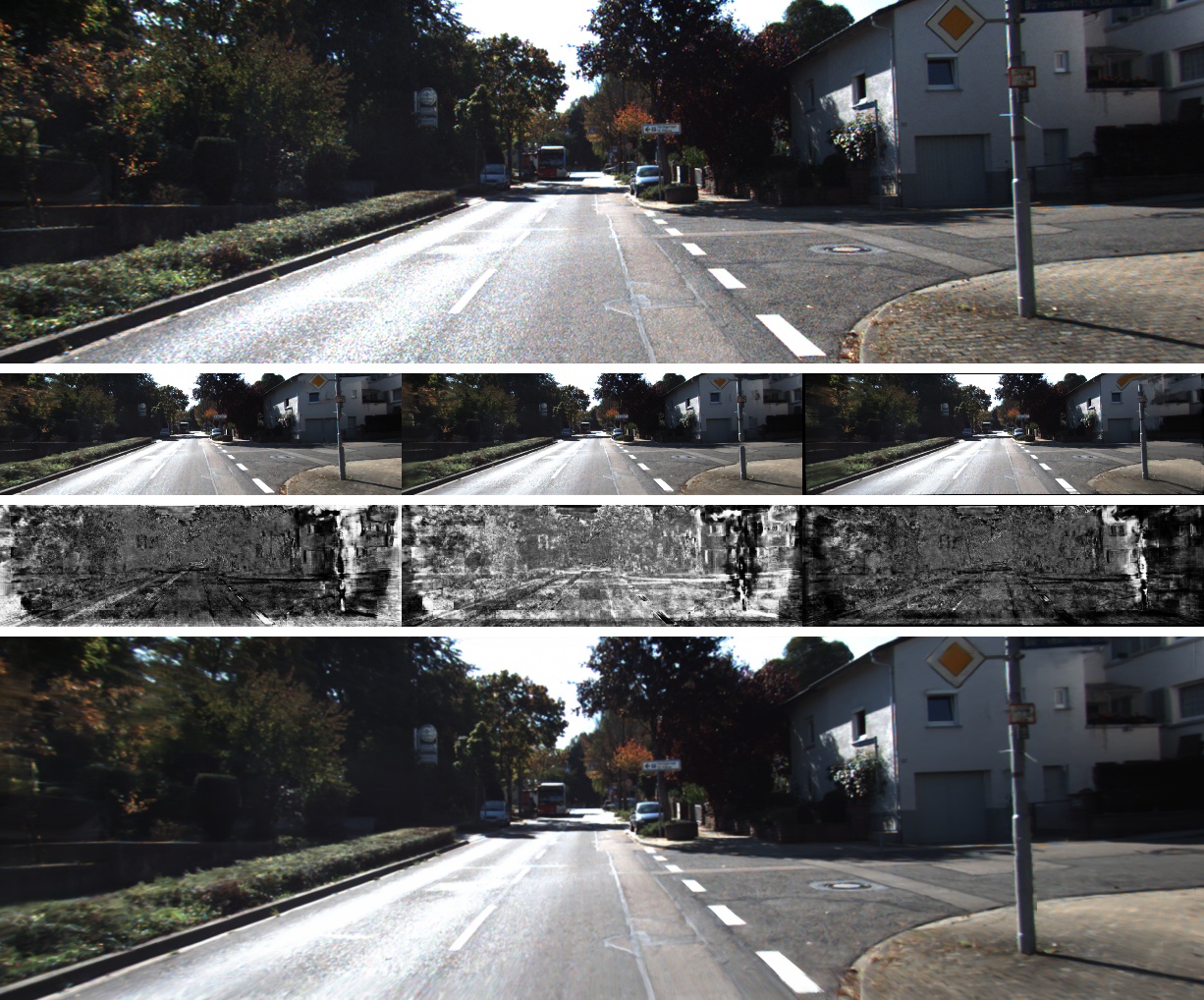}}~
{\includegraphics[scale=.165,trim={0 0 0 0},clip]{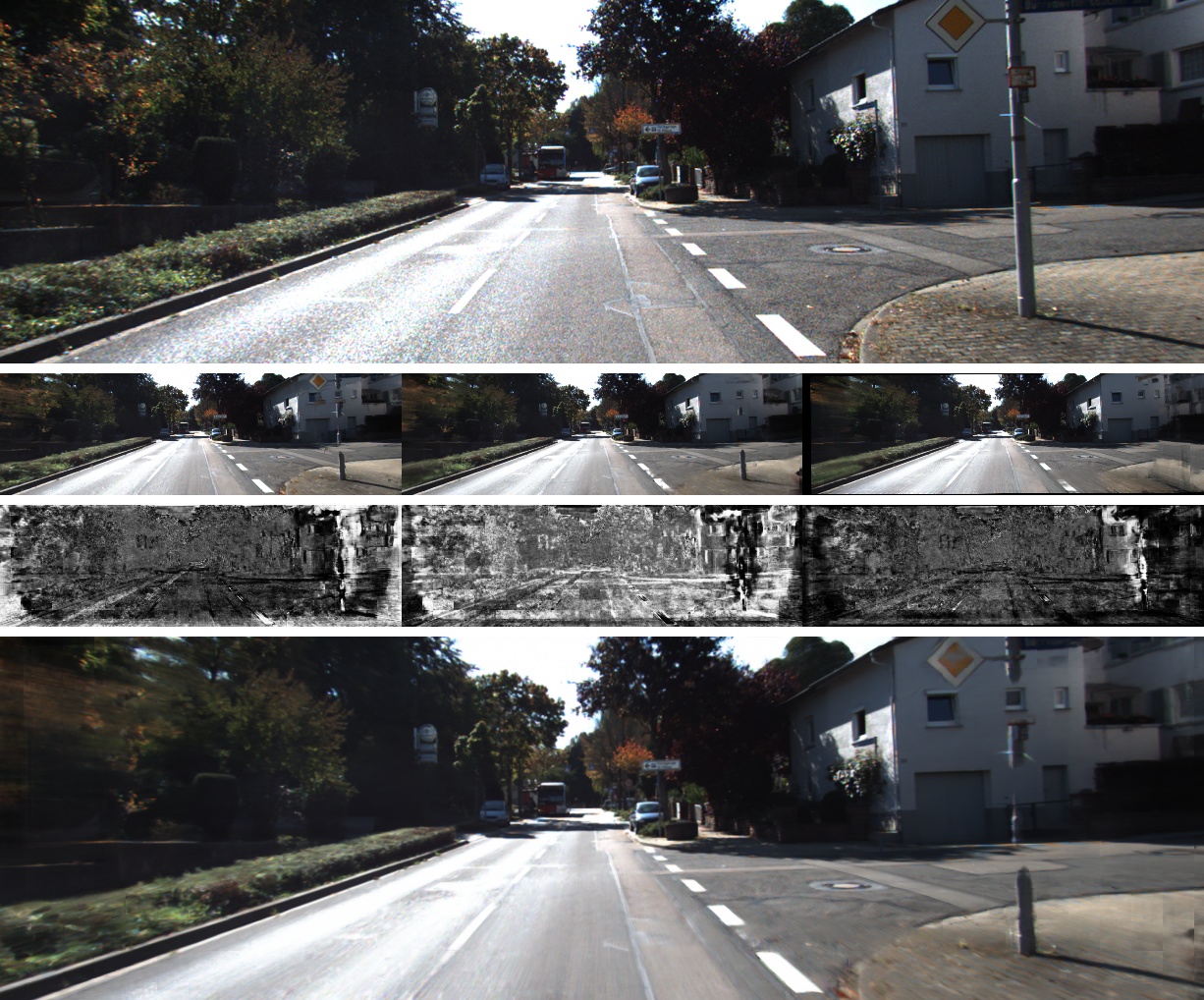}}
\caption{ Few more examples showing estimates similar to Fig.~\ref{fig:diff_baselines_1}. }
\label{fig:diff_baselines_2}
\end{figure*}

\begin{figure*}[h]
\centering
\textbf{\scriptsize 0.8m \hspace{5cm} \scriptsize 1.6m}\par\medskip
{\includegraphics[scale=.165,trim={0 0 0 0},clip]{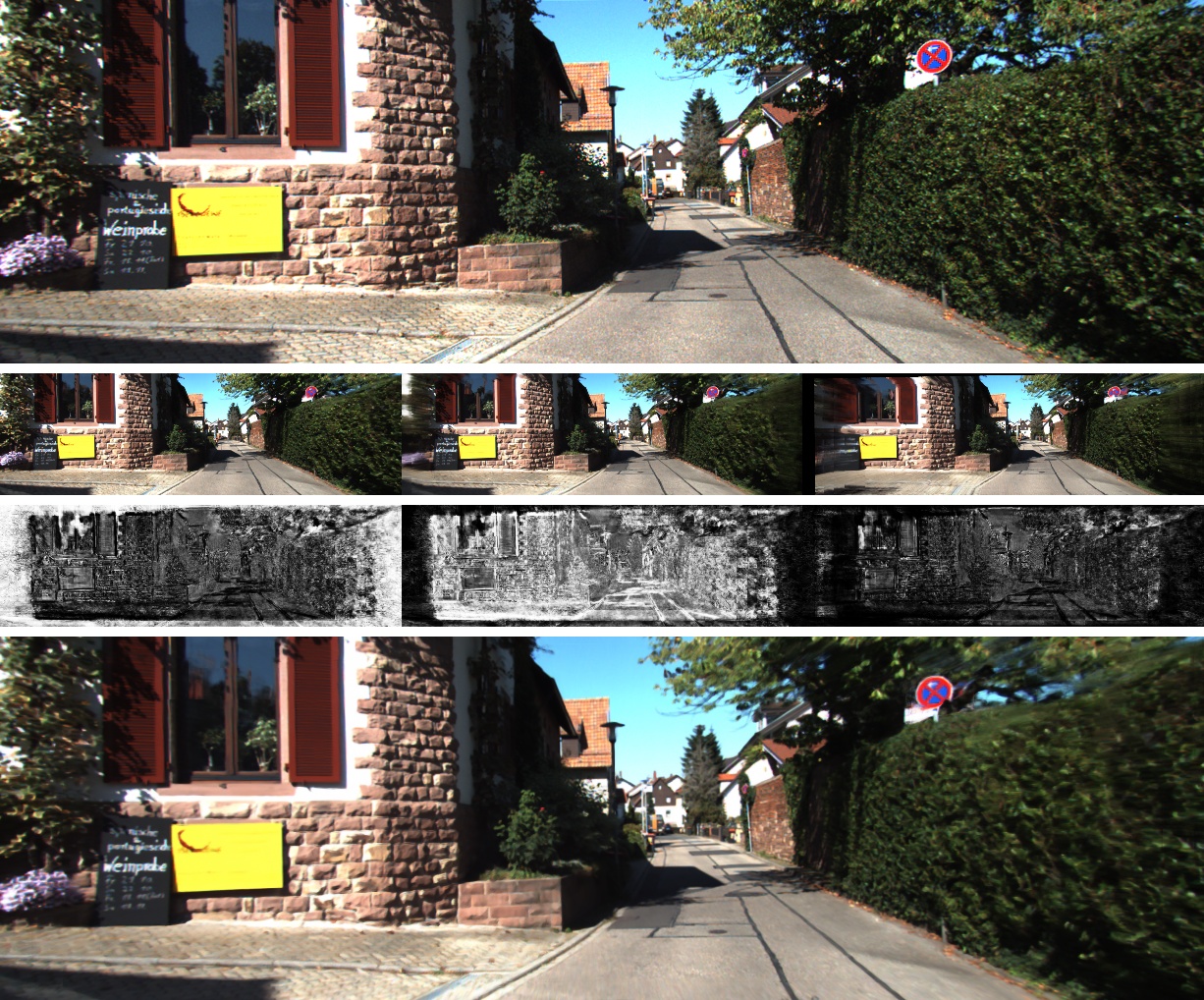}}~
{\includegraphics[scale=.165,trim={0 0 0 0},clip]{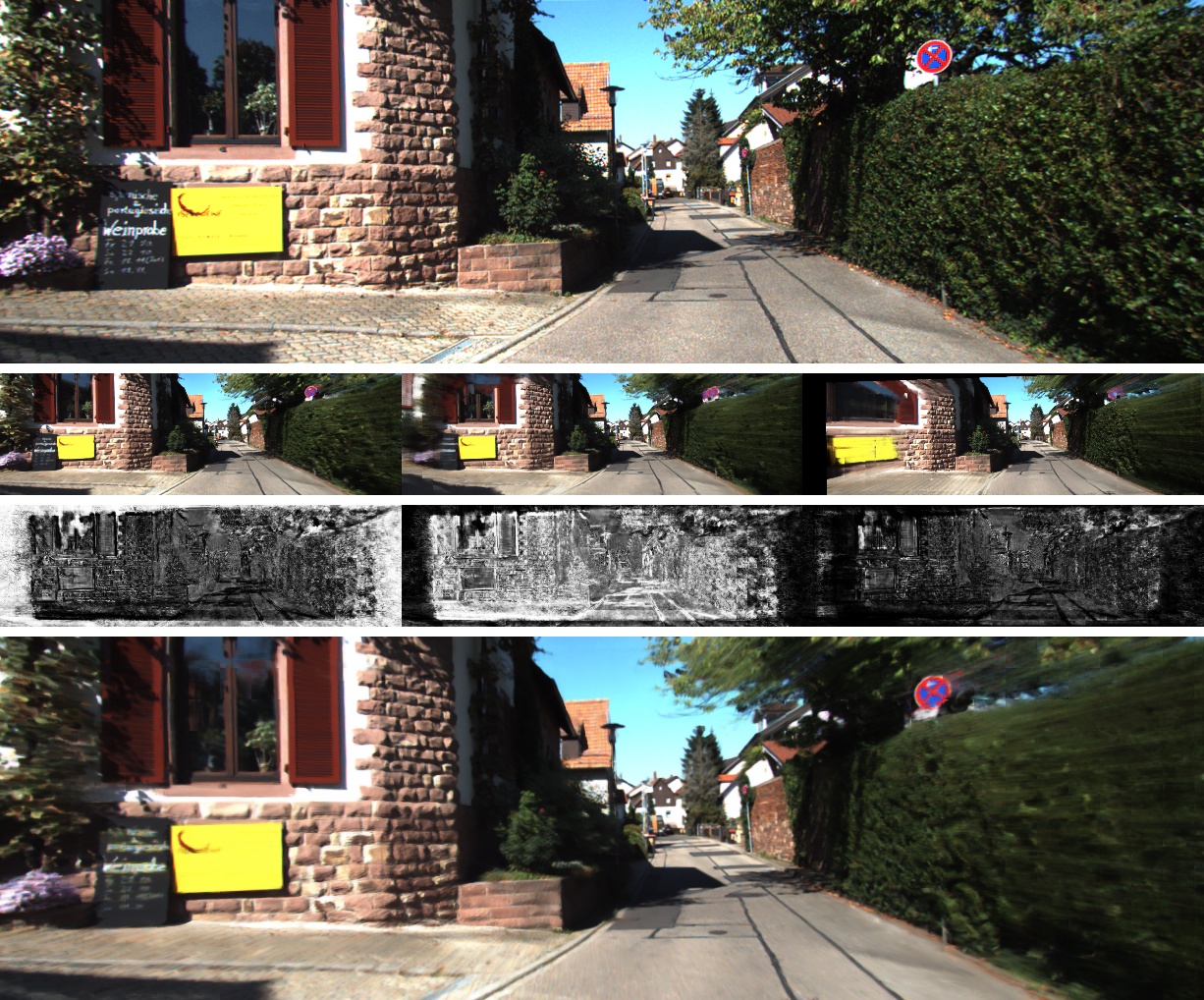}}\\
{\includegraphics[scale=.165,trim={0 0 0 0},clip]{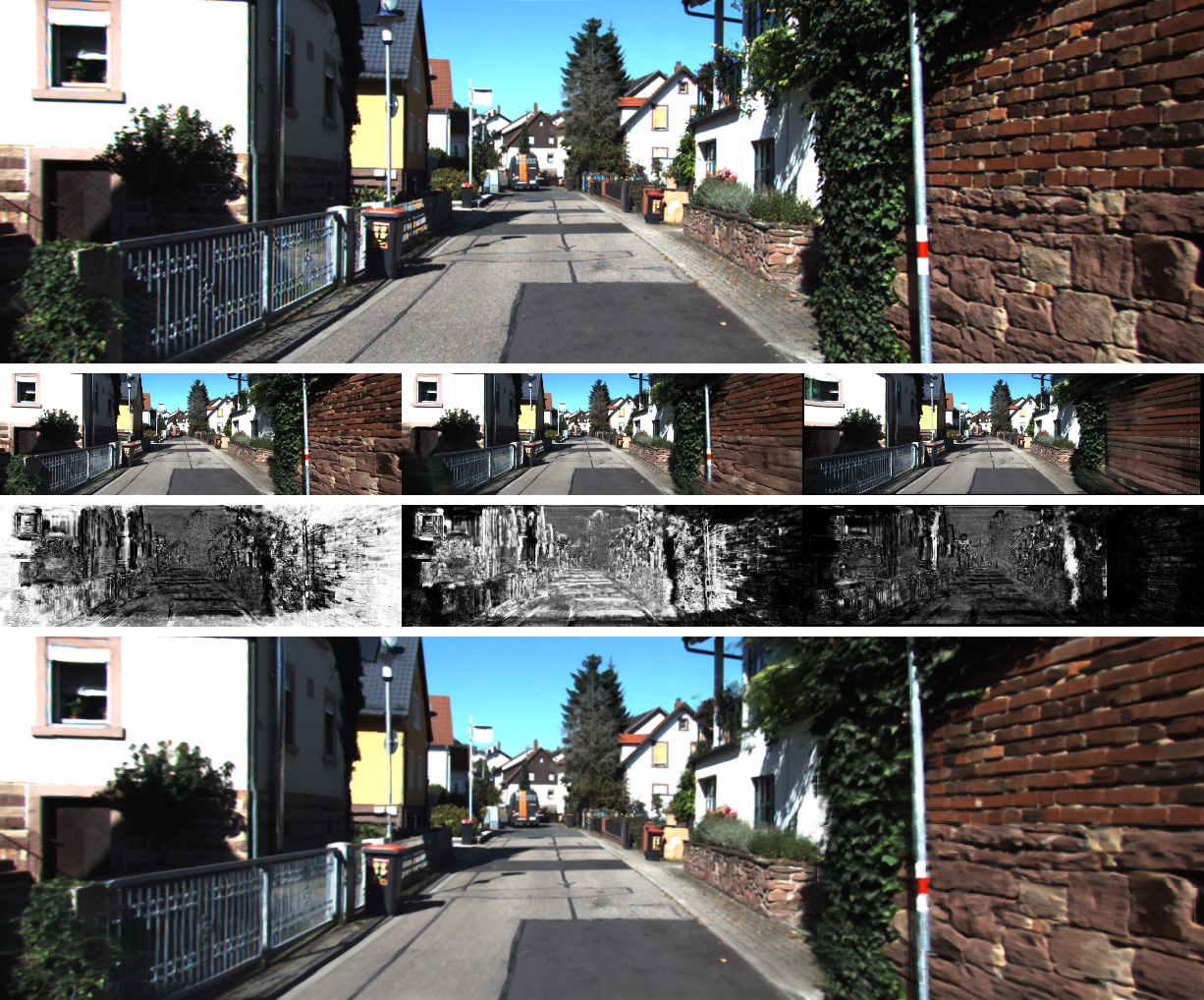}}~
{\includegraphics[scale=.165,trim={0 0 0 0},clip]{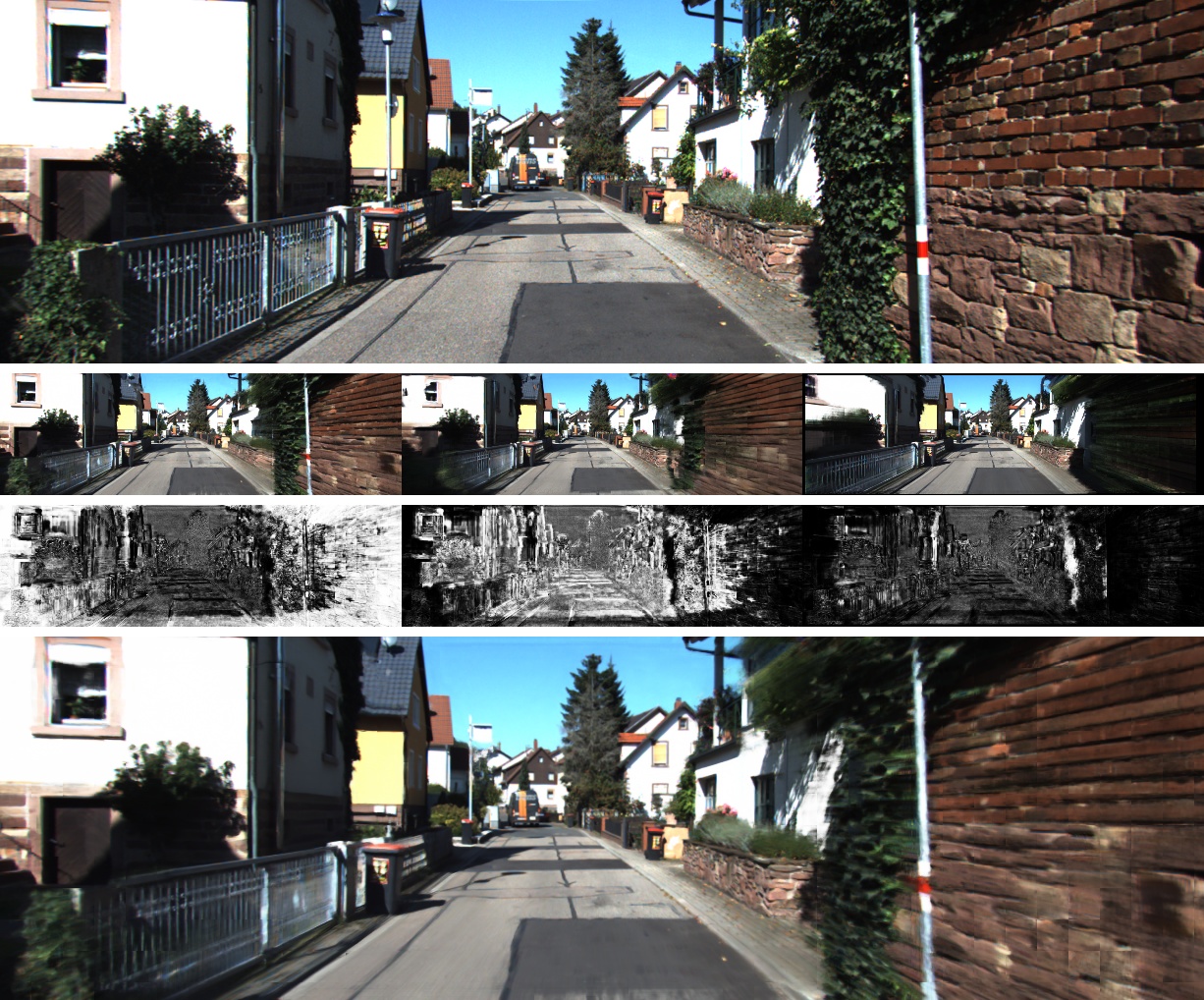}}\\
{\includegraphics[scale=.165,trim={0 0 0 0},clip]{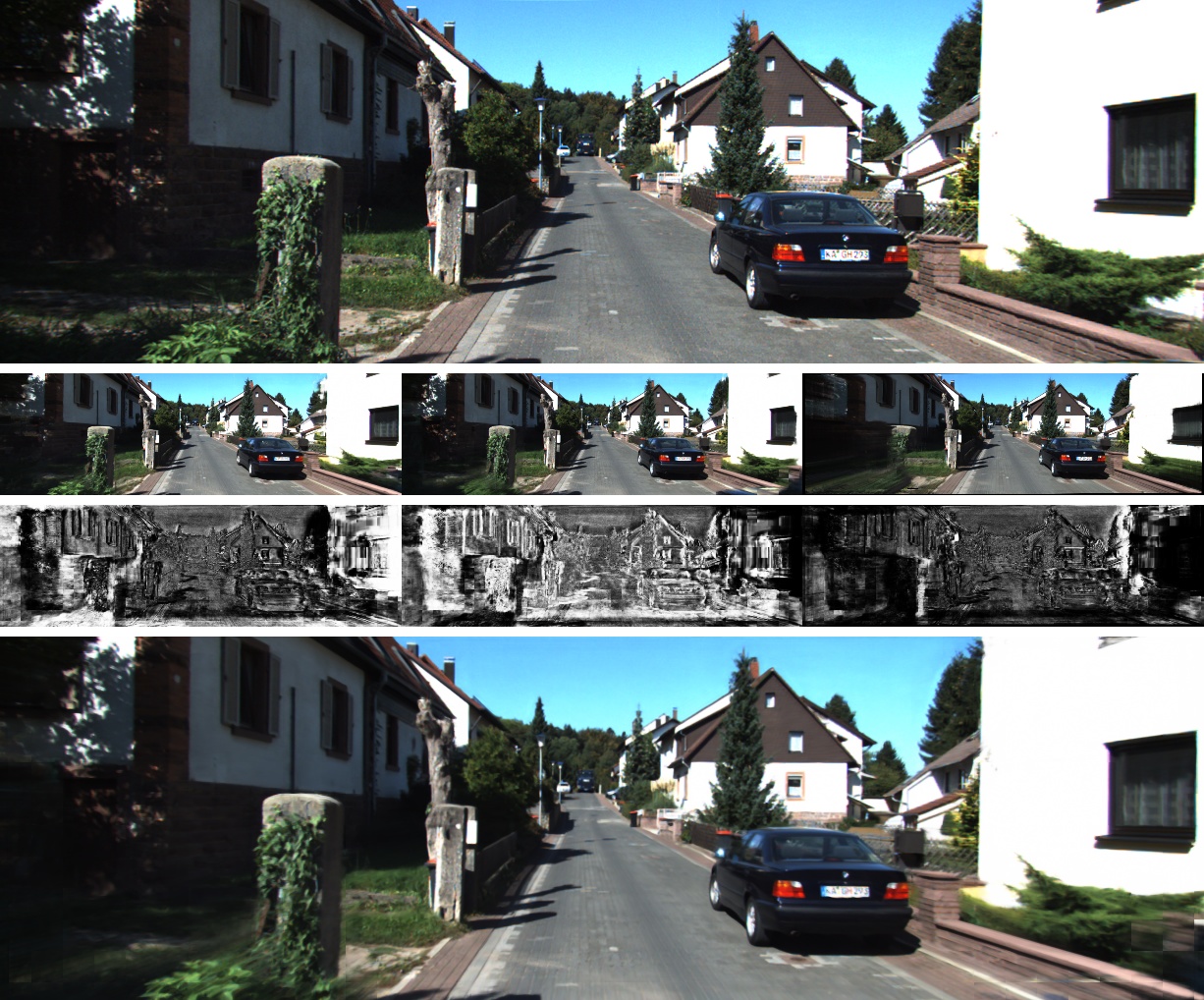}}~
{\includegraphics[scale=.165,trim={0 0 0 0},clip]{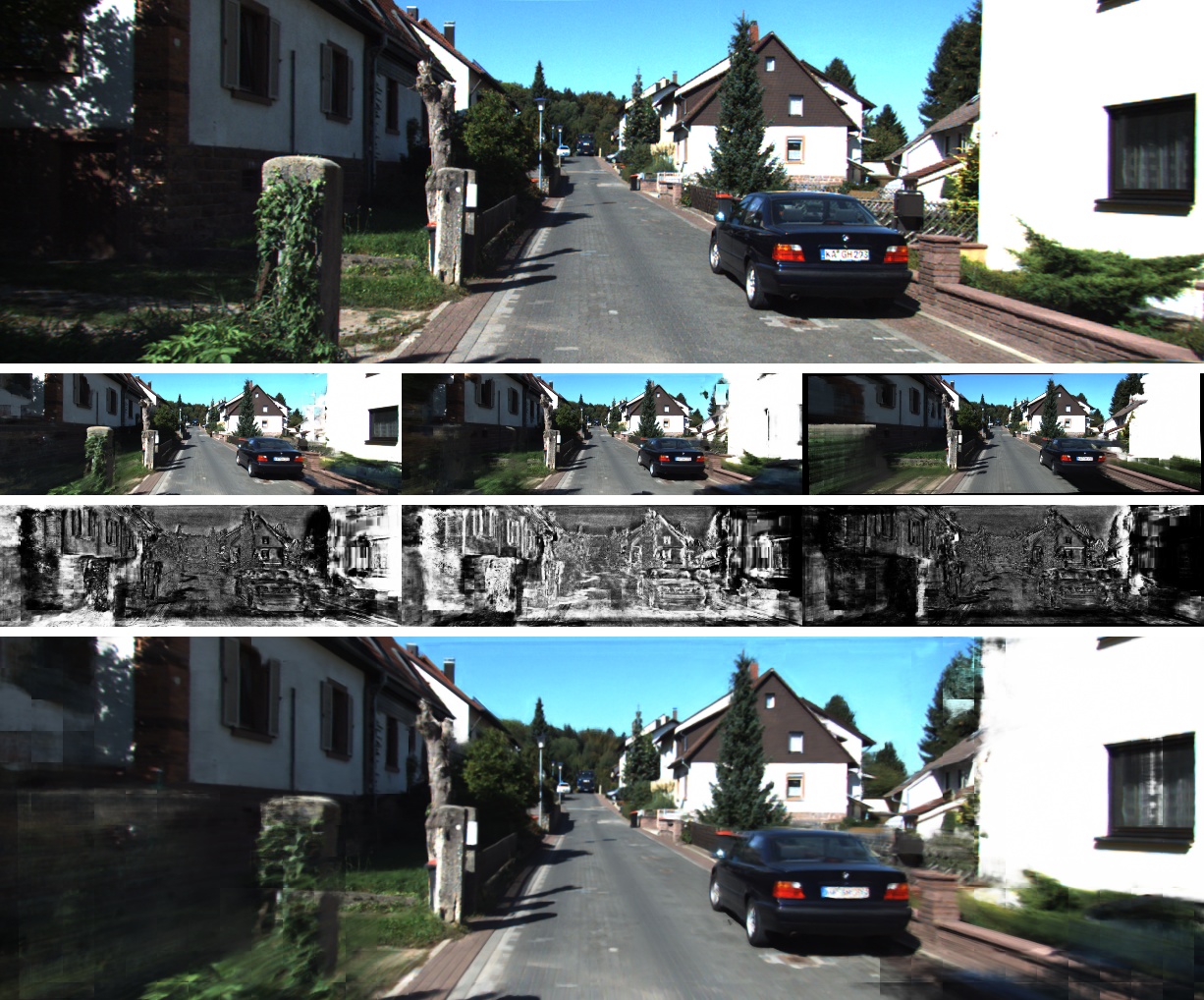}}
\caption{ Few more examples showing estimates similar to Fig.~\ref{fig:diff_baselines_1}. }
\label{fig:diff_baselines_3}
\end{figure*}

\textbf{Comparative analysis for different baselines}
We show additional results in Fig.~\ref{fig:diff_baselines_1}-\ref{fig:diff_baselines_3}. 
Each column in the figures show results using different sets of inputs with increasing baseline for comparative qualitative analysis. 
One can now analyze the final row in these figures to observe the effect of increasing baseline on view synthesis. 
Each figure also show estimated views using different input image pairs in second row and corresponding occlusion masks in third row. 
We show results with $64$ depth levels and using mr analysis. 

\subsection{Results on Spaces dataset}
\begin{figure*}[t]
\centering
\subfloat[]{\includegraphics[scale=0.25,clip]{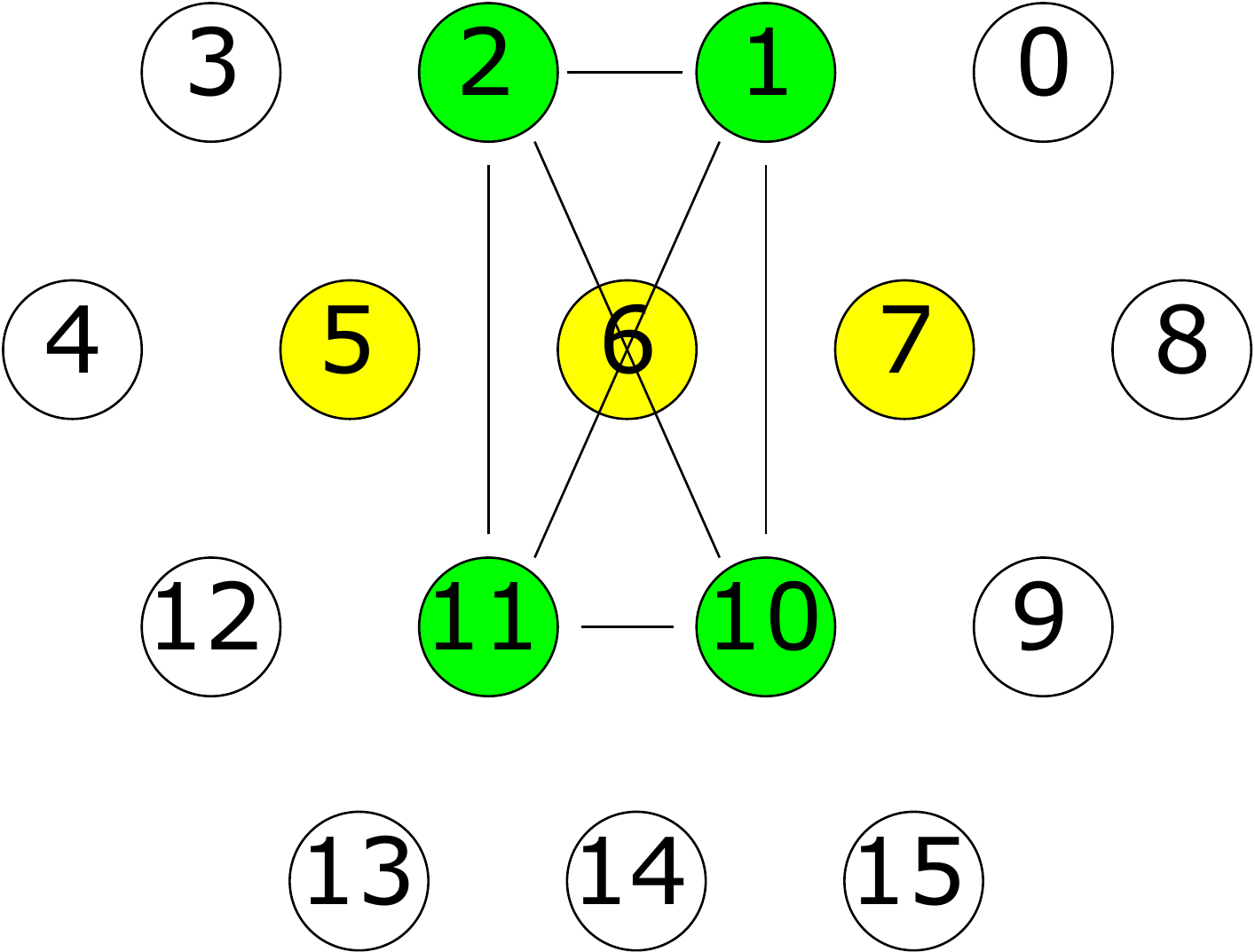}}~
\subfloat[]{\includegraphics[scale=0.25,clip]{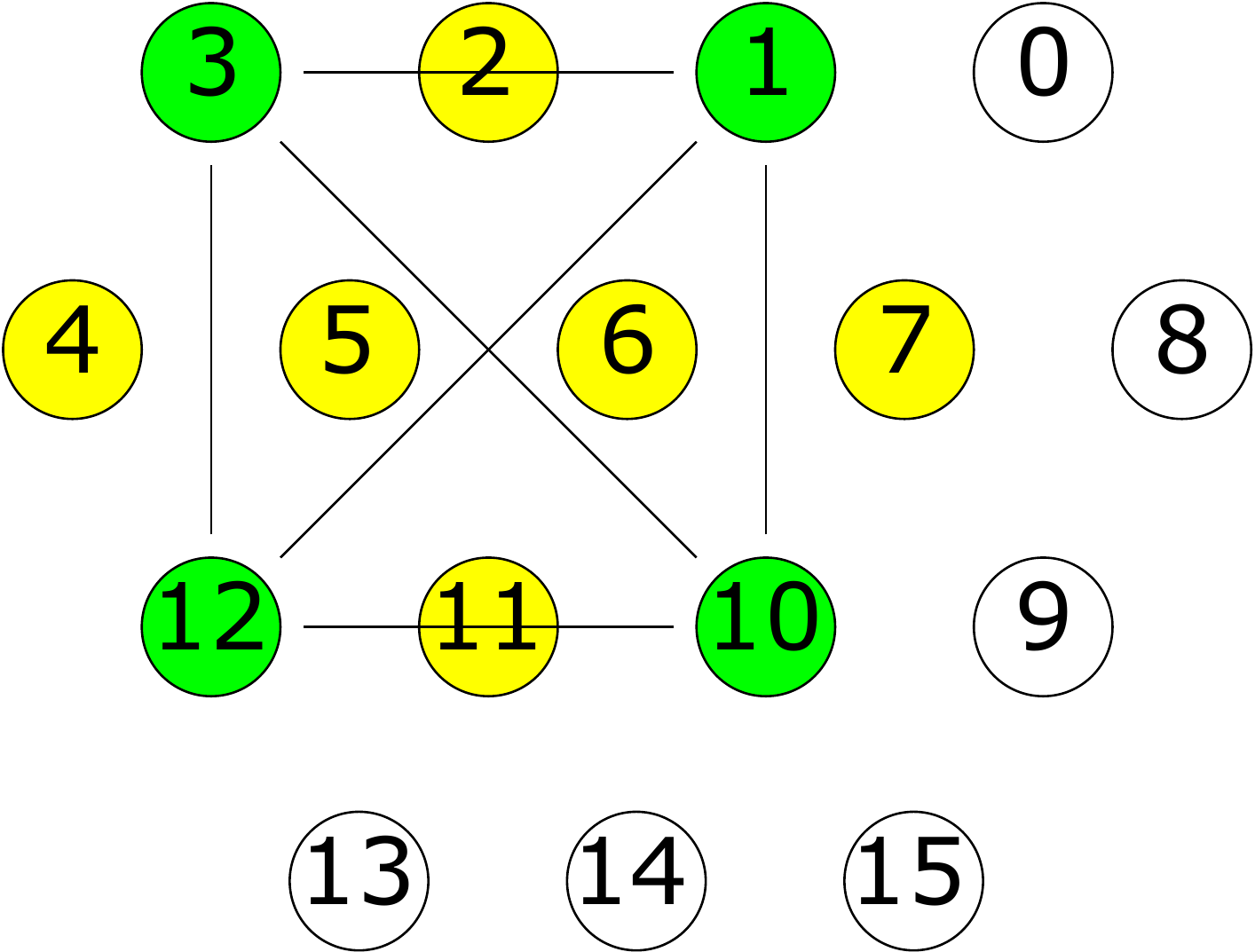}}~
\subfloat[]{\includegraphics[scale=0.25,clip]{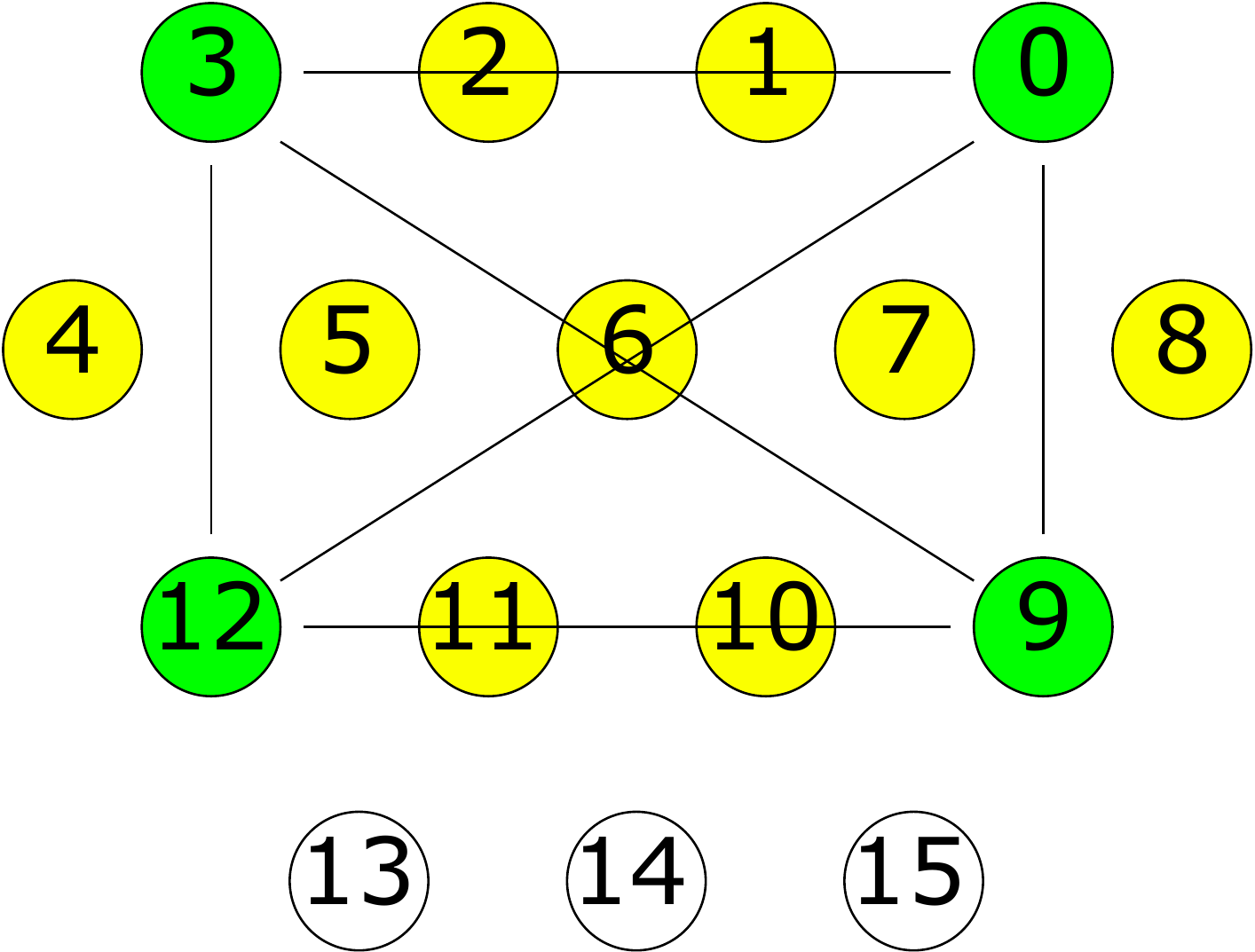}}~
\subfloat[]{\includegraphics[scale=0.25,clip]{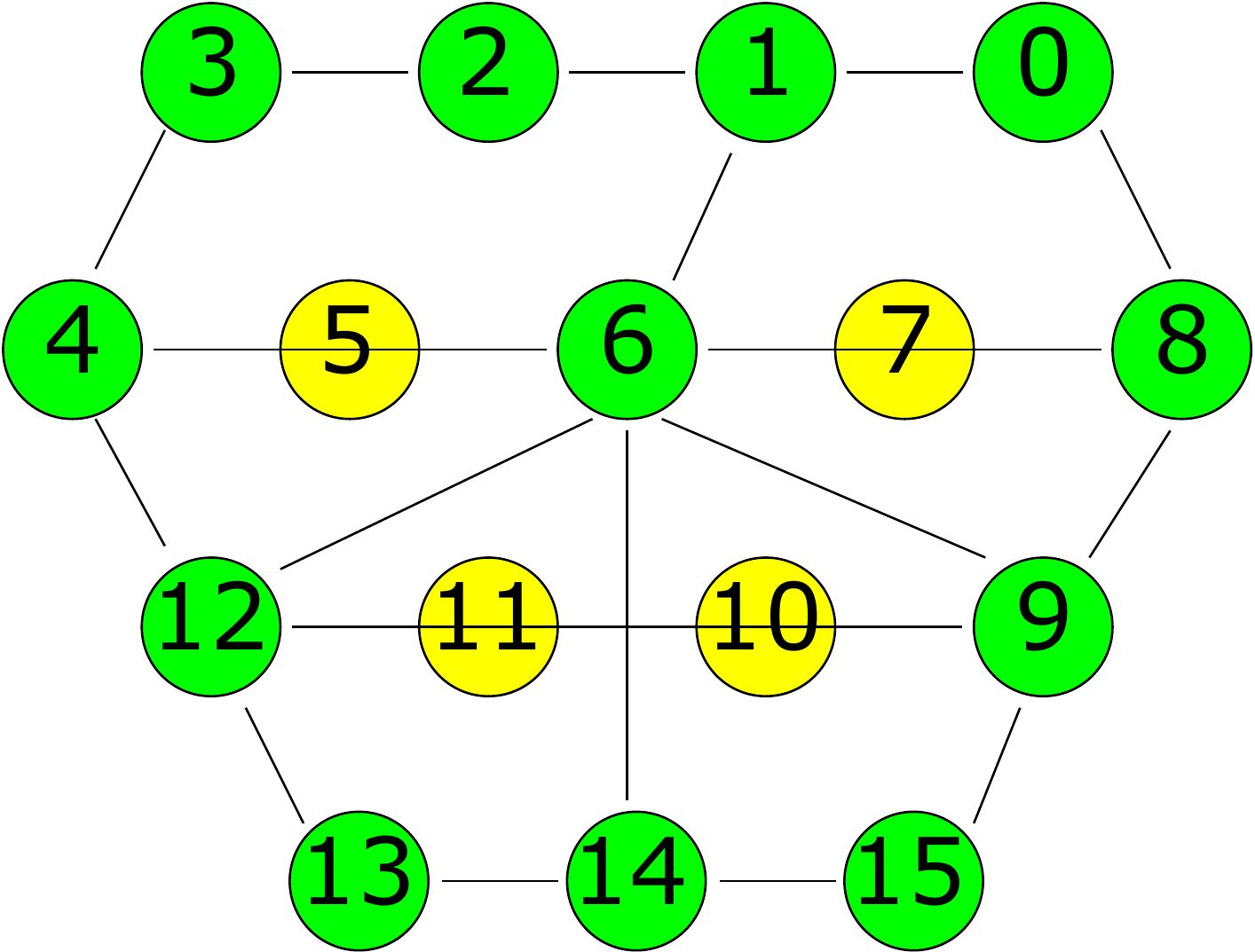}}
\caption{The figure shows the layout of the cameras in the Spaces dataset. We show details of all the experiments in this figure. Input image cameras are shown in green and an edge between cameras indicate a pair that we used for analysis. All the target views for respective experiments are shown in yellow. }
\label{fig:spaces_grid}
\end{figure*}

We show in Fig.~\ref{fig:spaces_grid} the layout of the light field cameras in Spaces dataset. 
The figure shows details of each experiment. 
The input images are shown in green and target images are shown in yellow. 
We show an edge in the figure indicating corresponding image pair used in the analysis. 

\begin{figure*}[t]
\centering
\textbf{\tiny Ground Truth \hspace{.4cm} \tiny 64 depths w/ mr \hspace{.4cm} \tiny 64 depths w/o mr \hspace{.4cm} \tiny 16 depths w/ mr \hspace{.4cm} \tiny 16 depths w/o mr \hspace{.4cm} \tiny 64 depths w/ mr err\hspace{.4cm} \tiny 64 depths w/o mr err\hspace{.4cm} \tiny 16 depths w/ mr err\hspace{.4cm} \tiny 16 depths w/o mr err}\par\medskip
{\includegraphics[scale=0.06,clip]{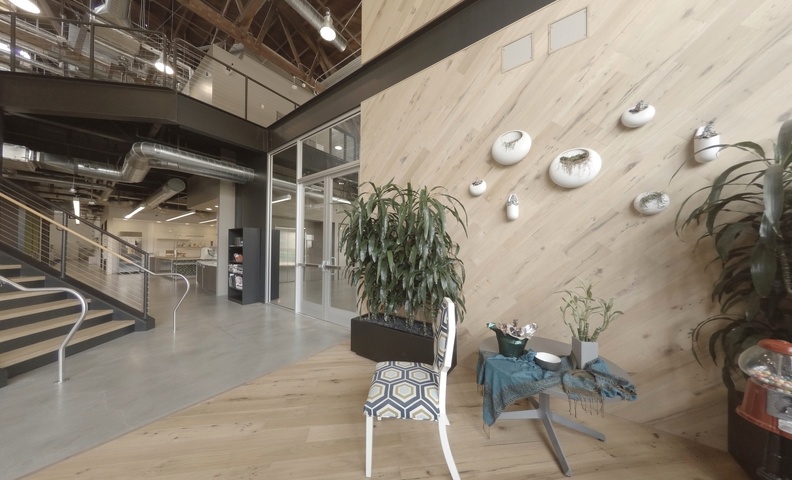}}
{\includegraphics[scale=0.06,clip]{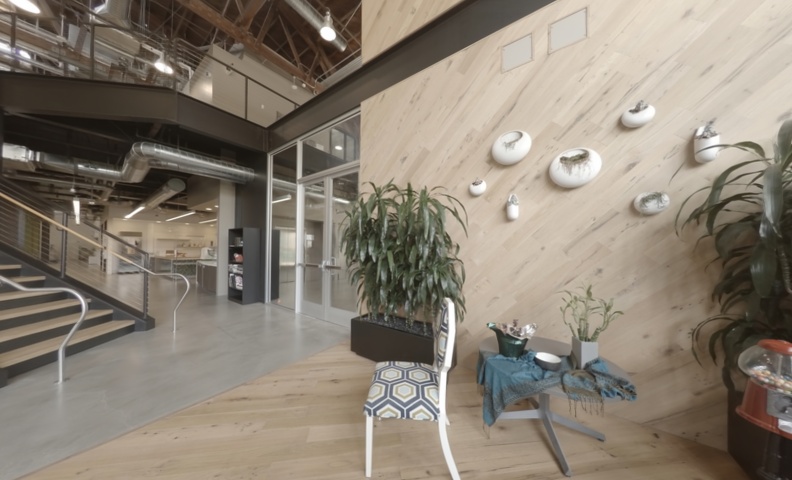}}
{\includegraphics[scale=0.06,clip]{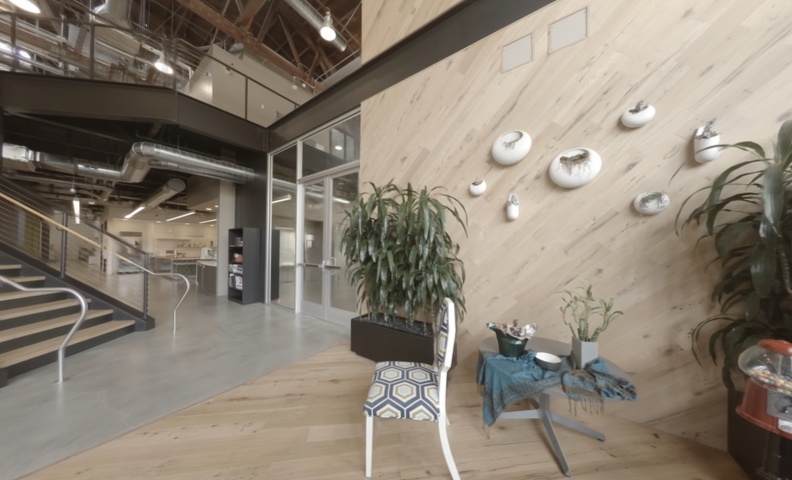}}
{\includegraphics[scale=0.06,clip]{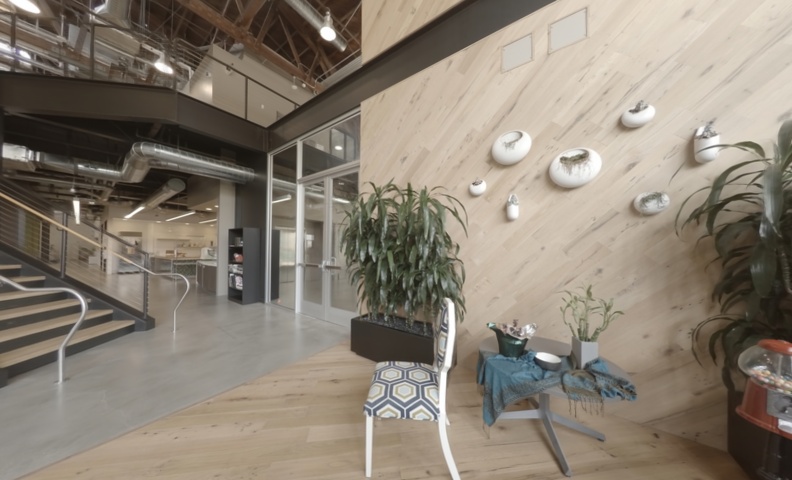}}
{\includegraphics[scale=0.06,clip]{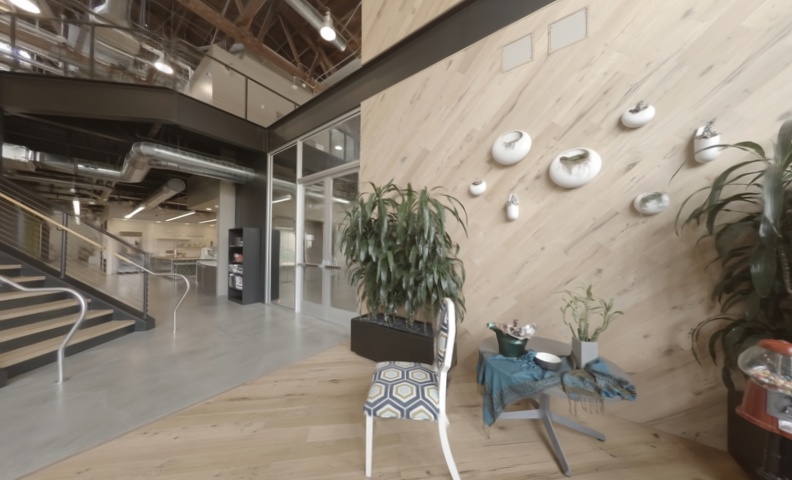}}
{\includegraphics[scale=0.06,clip]{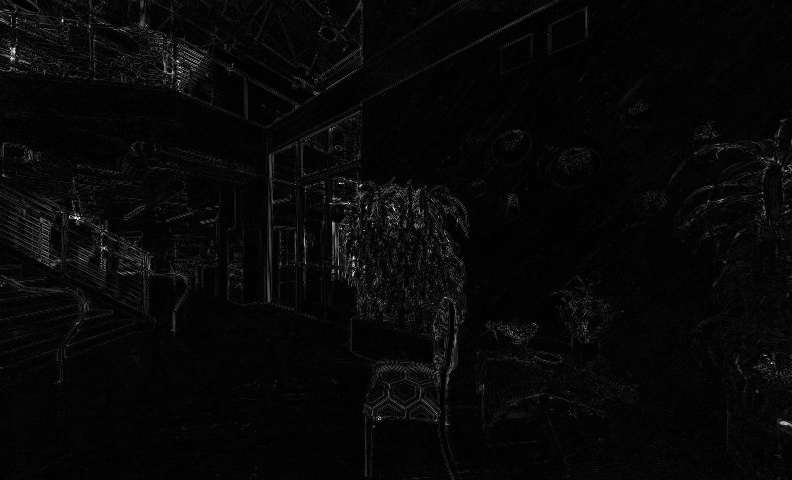}}
{\includegraphics[scale=0.06,clip]{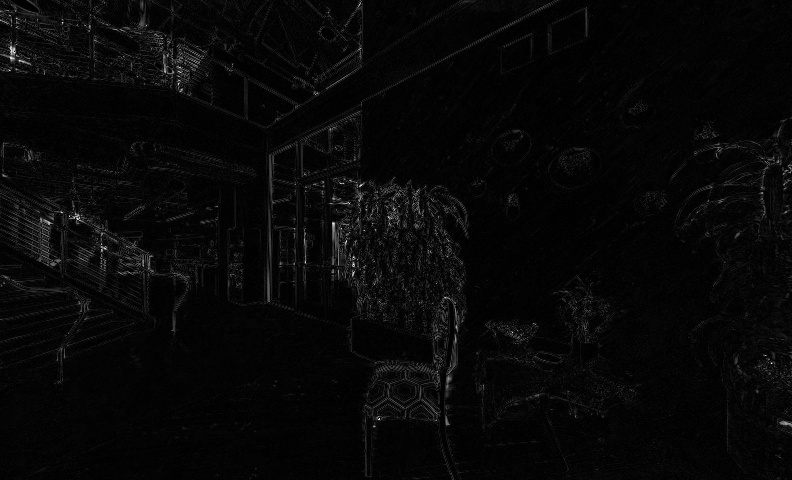}}
{\includegraphics[scale=0.06,clip]{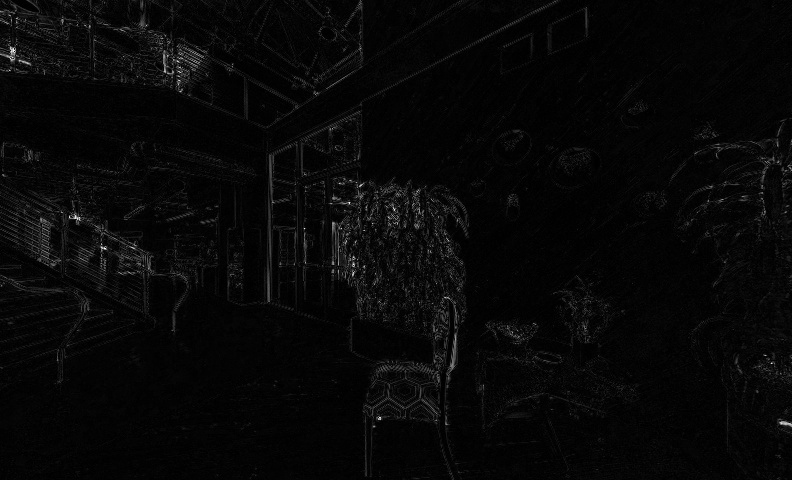}}
{\includegraphics[scale=0.06,clip]{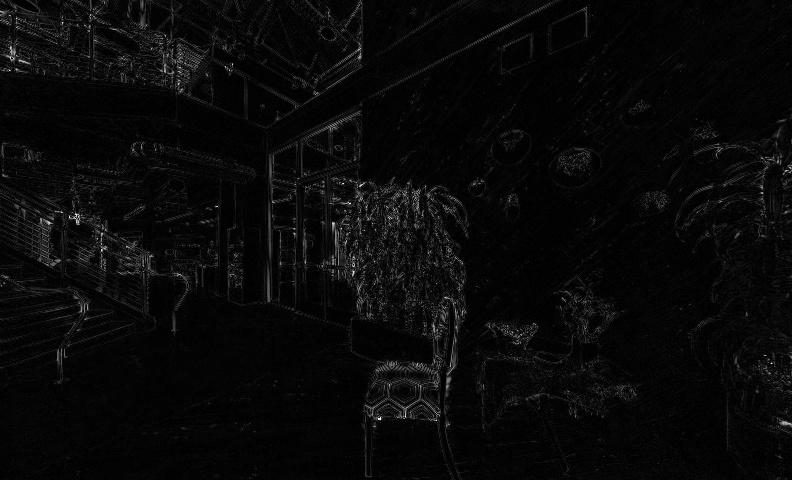}}\\

{\includegraphics[scale=0.06,clip]{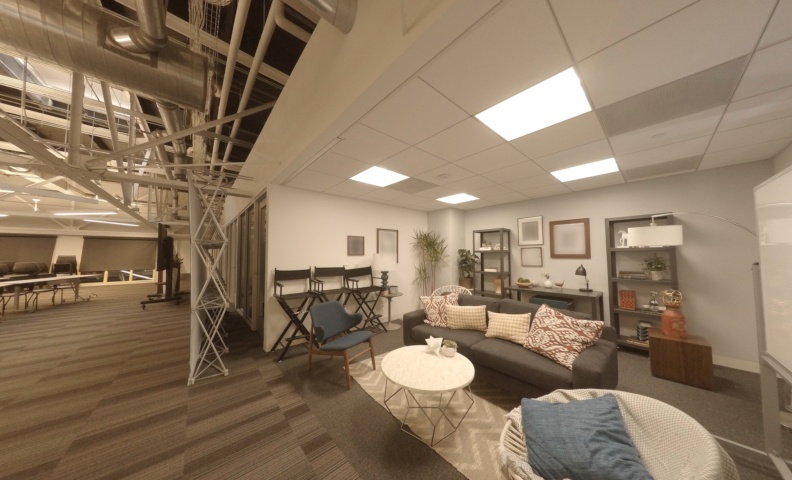}}
{\includegraphics[scale=0.06,clip]{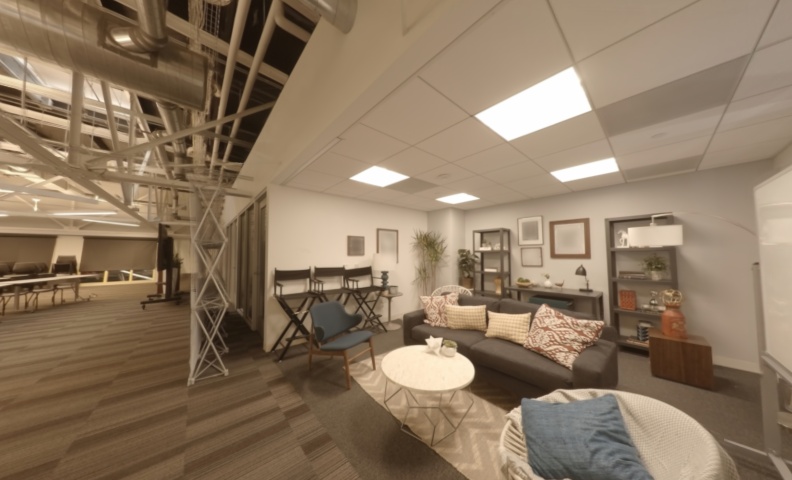}}
{\includegraphics[scale=0.06,clip]{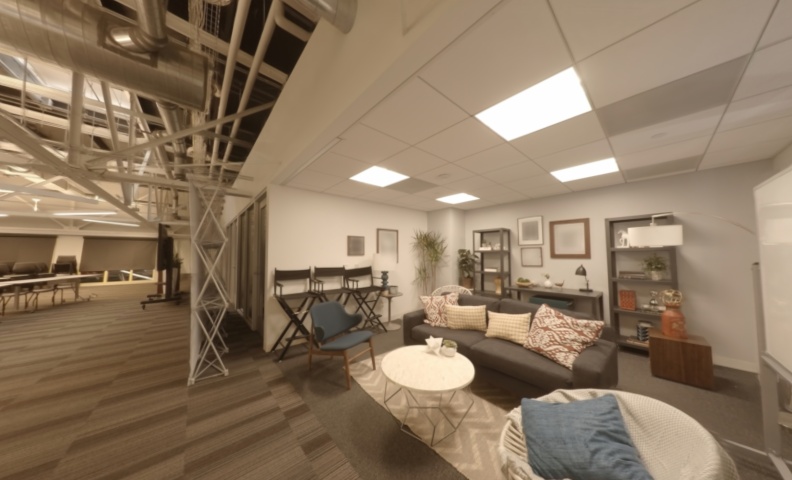}}
{\includegraphics[scale=0.06,clip]{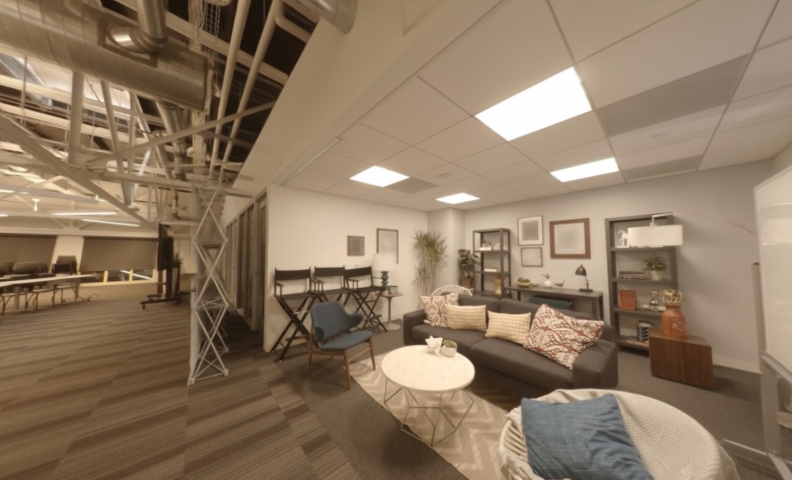}}
{\includegraphics[scale=0.06,clip]{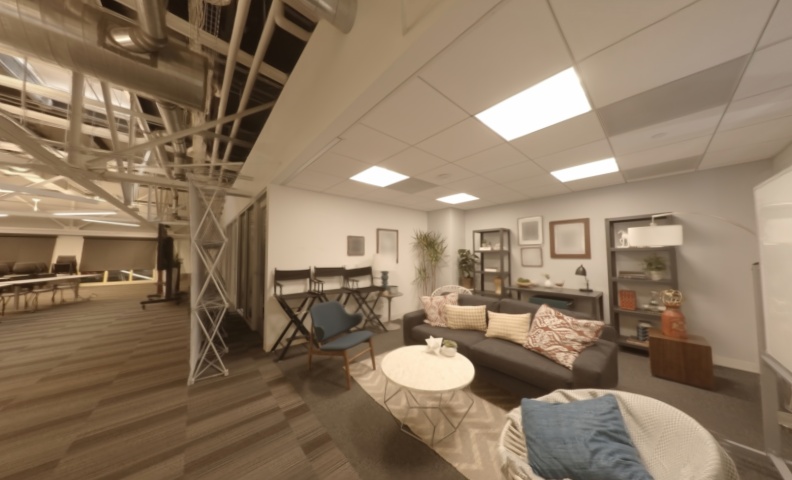}}
{\includegraphics[scale=0.06,clip]{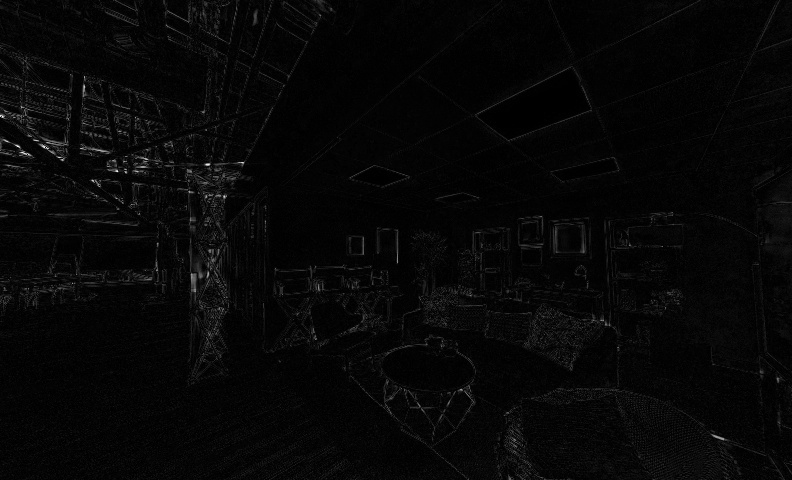}}
{\includegraphics[scale=0.06,clip]{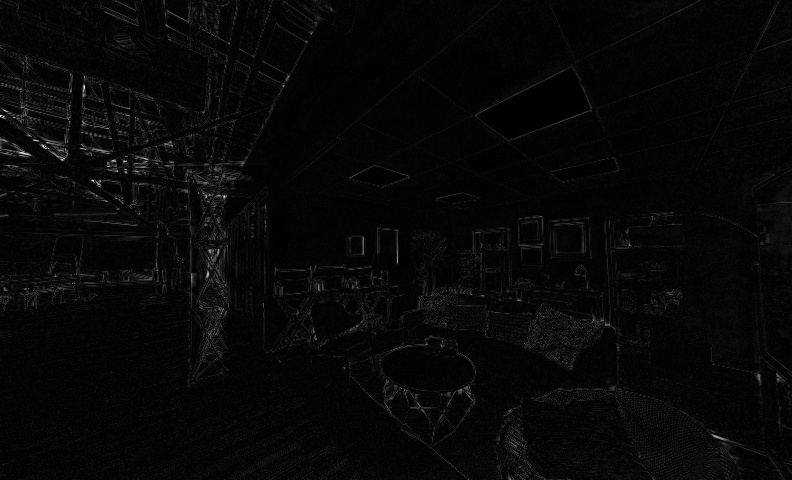}}
{\includegraphics[scale=0.06,clip]{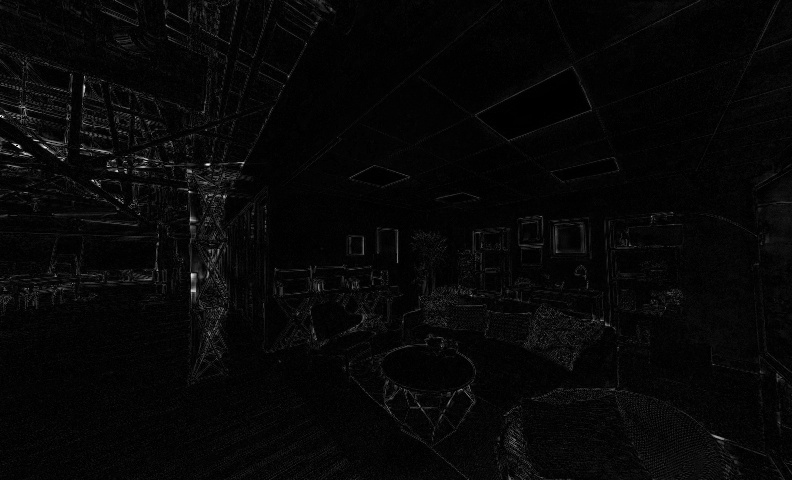}}
{\includegraphics[scale=0.06,clip]{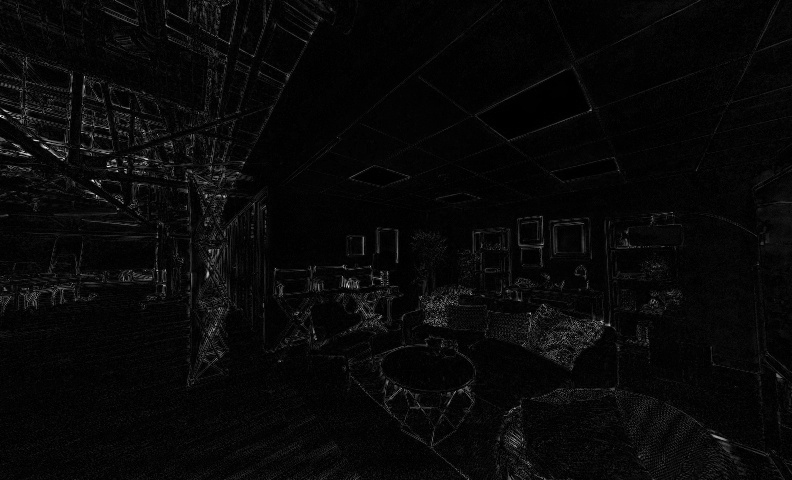}}

{\includegraphics[scale=0.06,clip]{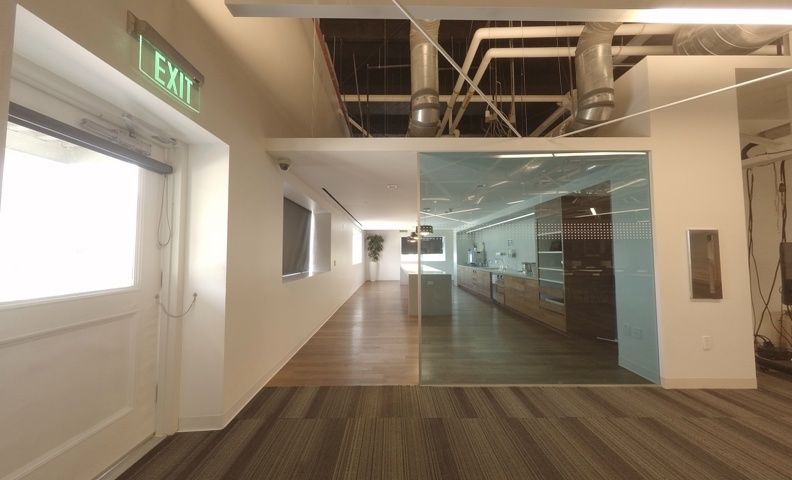}}
{\includegraphics[scale=0.06,clip]{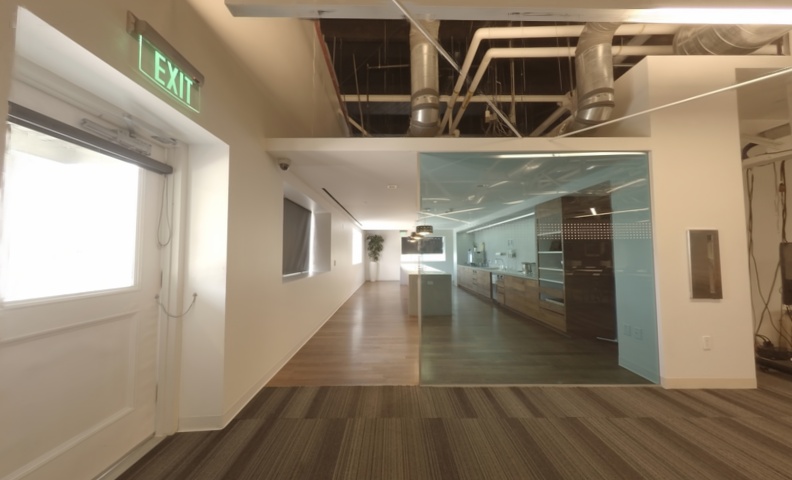}}
{\includegraphics[scale=0.06,clip]{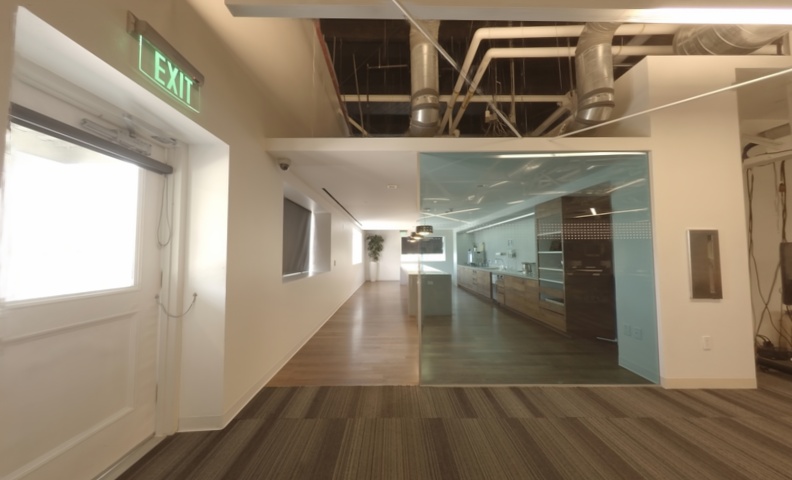}}
{\includegraphics[scale=0.06,clip]{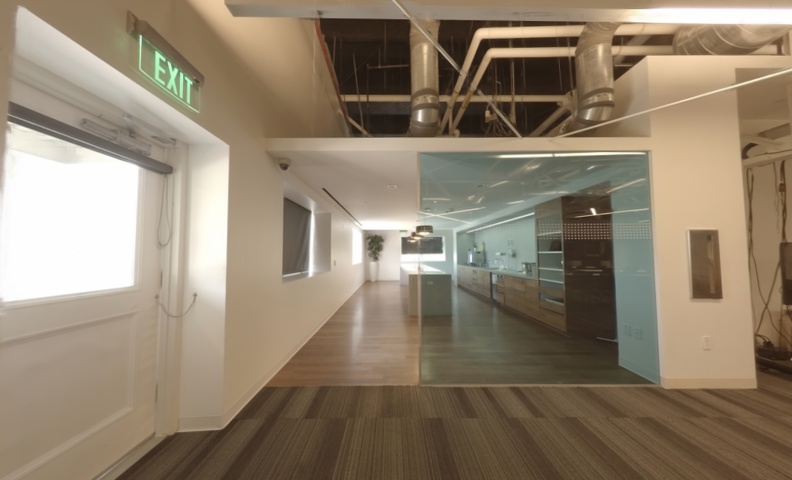}}
{\includegraphics[scale=0.06,clip]{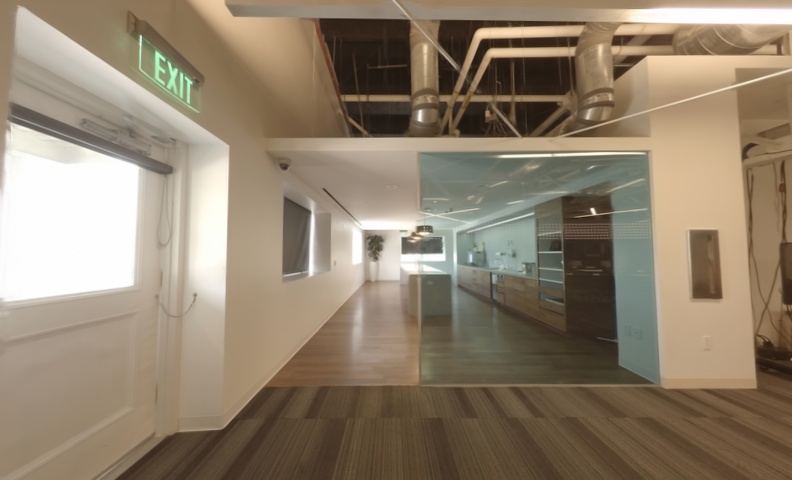}}
{\includegraphics[scale=0.06,clip]{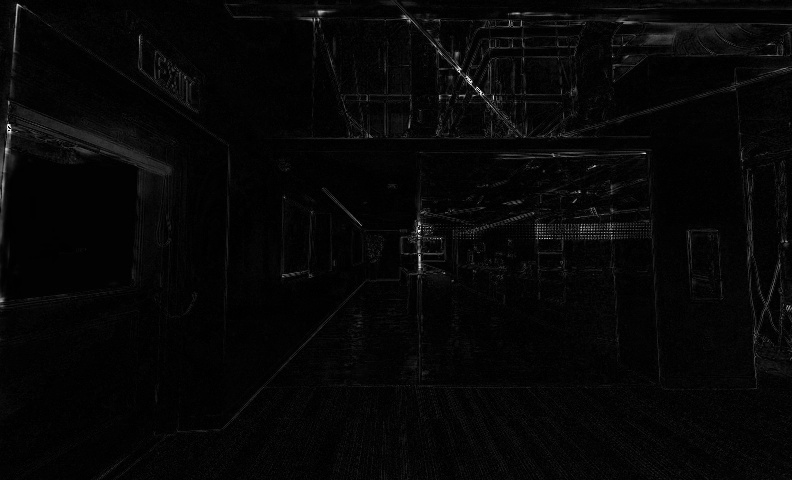}}
{\includegraphics[scale=0.06,clip]{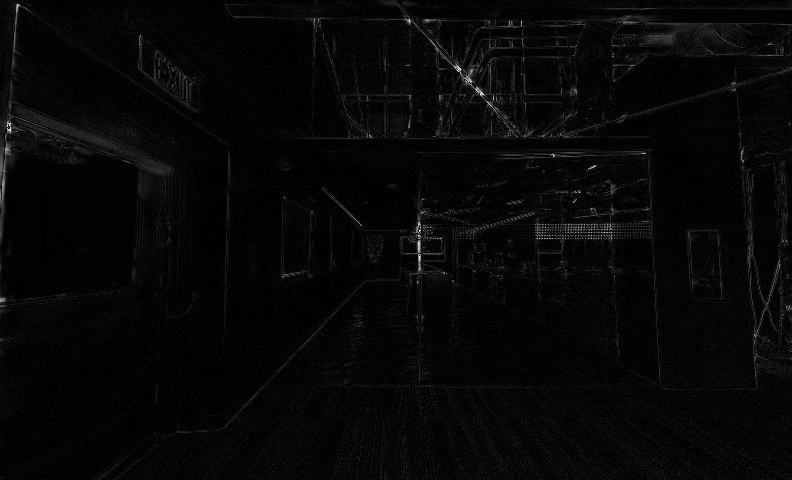}}
{\includegraphics[scale=0.06,clip]{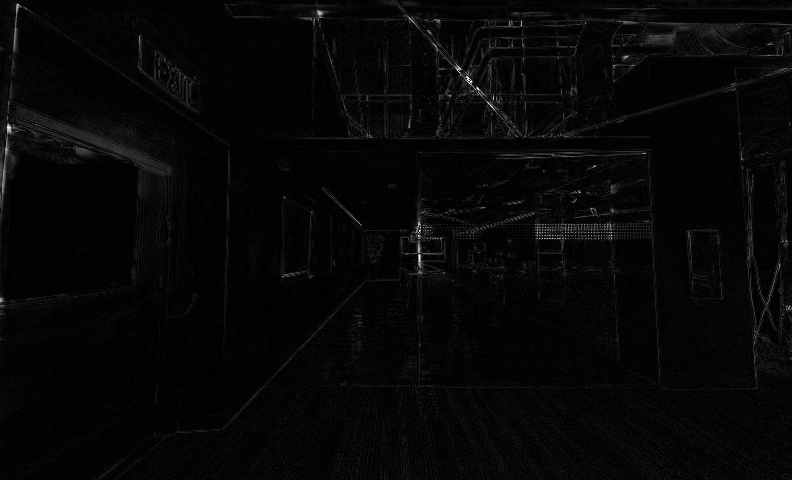}}
{\includegraphics[scale=0.06,clip]{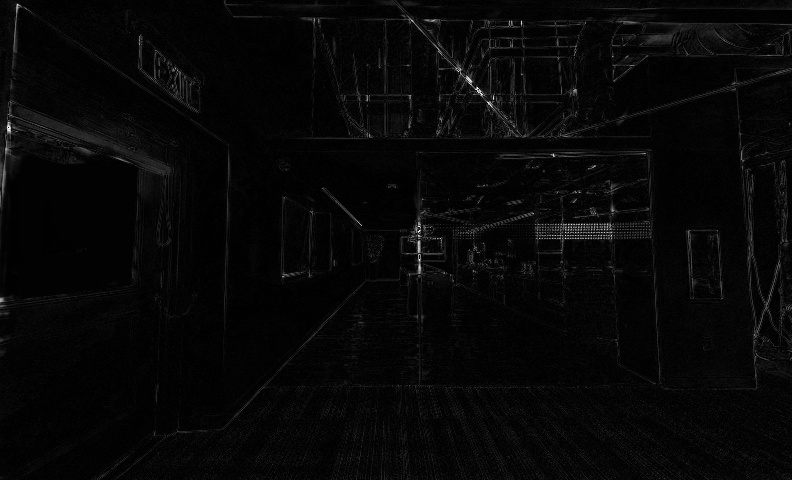}}

{\includegraphics[scale=0.06,clip]{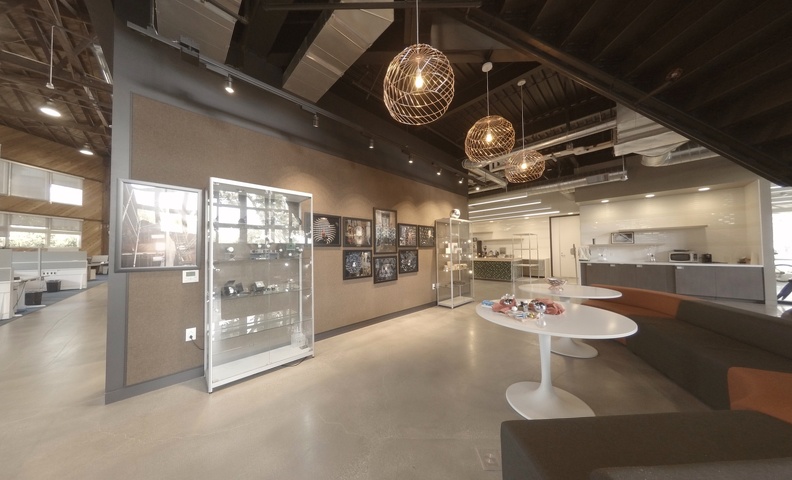}}
{\includegraphics[scale=0.06,clip]{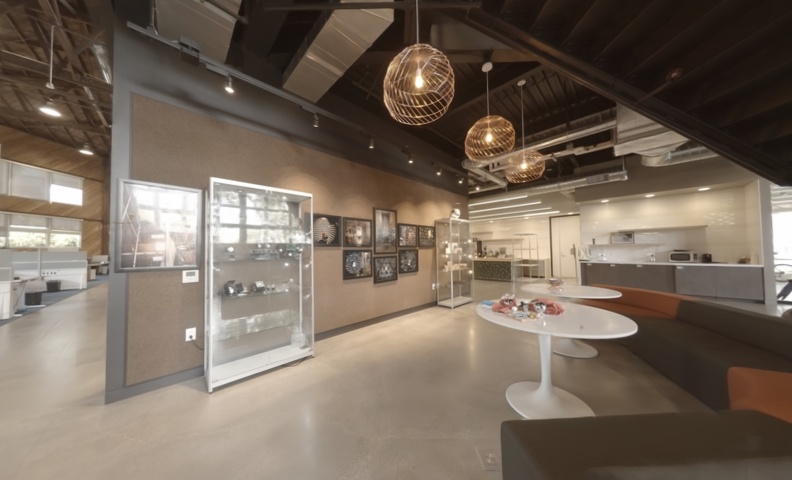}}
{\includegraphics[scale=0.06,clip]{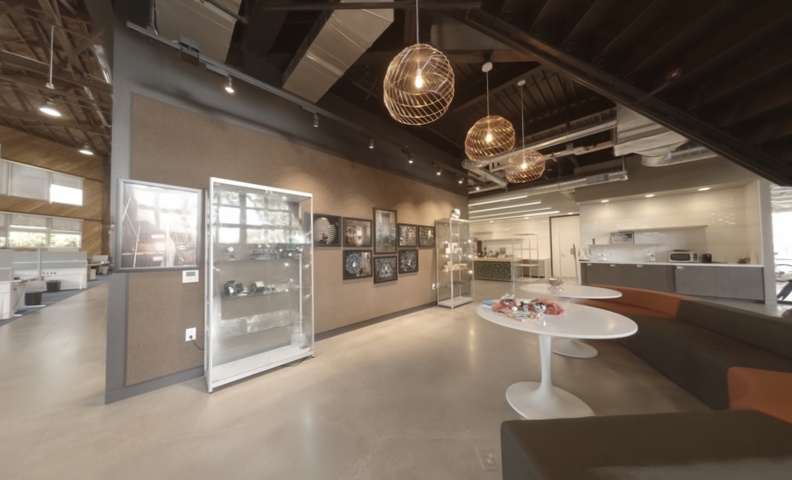}}
{\includegraphics[scale=0.06,clip]{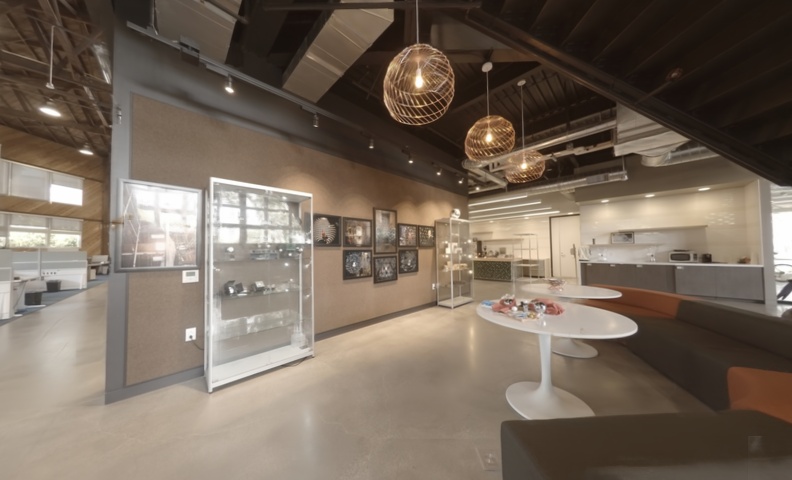}}
{\includegraphics[scale=0.06,clip]{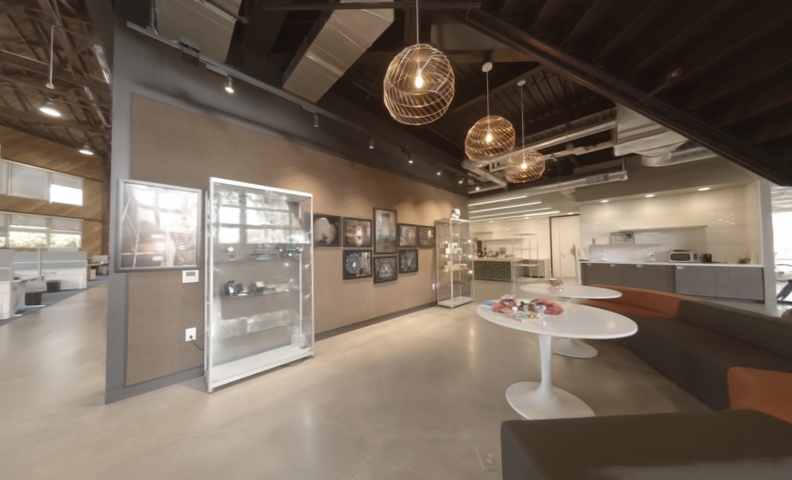}}
{\includegraphics[scale=0.06,clip]{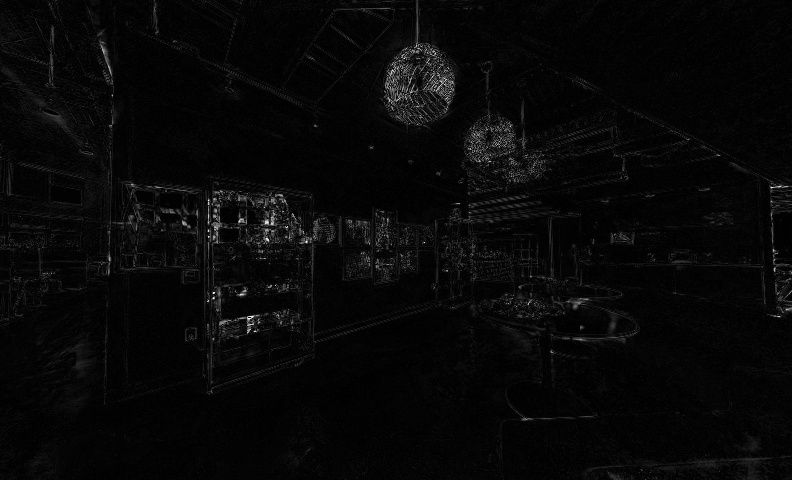}}
{\includegraphics[scale=0.06,clip]{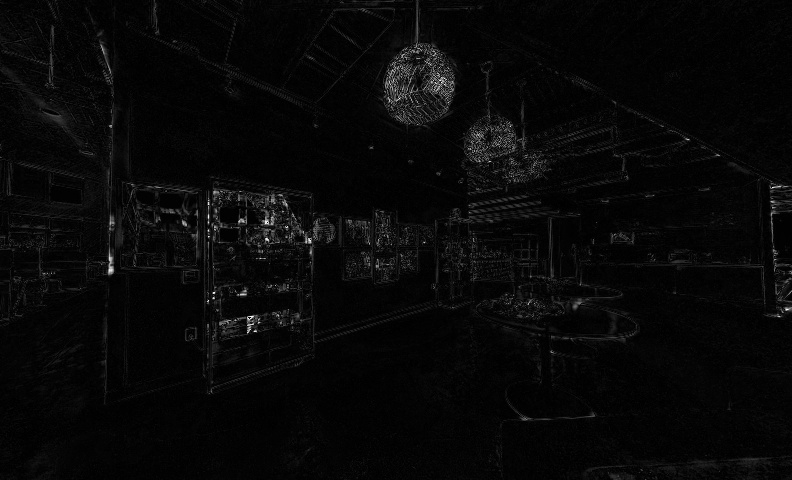}}
{\includegraphics[scale=0.06,clip]{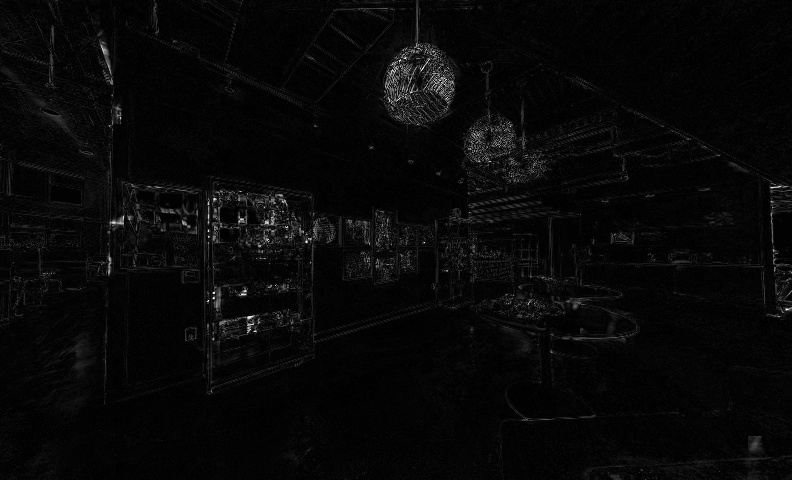}}
{\includegraphics[scale=0.06,clip]{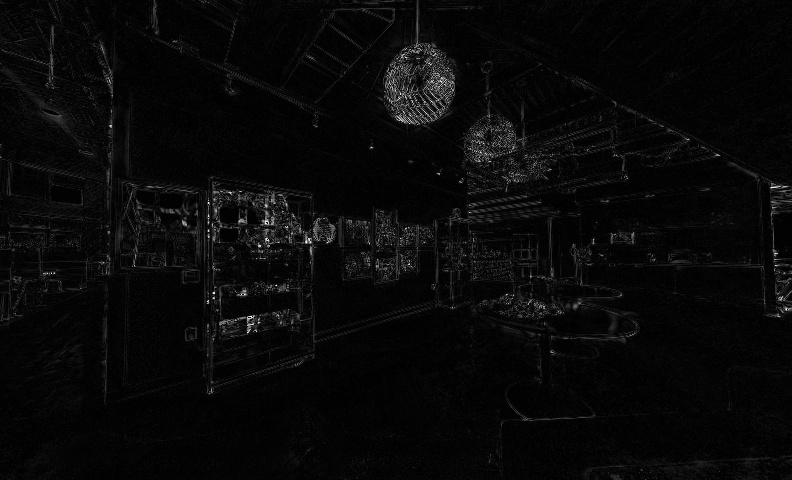}}

{\includegraphics[scale=0.06,clip]{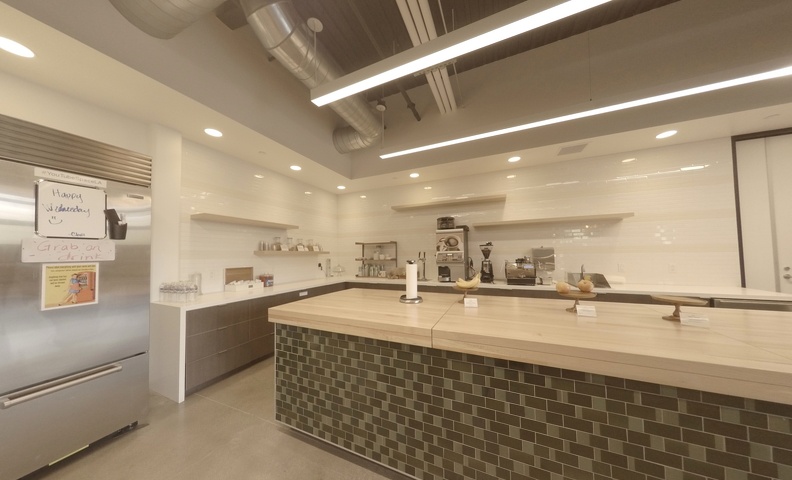}}
{\includegraphics[scale=0.06,clip]{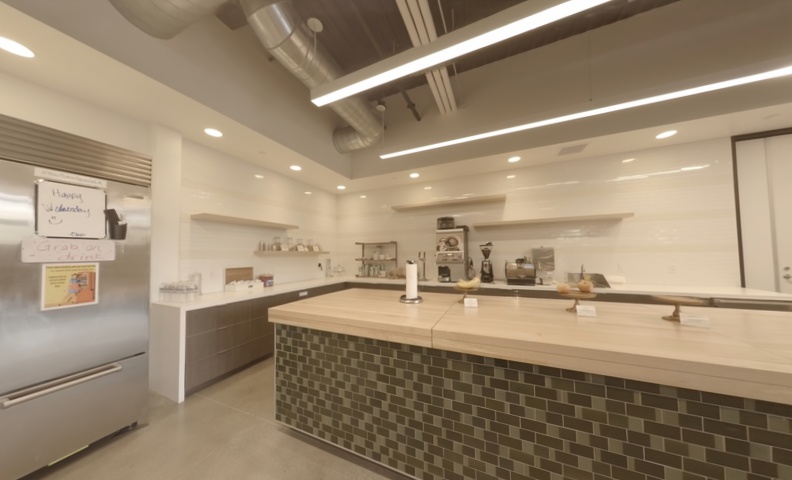}}
{\includegraphics[scale=0.06,clip]{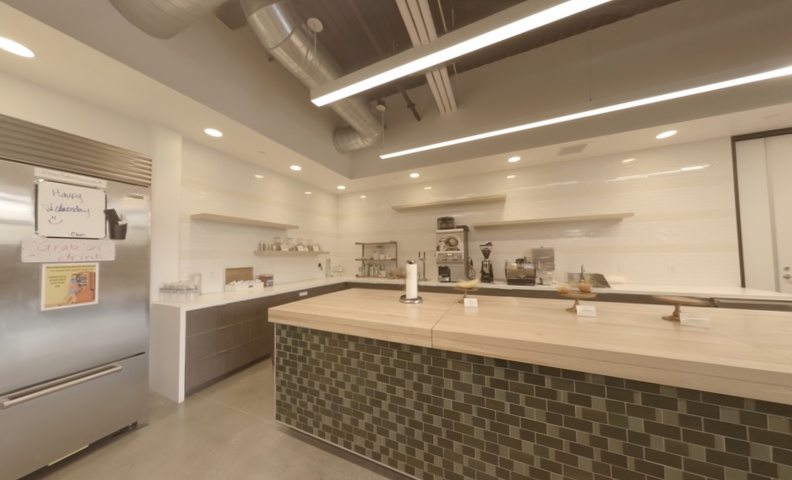}}
{\includegraphics[scale=0.06,clip]{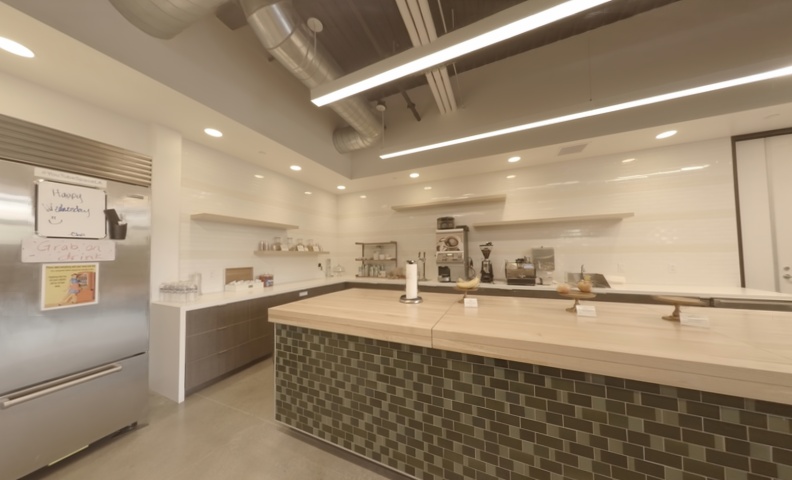}}
{\includegraphics[scale=0.06,clip]{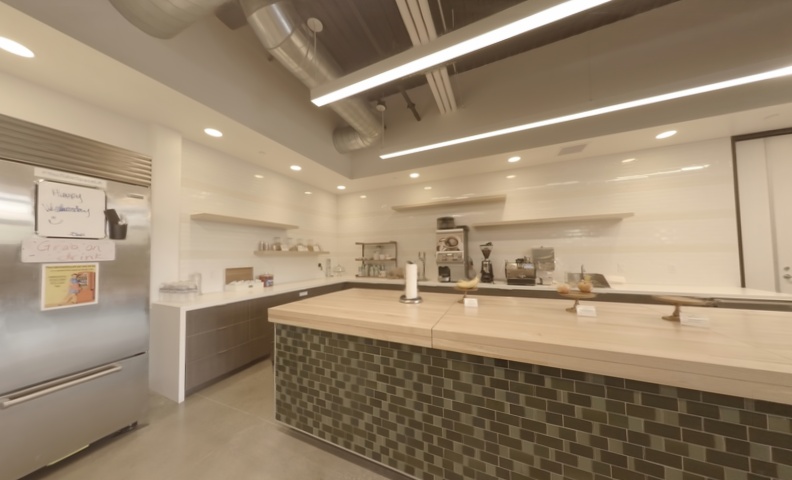}}
{\includegraphics[scale=0.06,clip]{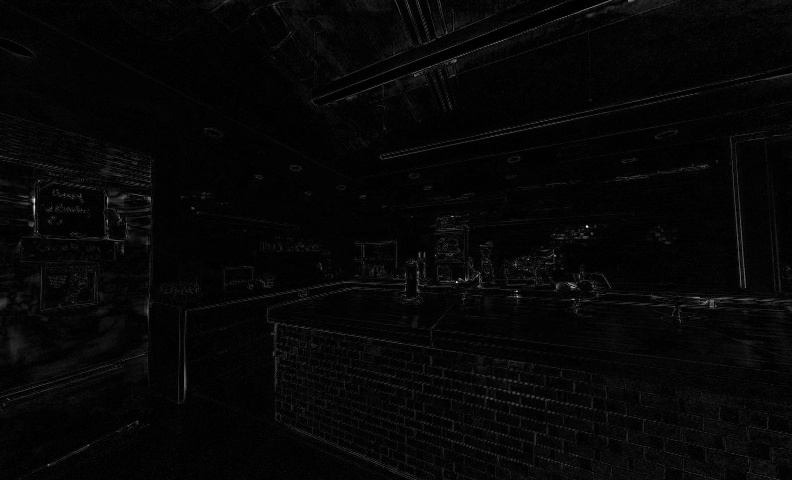}}
{\includegraphics[scale=0.06,clip]{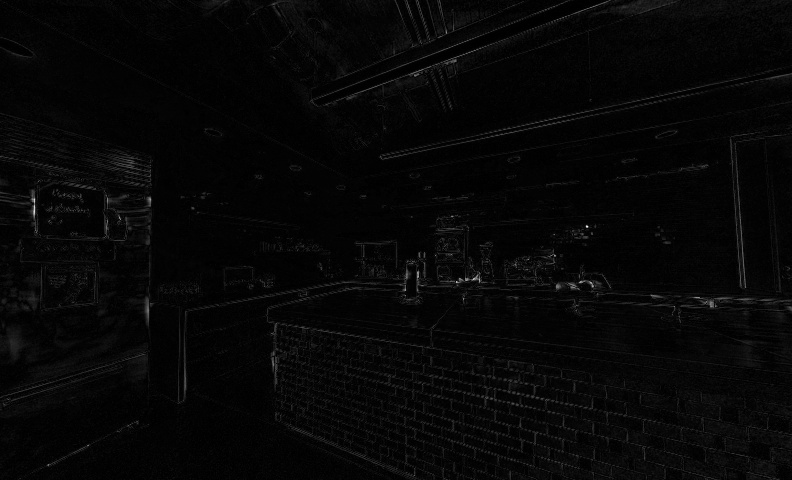}}
{\includegraphics[scale=0.06,clip]{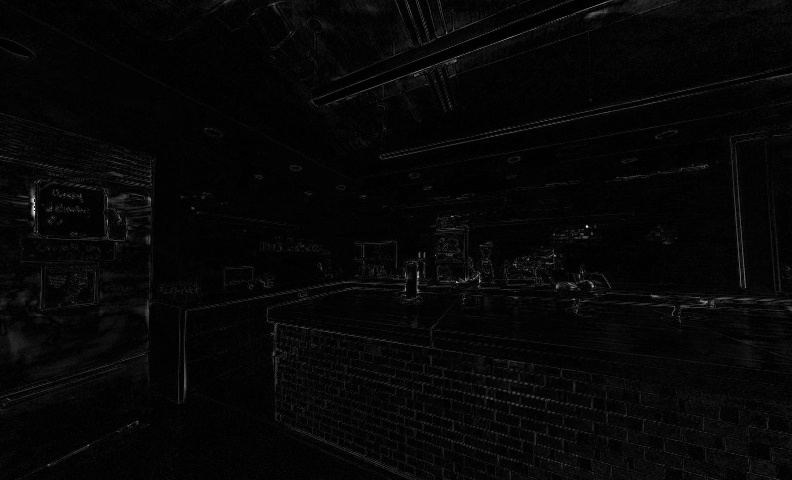}}
{\includegraphics[scale=0.06,clip]{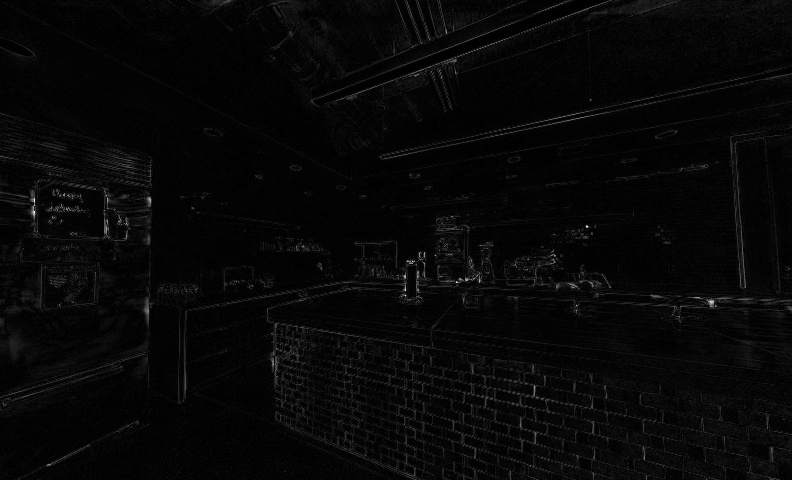}}

{\includegraphics[scale=0.06,clip]{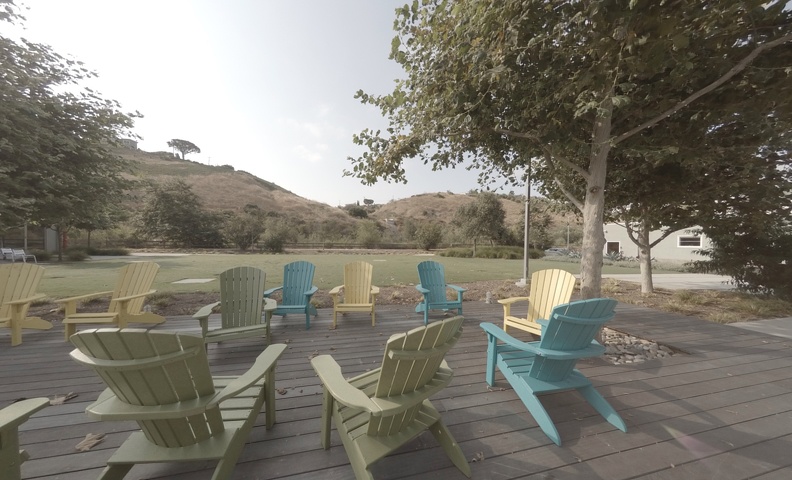}}
{\includegraphics[scale=0.06,clip]{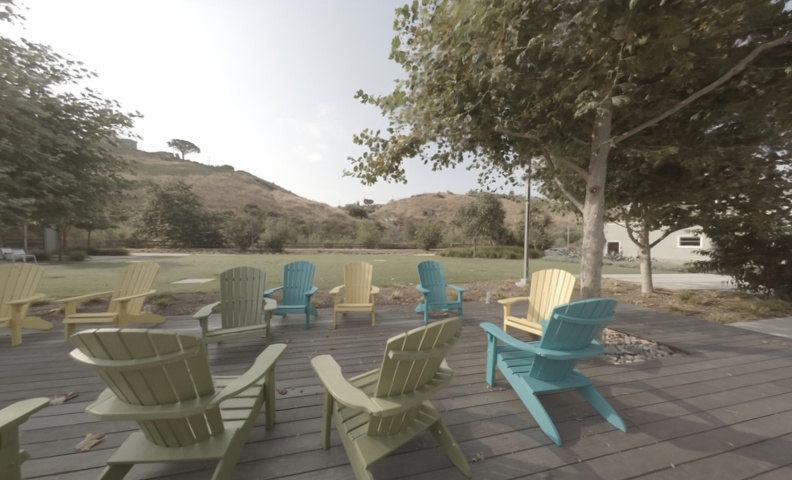}}
{\includegraphics[scale=0.06,clip]{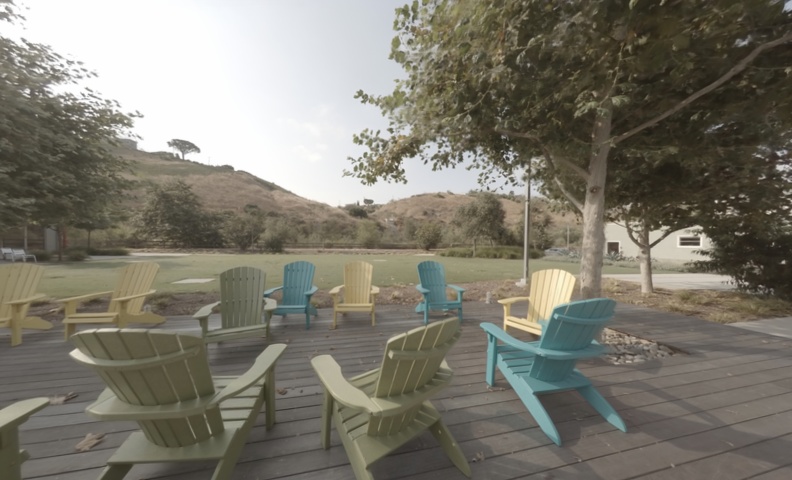}}
{\includegraphics[scale=0.06,clip]{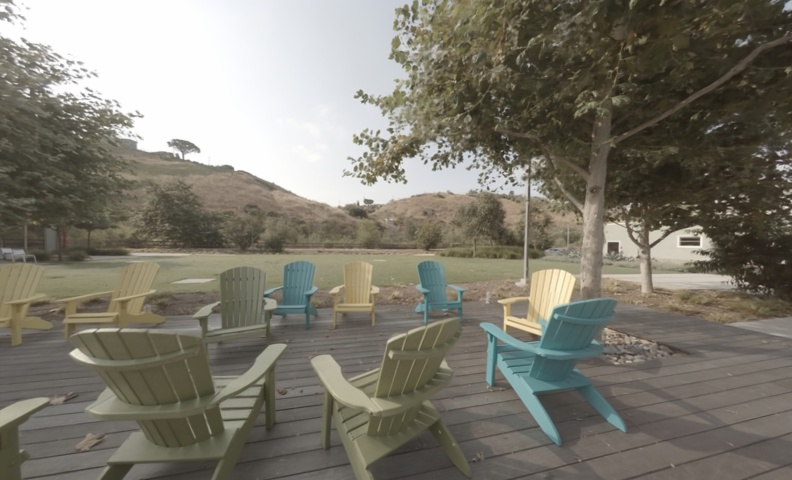}}
{\includegraphics[scale=0.06,clip]{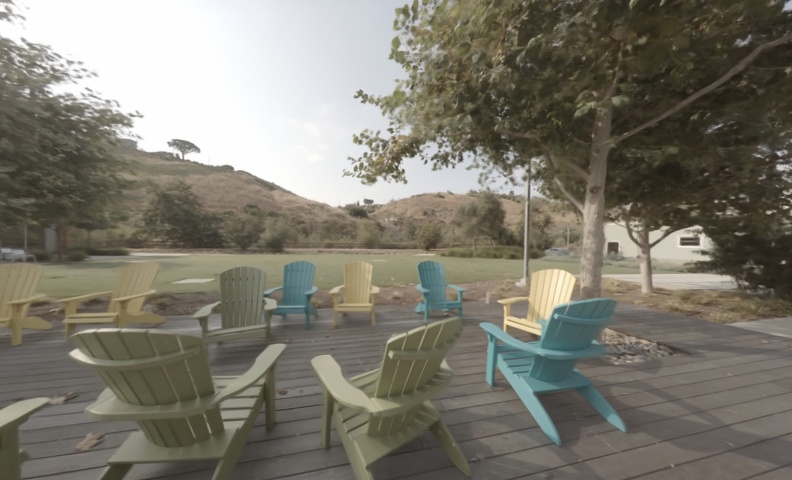}}
{\includegraphics[scale=0.06,clip]{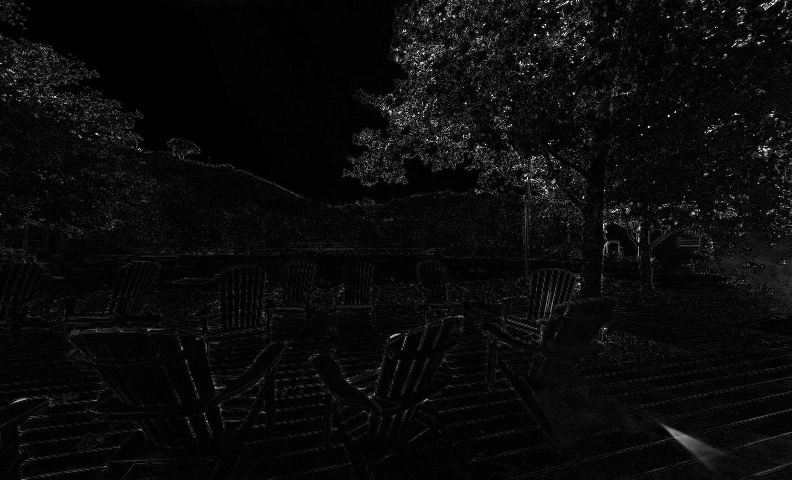}}
{\includegraphics[scale=0.06,clip]{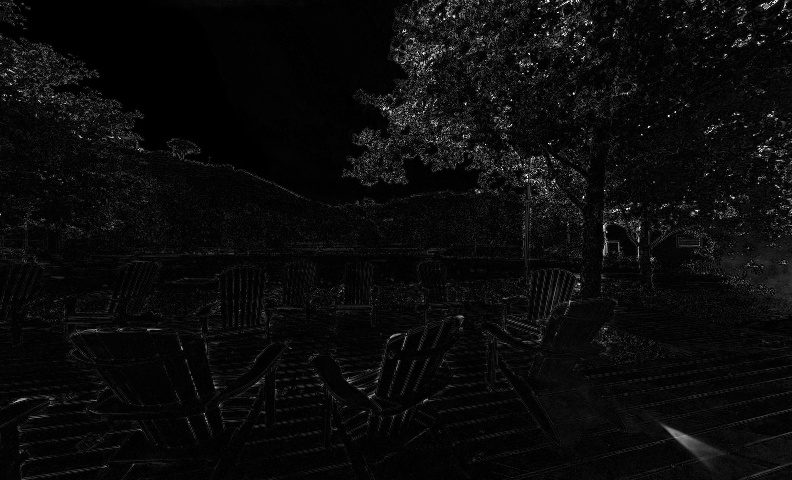}}
{\includegraphics[scale=0.06,clip]{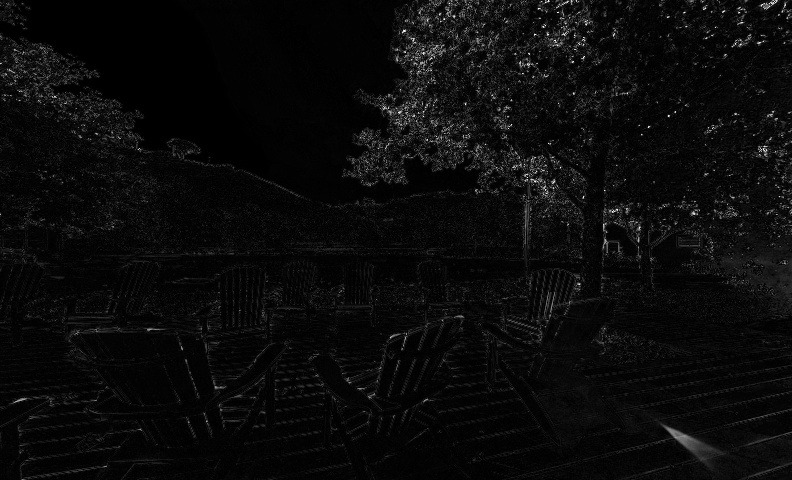}}
{\includegraphics[scale=0.06,clip]{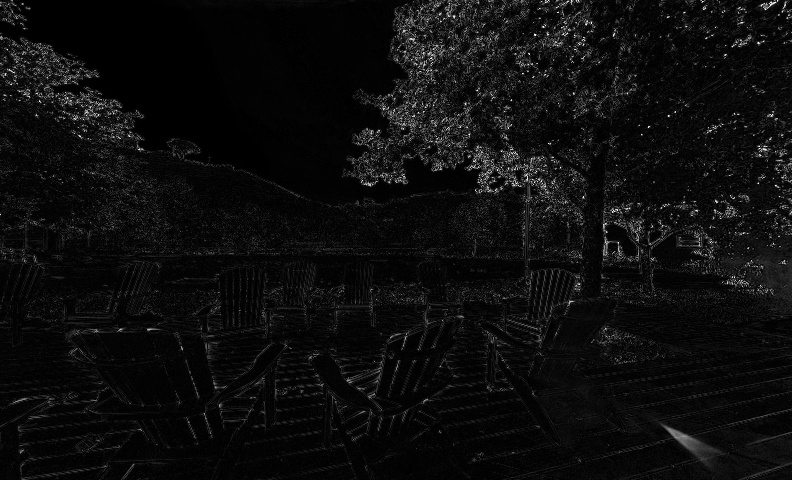}}

{\includegraphics[scale=0.06,clip]{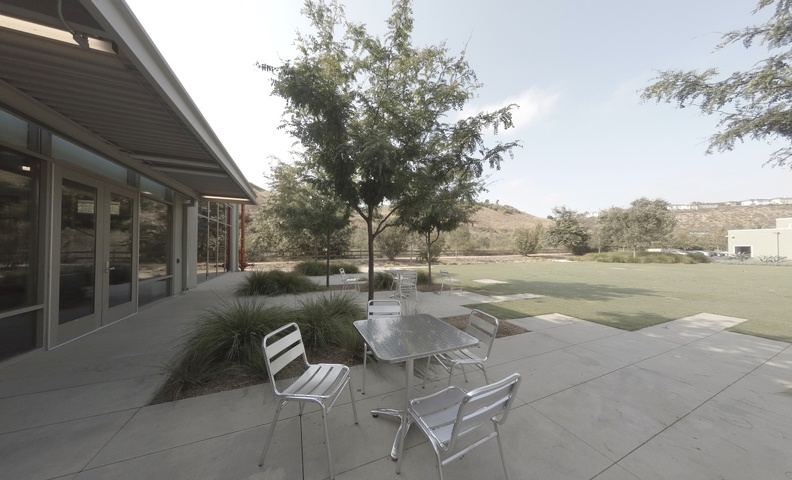}}
{\includegraphics[scale=0.06,clip]{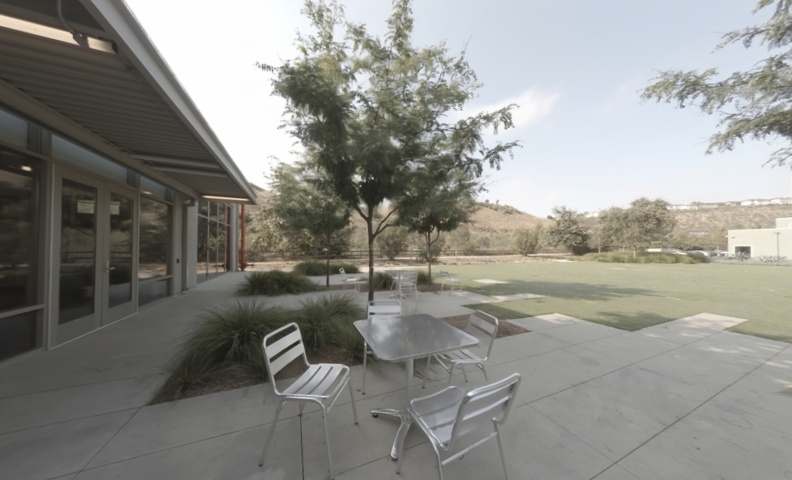}}
{\includegraphics[scale=0.06,clip]{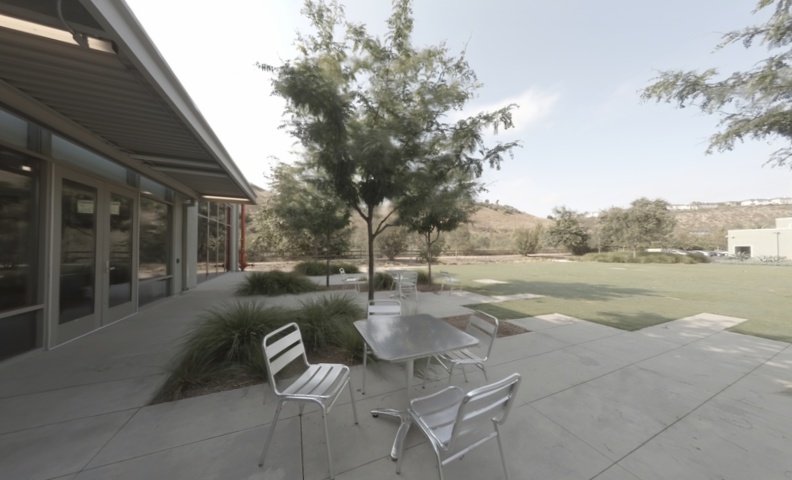}}
{\includegraphics[scale=0.06,clip]{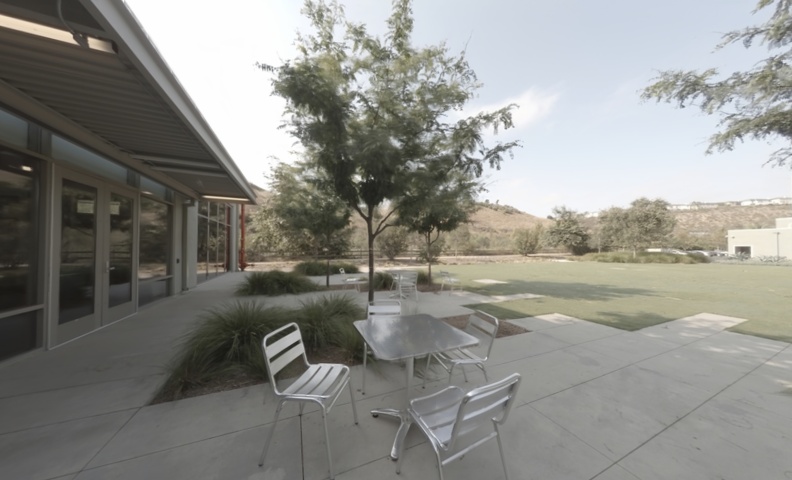}}
{\includegraphics[scale=0.06,clip]{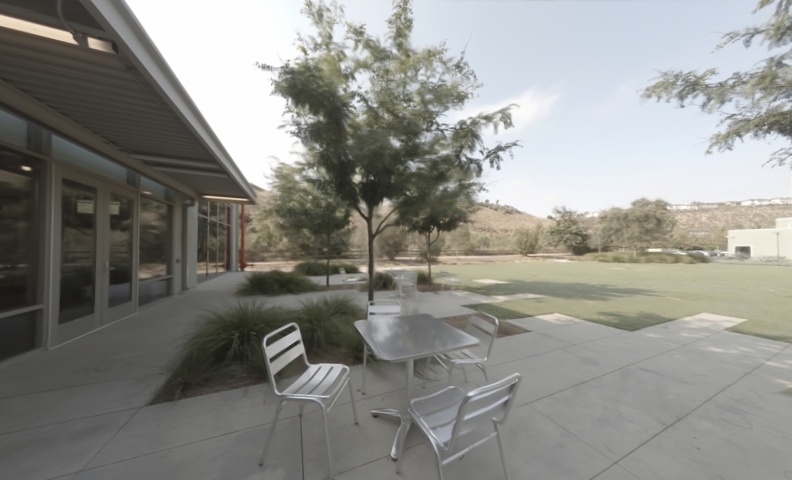}}
{\includegraphics[scale=0.06,clip]{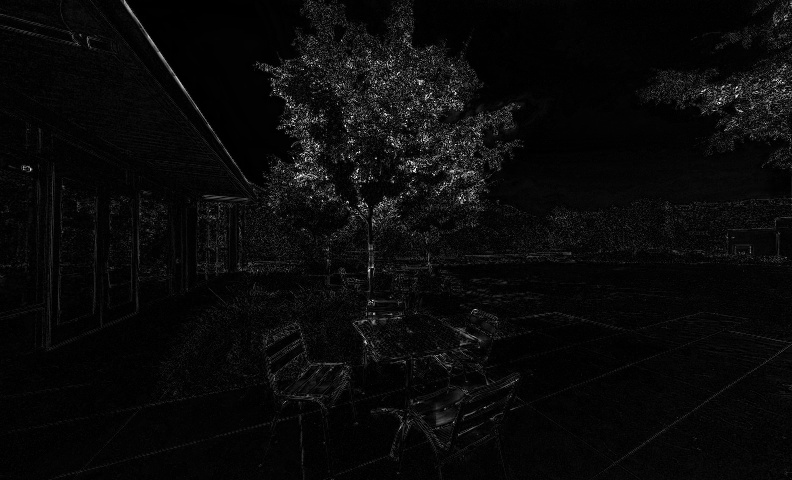}}
{\includegraphics[scale=0.06,clip]{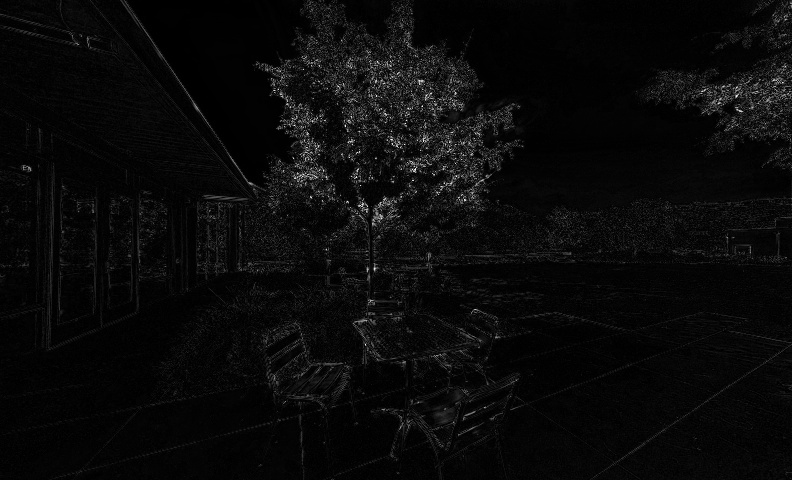}}
{\includegraphics[scale=0.06,clip]{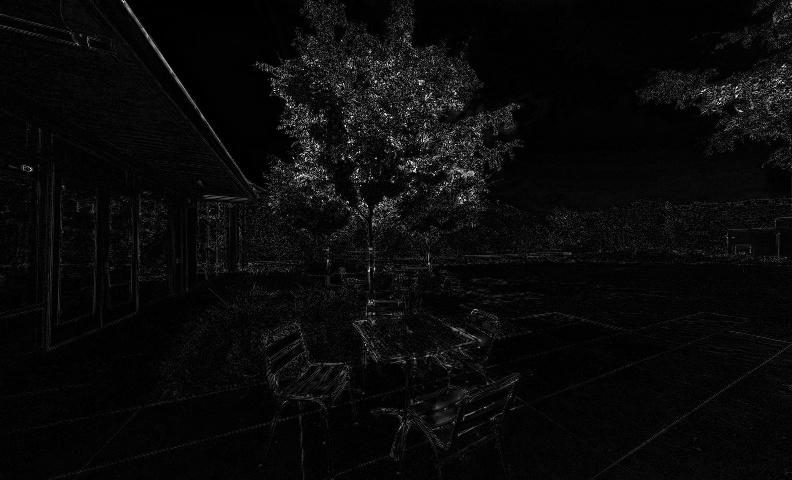}}
{\includegraphics[scale=0.06,clip]{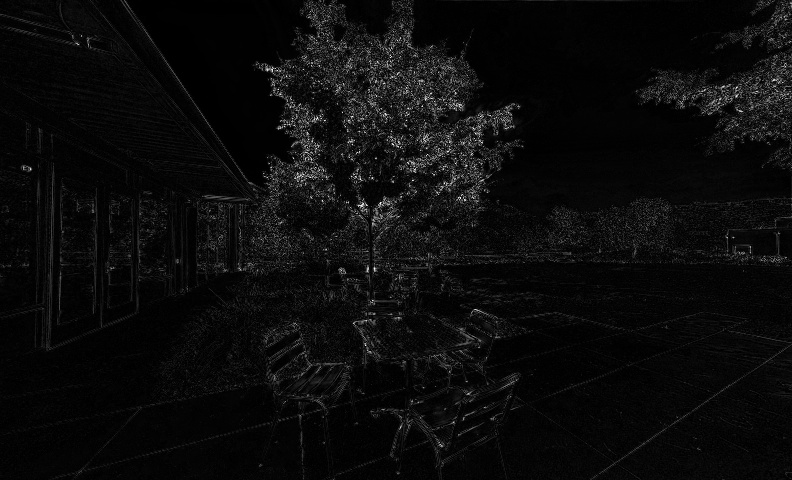}}

{\includegraphics[scale=0.06,clip]{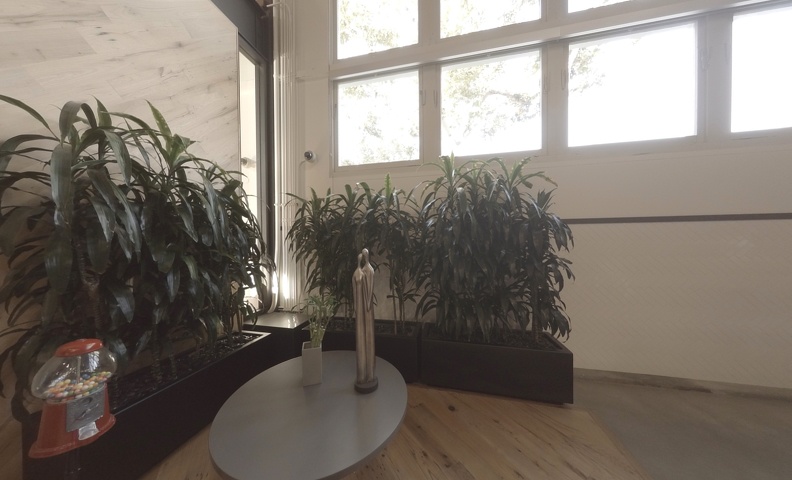}}
{\includegraphics[scale=0.06,clip]{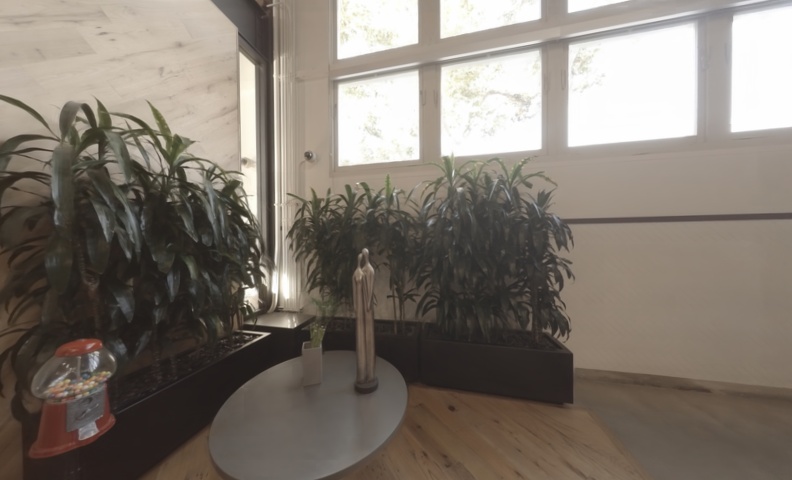}}
{\includegraphics[scale=0.06,clip]{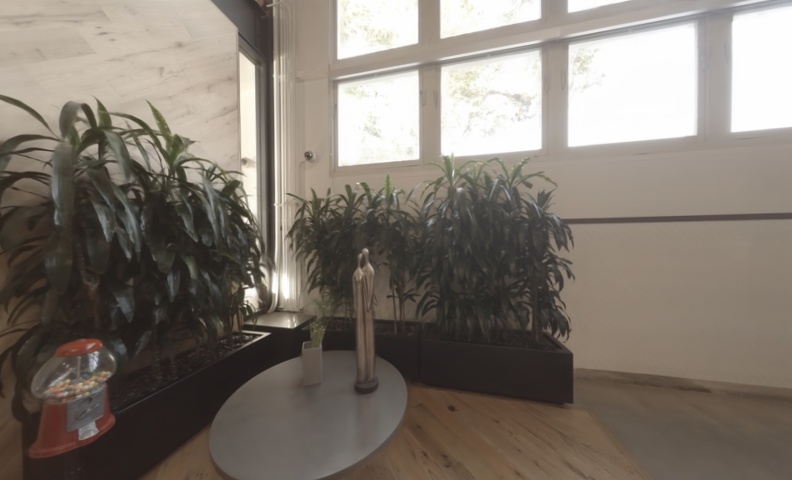}}
{\includegraphics[scale=0.06,clip]{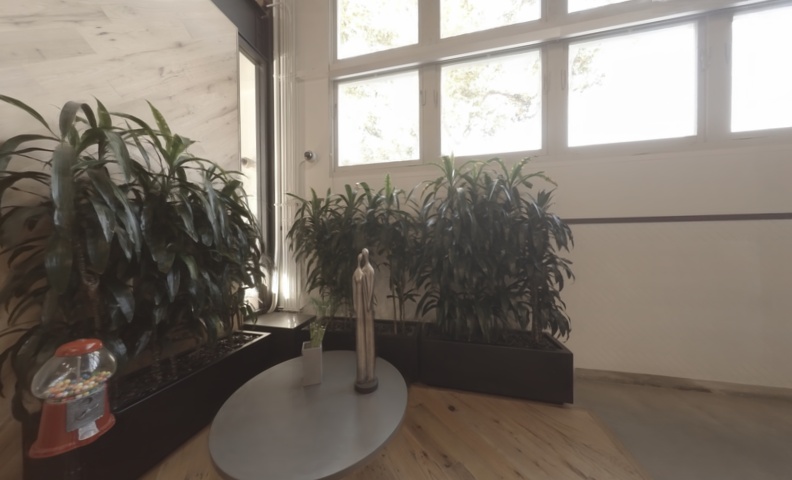}}
{\includegraphics[scale=0.06,clip]{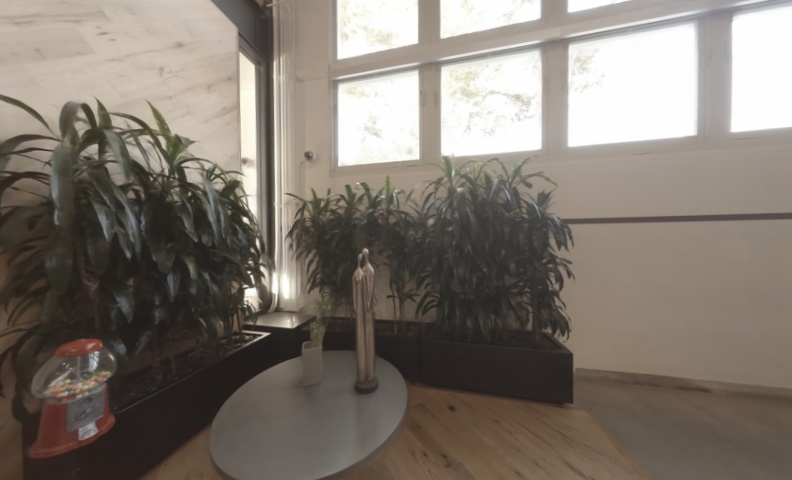}}
{\includegraphics[scale=0.06,clip]{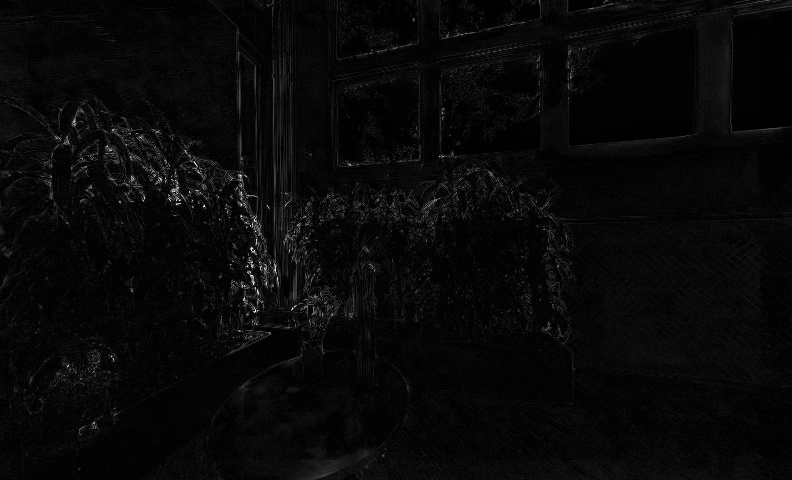}}
{\includegraphics[scale=0.06,clip]{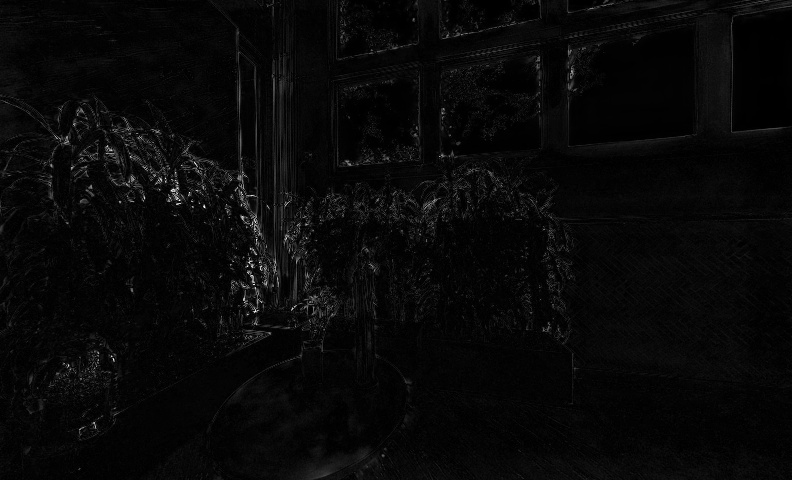}}
{\includegraphics[scale=0.06,clip]{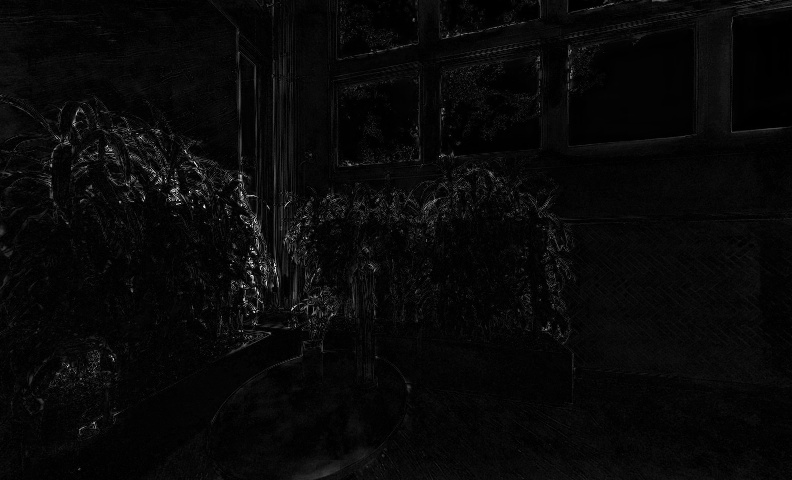}}
{\includegraphics[scale=0.06,clip]{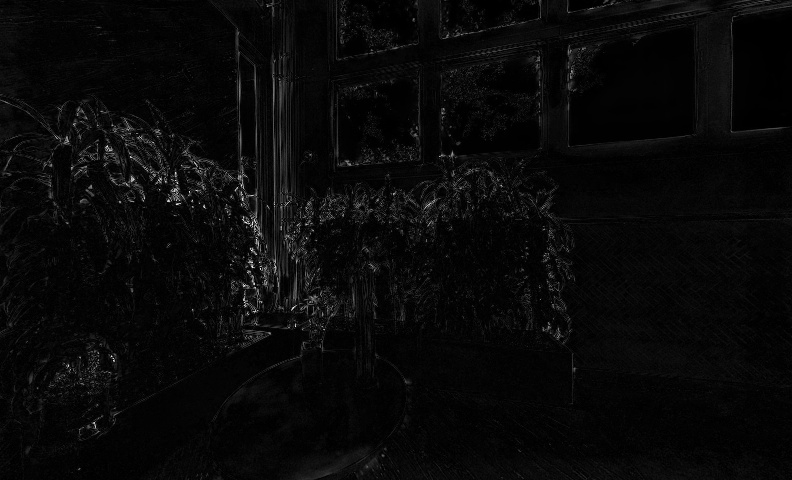}}

{\includegraphics[scale=0.06,clip]{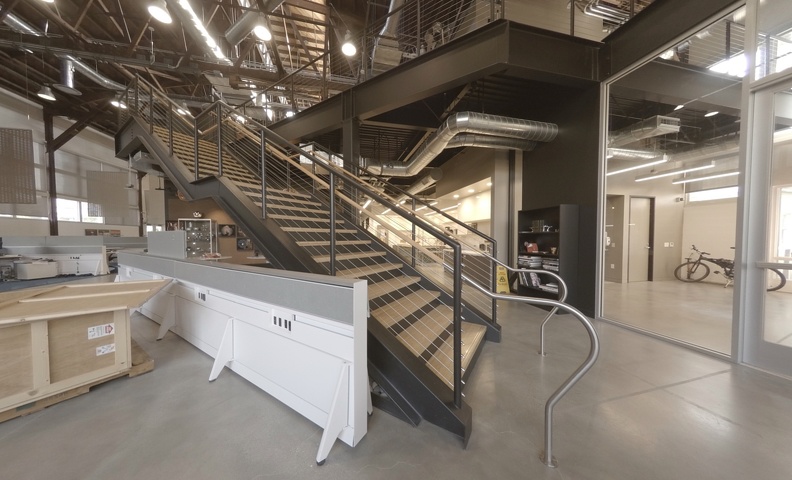}}
{\includegraphics[scale=0.06,clip]{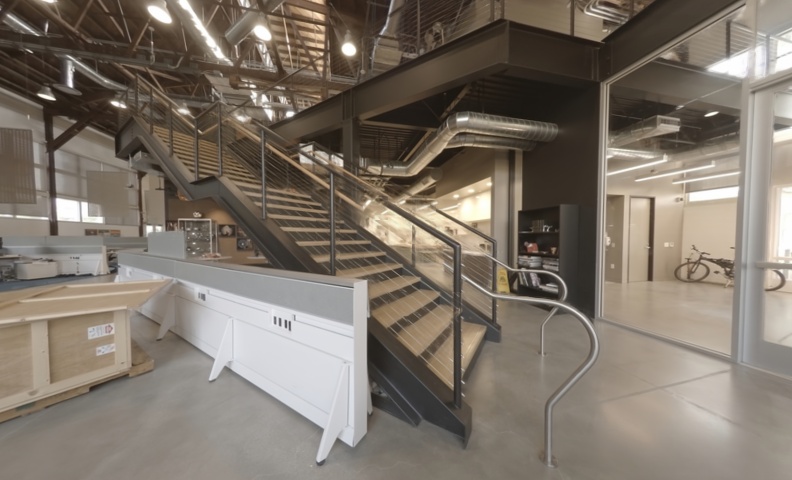}}
{\includegraphics[scale=0.06,clip]{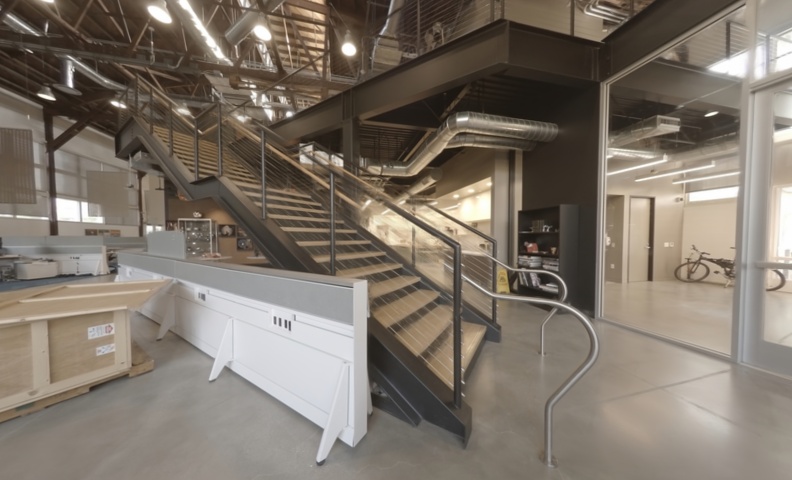}}
{\includegraphics[scale=0.06,clip]{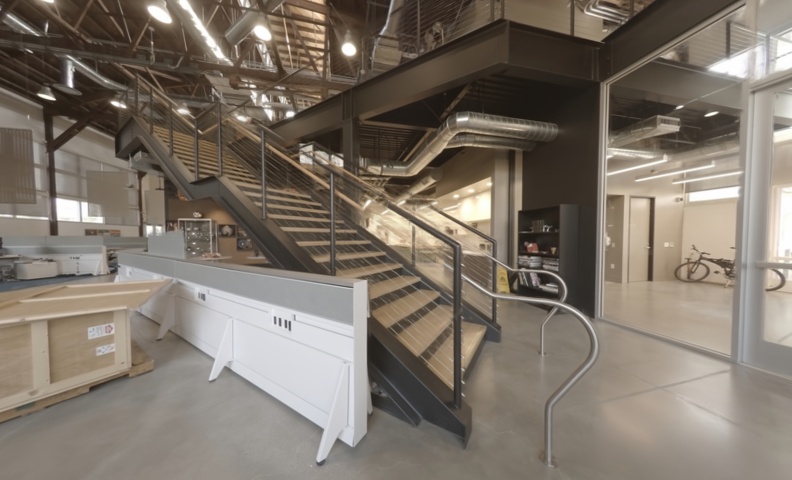}}
{\includegraphics[scale=0.06,clip]{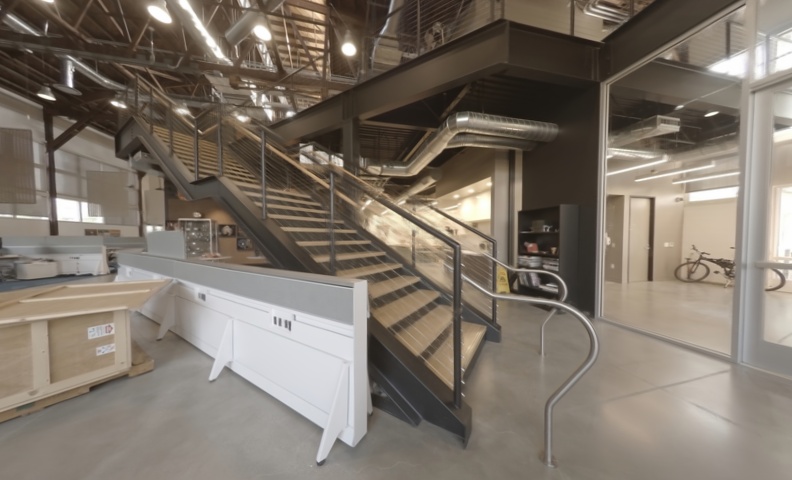}}
{\includegraphics[scale=0.06,clip]{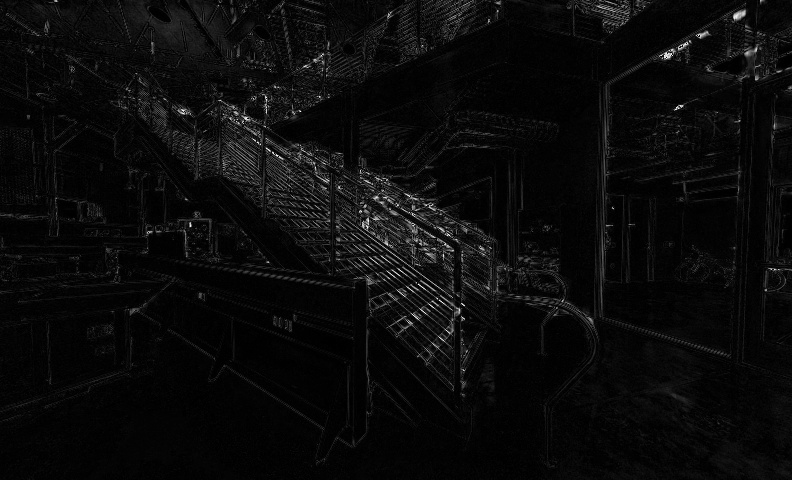}}
{\includegraphics[scale=0.06,clip]{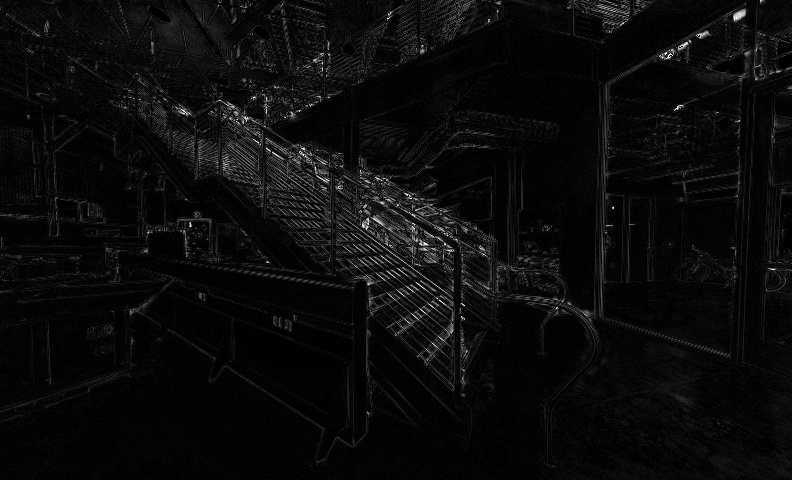}}
{\includegraphics[scale=0.06,clip]{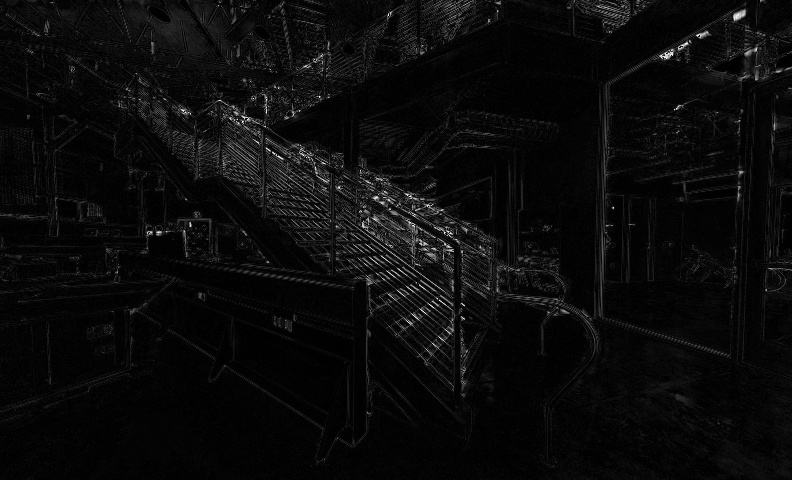}}
{\includegraphics[scale=0.06,clip]{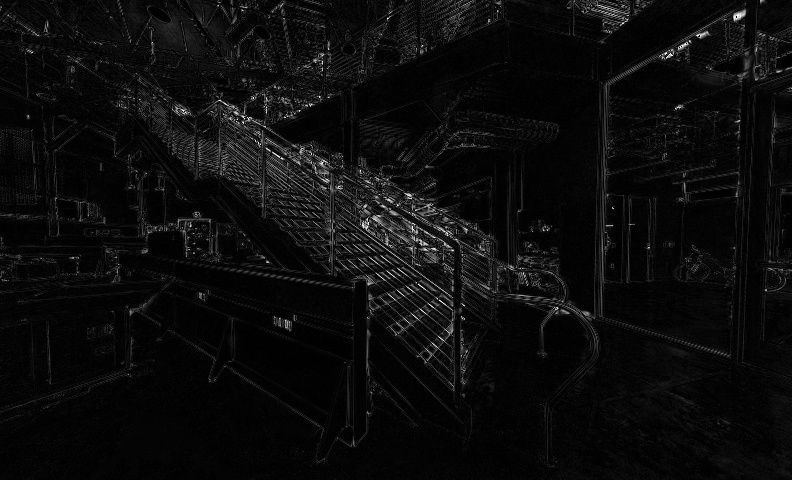}}

{\includegraphics[scale=0.06,clip]{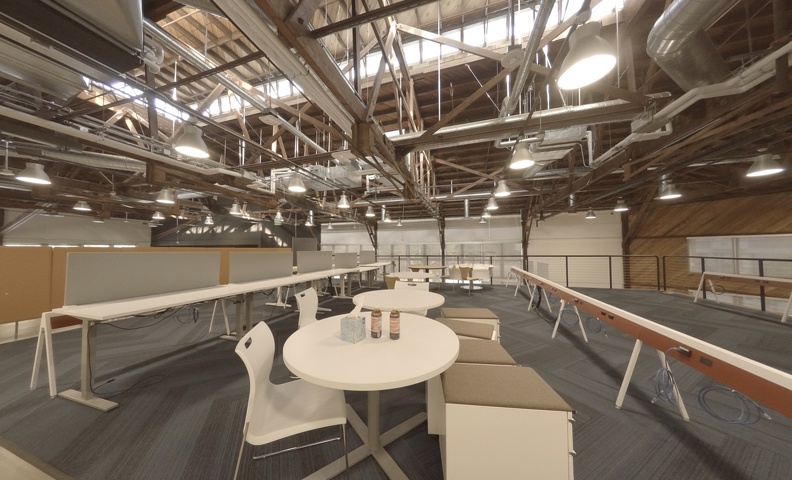}}
{\includegraphics[scale=0.06,clip]{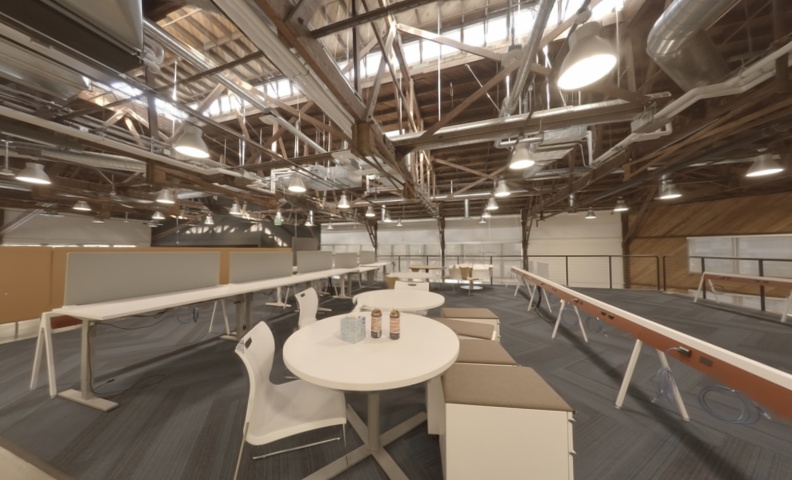}}
{\includegraphics[scale=0.06,clip]{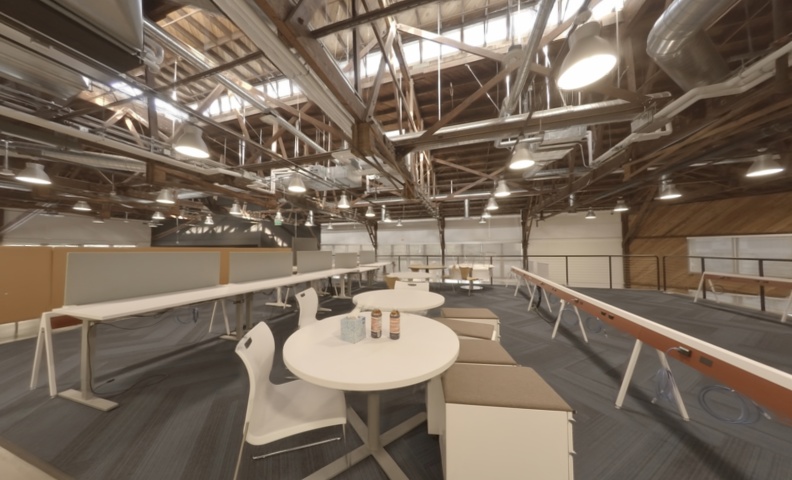}}
{\includegraphics[scale=0.06,clip]{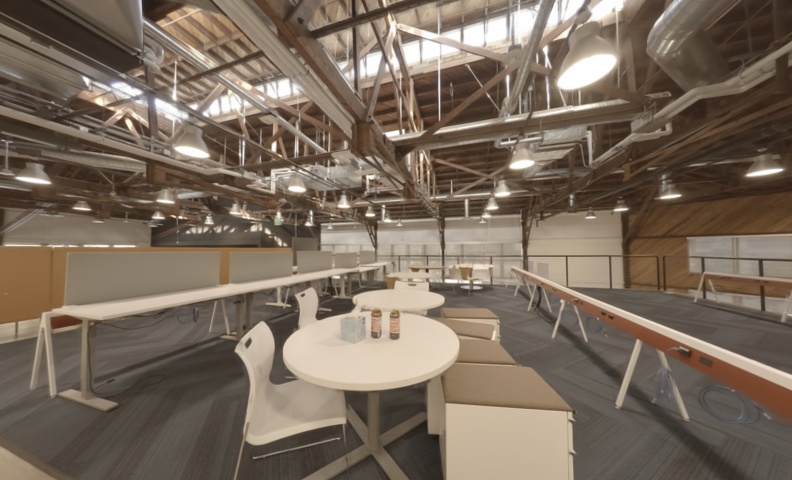}}
{\includegraphics[scale=0.06,clip]{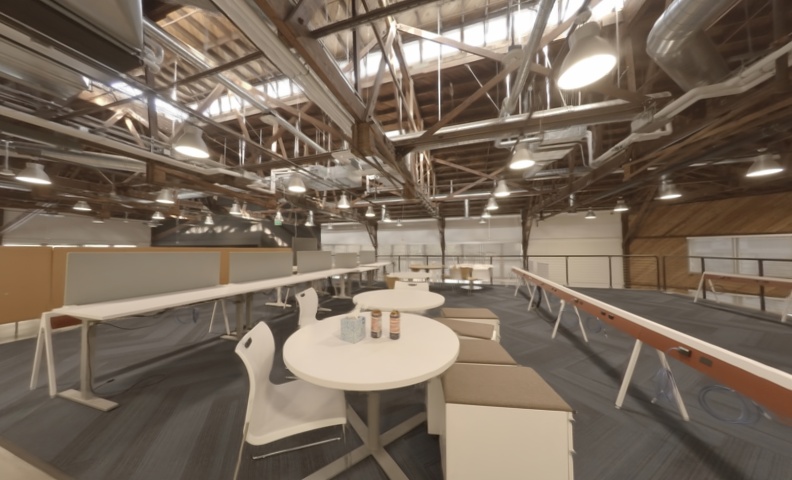}}
{\includegraphics[scale=0.06,clip]{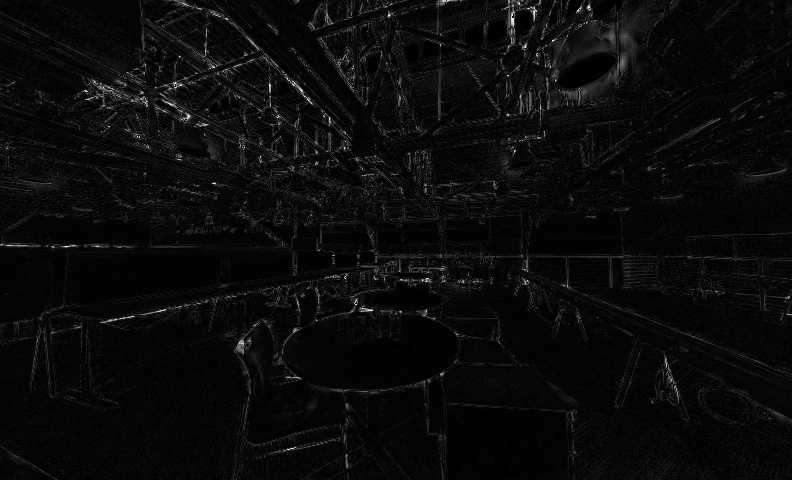}}
{\includegraphics[scale=0.06,clip]{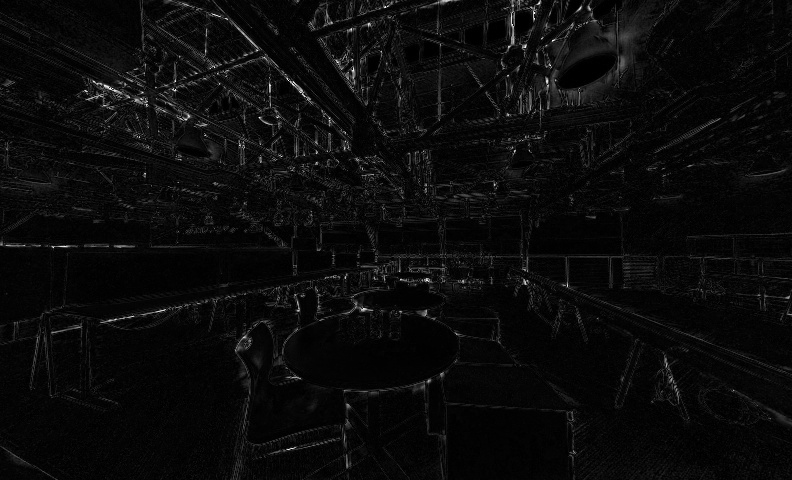}}
{\includegraphics[scale=0.06,clip]{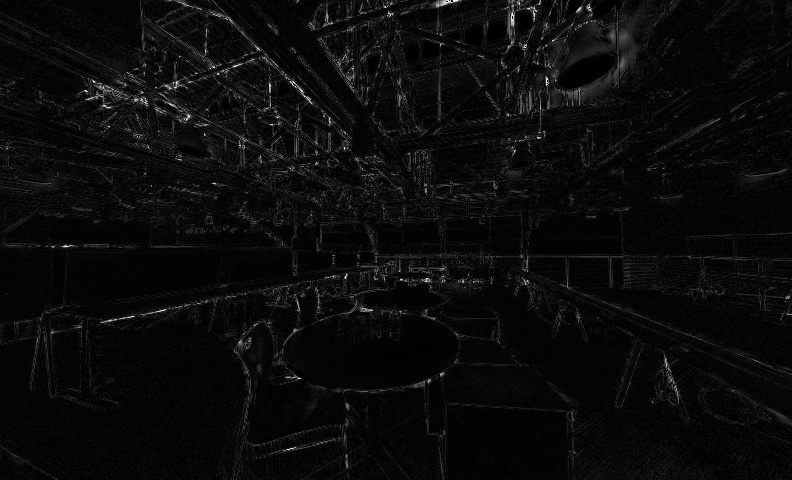}}
{\includegraphics[scale=0.06,clip]{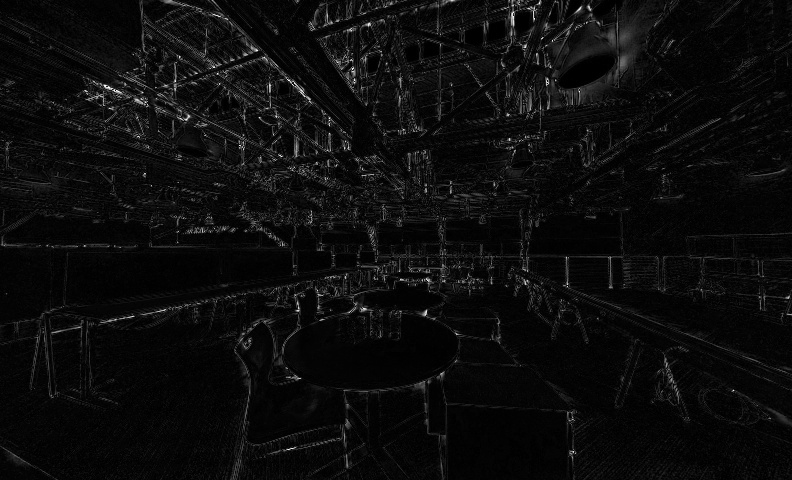}}

\caption{ We show results on Spaces dataset. We show view estimated using different depth levels and effect of multi-resolution analysis. We also show corresponding error images for respective predictions which can be used to compare the quality of the estimates.}
\label{fig:spaces1}
\end{figure*}

\begin{figure*}[t]
\centering
\textbf{\scriptsize Ground Truth \hspace{1.5cm} \scriptsize Ours 64 depths w/ mr \hspace{1.5cm} \scriptsize Soft3d \hspace{2.5cm} \scriptsize ours error\hspace{2.5cm} \scriptsize soft3d error}\par\medskip
{\includegraphics[scale=0.12,clip]{spaces/dense/64mr/scene_000_7_final_gt.jpg}}
{\includegraphics[scale=0.12,clip]{spaces/dense/64mr/scene_000_7_final_pred.jpg}}
{\includegraphics[scale=0.12,clip]{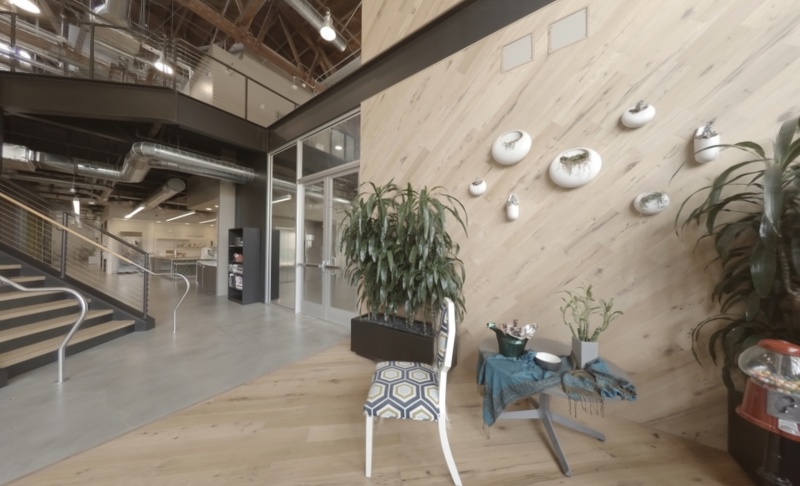}}~
{\includegraphics[scale=0.12,clip]{spaces/dense/64mr/scene_000_7_final_err.jpg}}~
{\includegraphics[scale=0.12,clip]{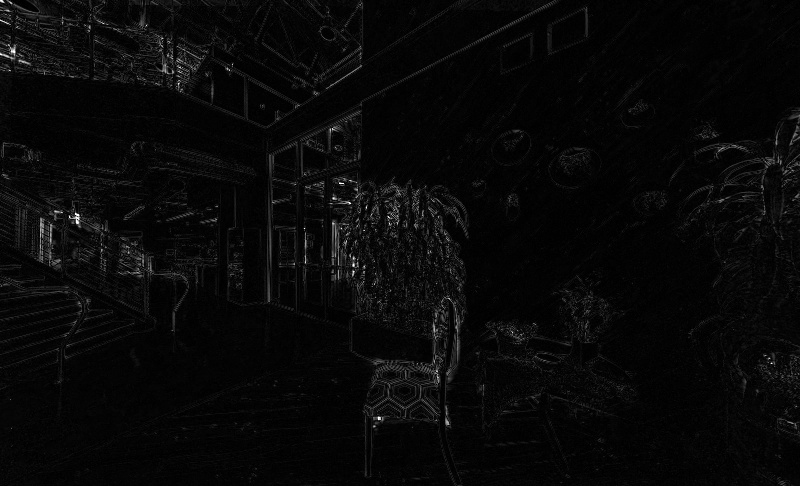}}\\

{\includegraphics[scale=0.12,clip]{spaces/dense/64mr/scene_009_7_final_gt.jpg}}
{\includegraphics[scale=0.12,clip]{spaces/dense/64mr/scene_009_7_final_pred.jpg}}
{\includegraphics[scale=0.12,clip]{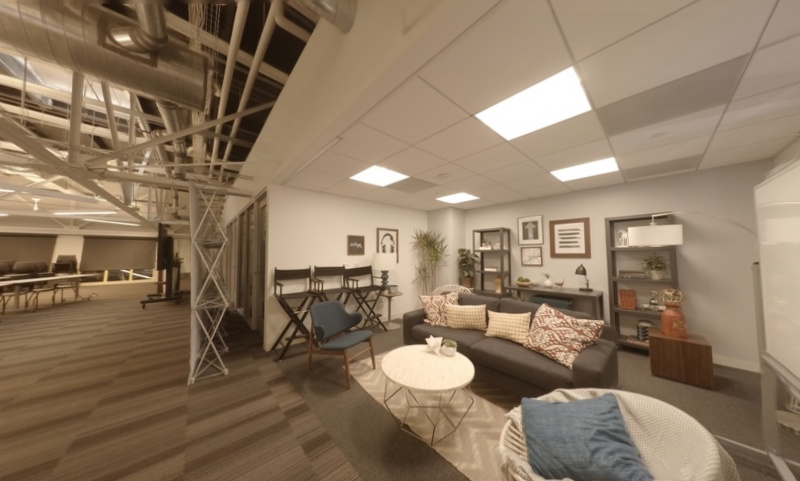}}~
{\includegraphics[scale=0.12,clip]{spaces/dense/64mr/scene_009_7_final_err.jpg}}~
{\includegraphics[scale=0.12,clip]{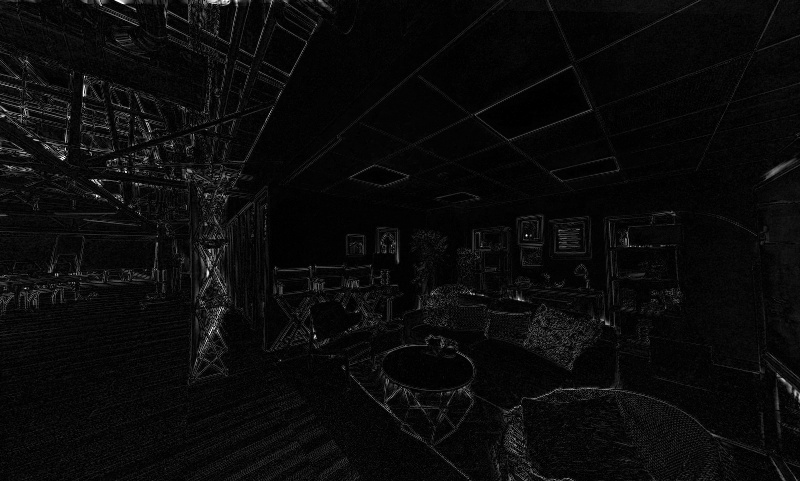}}\\

{\includegraphics[scale=0.12,clip]{spaces/dense/64mr/scene_010_7_final_gt.jpg}}
{\includegraphics[scale=0.12,clip]{spaces/dense/64mr/scene_010_7_final_pred.jpg}}
{\includegraphics[scale=0.12,clip]{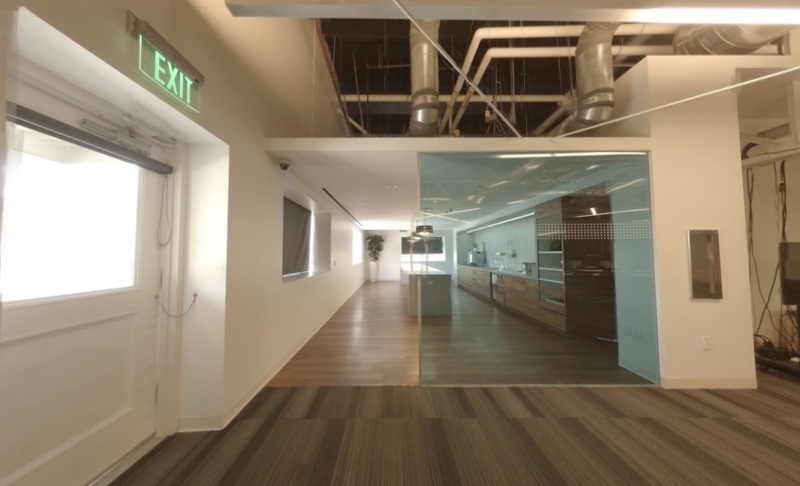}}~
{\includegraphics[scale=0.12,clip]{spaces/dense/64mr/scene_010_7_final_err.jpg}}~
{\includegraphics[scale=0.12,clip]{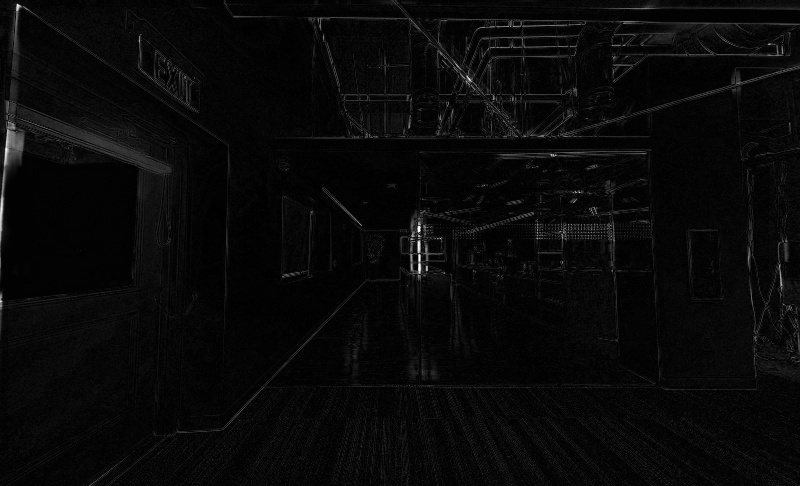}}\\

{\includegraphics[scale=0.12,clip]{spaces/dense/64mr/scene_023_7_final_gt.jpg}}
{\includegraphics[scale=0.12,clip]{spaces/dense/64mr/scene_023_7_final_pred.jpg}}
{\includegraphics[scale=0.12,clip]{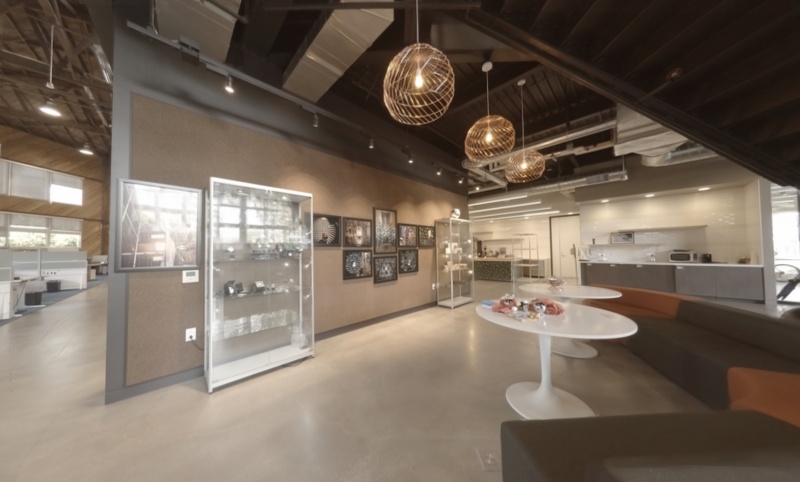}}~
{\includegraphics[scale=0.12,clip]{spaces/dense/64mr/scene_023_7_final_err.jpg}}~
{\includegraphics[scale=0.12,clip]{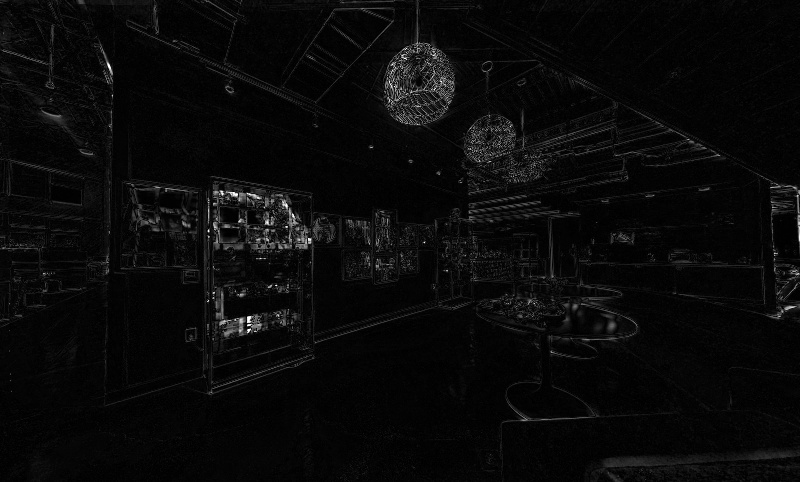}}\\

{\includegraphics[scale=0.12,clip]{spaces/dense/64mr/scene_024_7_final_gt.jpg}}
{\includegraphics[scale=0.12,clip]{spaces/dense/64mr/scene_024_7_final_pred.jpg}}
{\includegraphics[scale=0.12,clip]{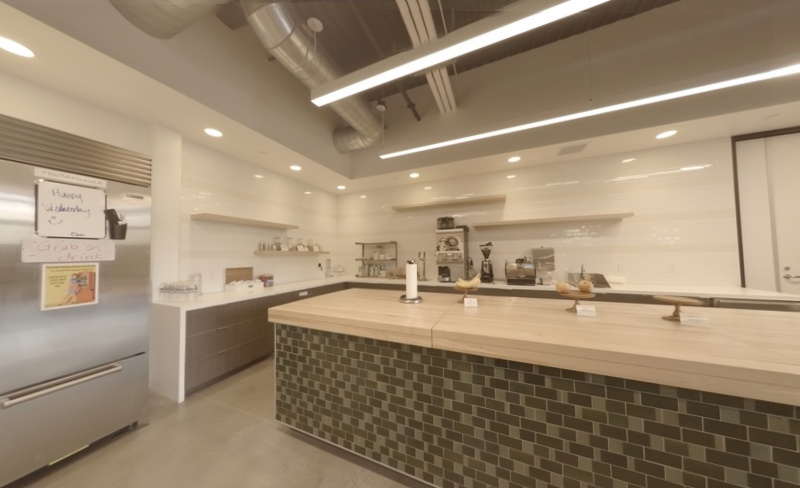}}~
{\includegraphics[scale=0.12,clip]{spaces/dense/64mr/scene_024_7_final_err.jpg}}~
{\includegraphics[scale=0.12,clip]{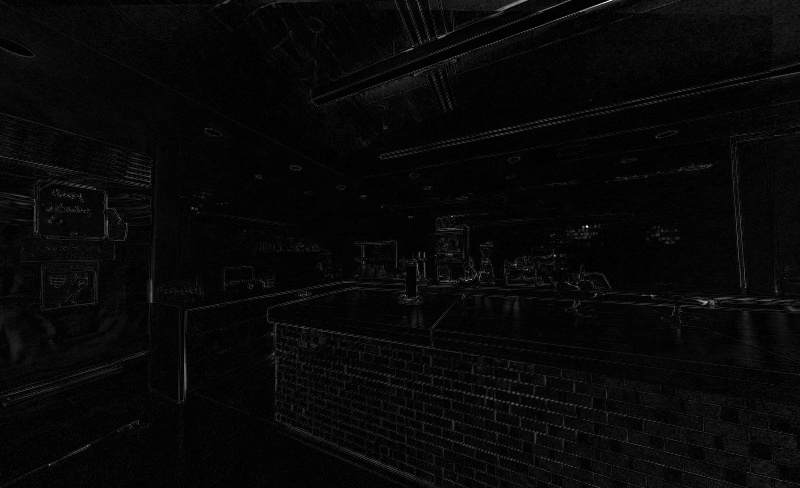}}\\

{\includegraphics[scale=0.12,clip]{spaces/dense/64mr/scene_052_7_final_gt.jpg}}
{\includegraphics[scale=0.12,clip]{spaces/dense/64mr/scene_052_7_final_pred.jpg}}
{\includegraphics[scale=0.12,clip]{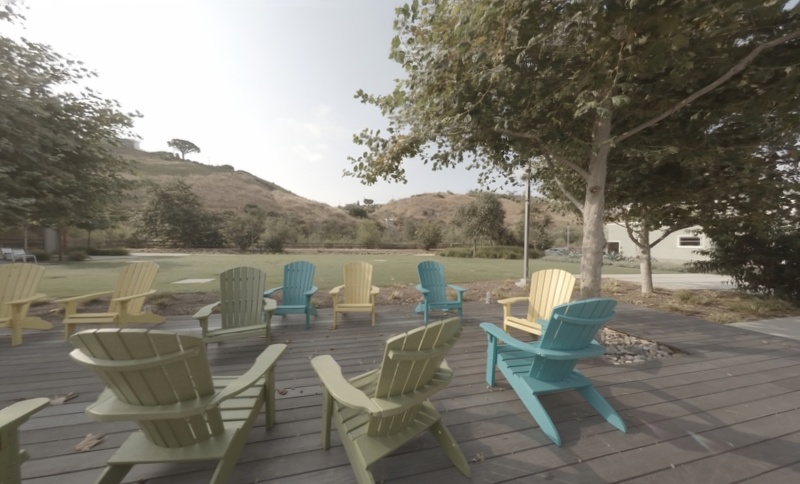}}~
{\includegraphics[scale=0.12,clip]{spaces/dense/64mr/scene_052_7_final_err.jpg}}~
{\includegraphics[scale=0.12,clip]{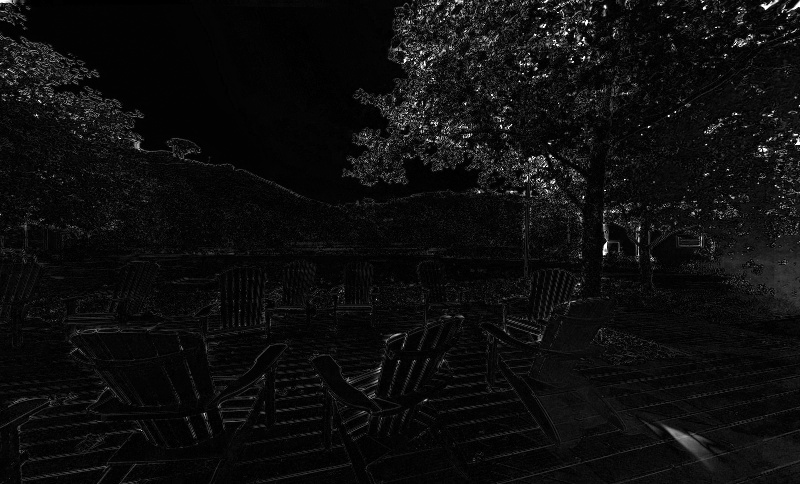}}\\

{\includegraphics[scale=0.12,clip]{spaces/dense/64mr/scene_056_7_final_gt.jpg}}
{\includegraphics[scale=0.12,clip]{spaces/dense/64mr/scene_056_7_final_pred.jpg}}
{\includegraphics[scale=0.12,clip]{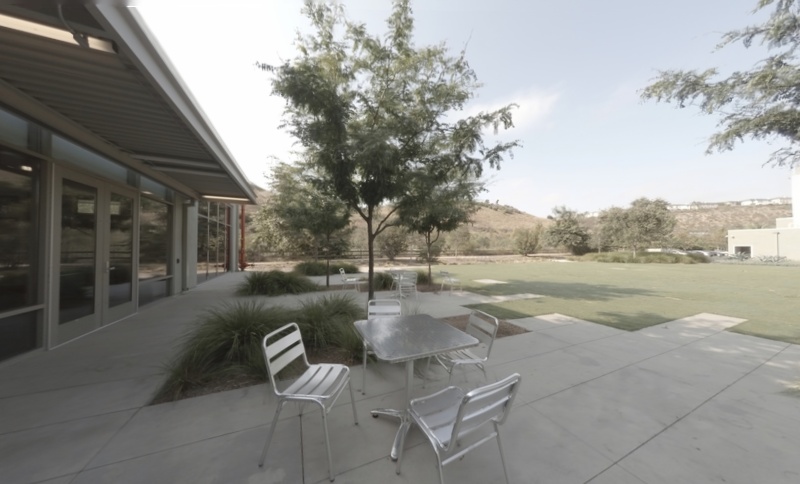}}~
{\includegraphics[scale=0.12,clip]{spaces/dense/64mr/scene_056_7_final_err.jpg}}~
{\includegraphics[scale=0.12,clip]{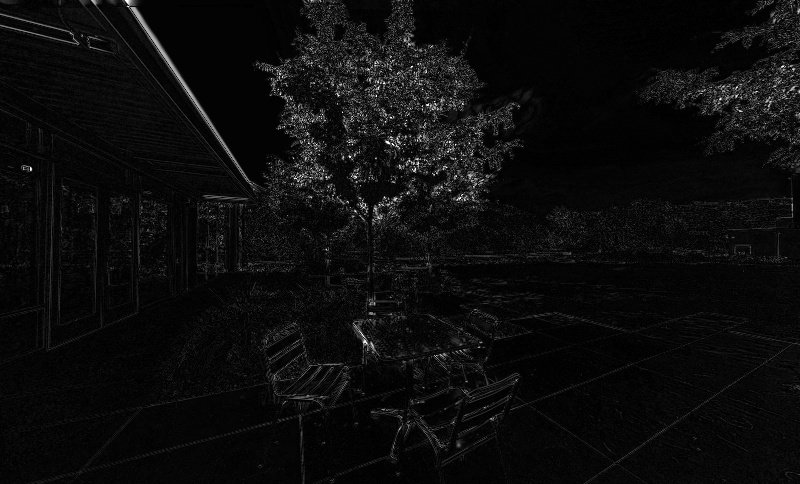}}\\

{\includegraphics[scale=0.12,clip]{spaces/dense/64mr/scene_062_7_final_gt.jpg}}
{\includegraphics[scale=0.12,clip]{spaces/dense/64mr/scene_062_7_final_pred.jpg}}
{\includegraphics[scale=0.12,clip]{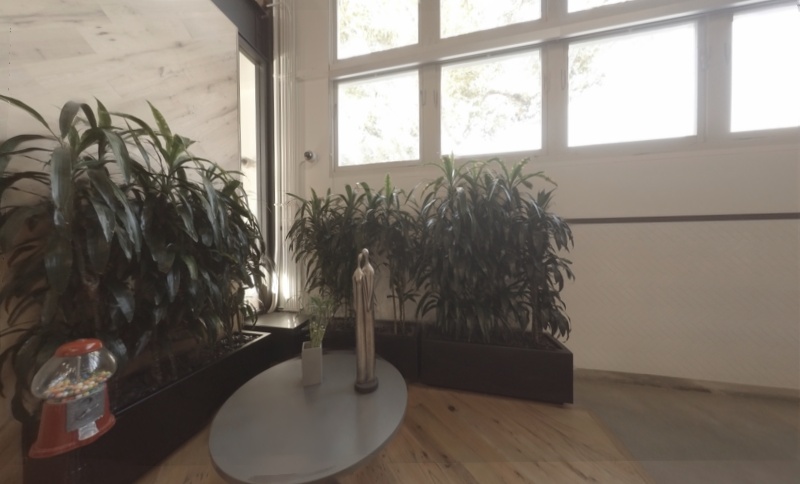}}~
{\includegraphics[scale=0.12,clip]{spaces/dense/64mr/scene_062_7_final_err.jpg}}~
{\includegraphics[scale=0.12,clip]{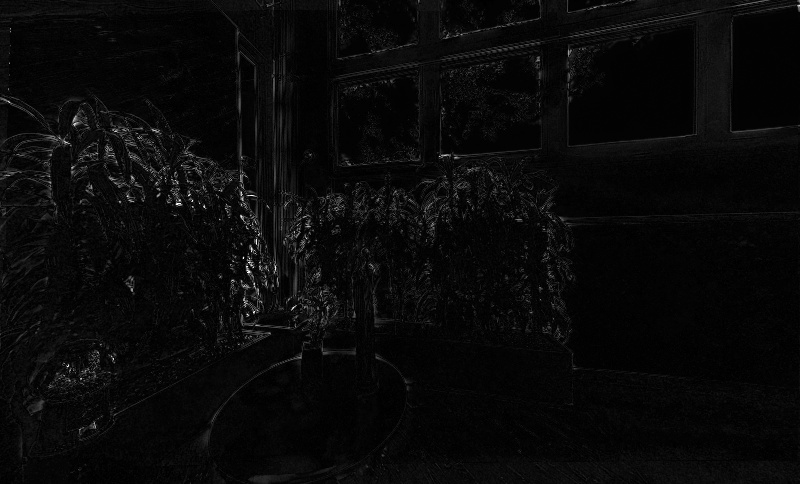}}\\

{\includegraphics[scale=0.12,clip]{spaces/dense/64mr/scene_063_7_final_gt.jpg}}
{\includegraphics[scale=0.12,clip]{spaces/dense/64mr/scene_063_7_final_pred.jpg}}
{\includegraphics[scale=0.12,clip]{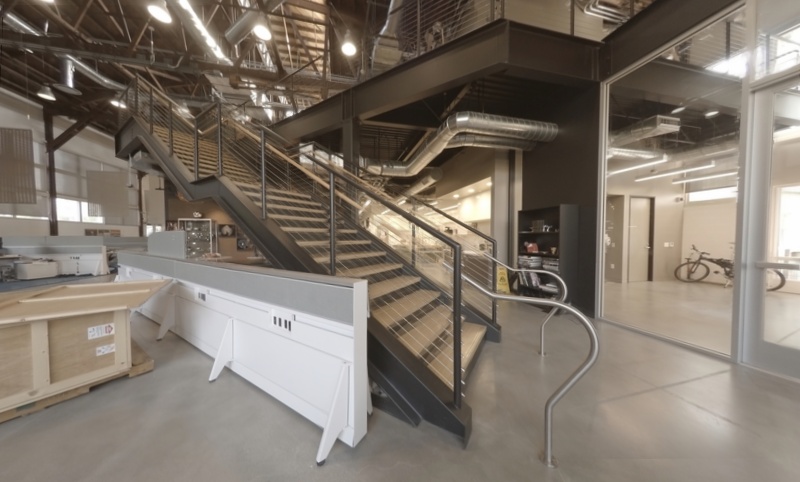}}~
{\includegraphics[scale=0.12,clip]{spaces/dense/64mr/scene_063_7_final_err.jpg}}~
{\includegraphics[scale=0.12,clip]{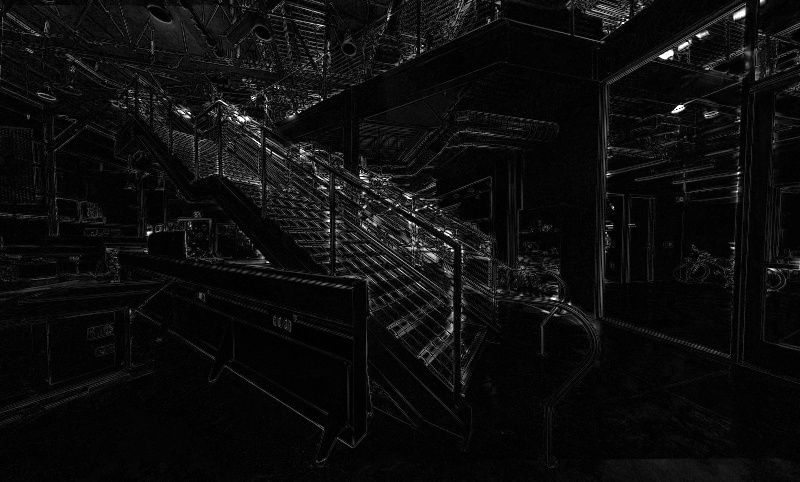}}\\

{\includegraphics[scale=0.12,clip]{spaces/dense/64mr/scene_073_7_final_gt.jpg}}
{\includegraphics[scale=0.12,clip]{spaces/dense/64mr/scene_073_7_final_pred.jpg}}
{\includegraphics[scale=0.12,clip]{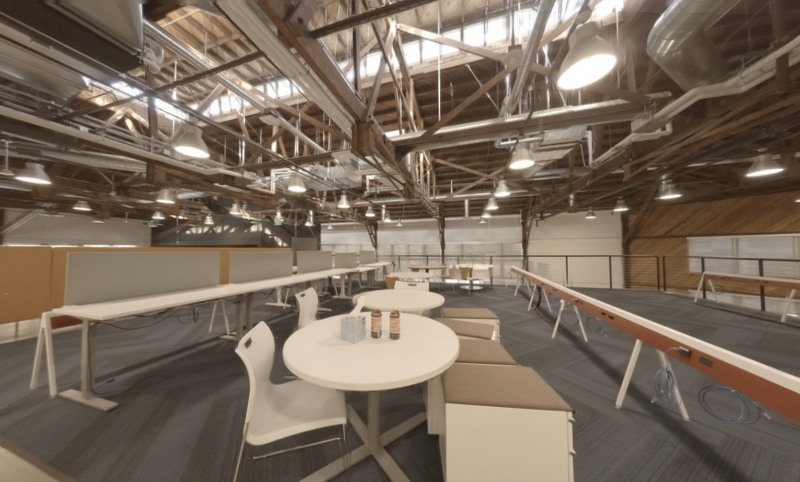}}~
{\includegraphics[scale=0.12,clip]{spaces/dense/64mr/scene_073_7_final_err.jpg}}~
{\includegraphics[scale=0.12,clip]{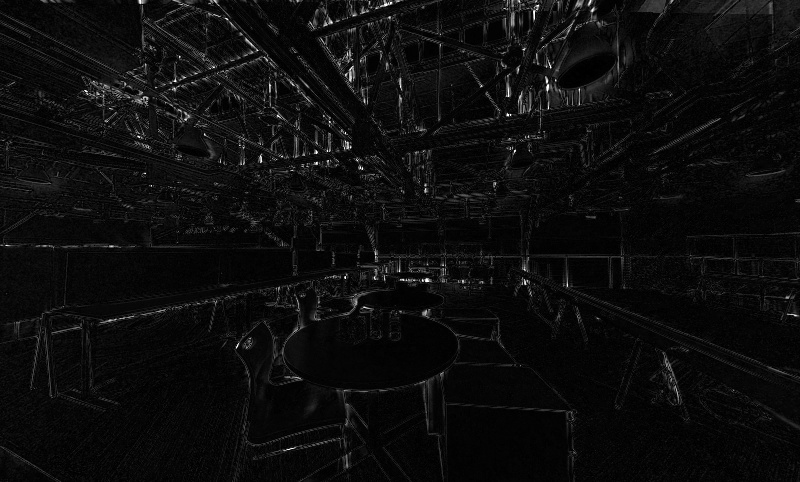}}\\

\caption{ We show results on $10$ scenes in the spaces dataset. Each row show results on different scene. We also show an error images for respective predictions which can be used to compare the quality of the estimates.}
\label{fig:spaces0}
\end{figure*}

We show effect of different depth levels in the estimated views on Spaces dataset in Fig.~\ref{fig:spaces1}. 
We show results on all $10$ scenes from the dataset in Fig.~\ref{fig:spaces0}. The figure show results using $64$ depth levels with multi-resolution analysis.

\subsection{Additional Results on light field dataset}
\begin{figure*}[t]
\centering
\textbf{\scriptsize Ground Truth \hspace{1.5cm} \scriptsize Ours (Spaces) \hspace{1.5cm} \scriptsize Ours (KITTI) \hspace{2.5cm} \scriptsize Soft3d}\par\medskip
{\includegraphics[scale=0.2,clip]{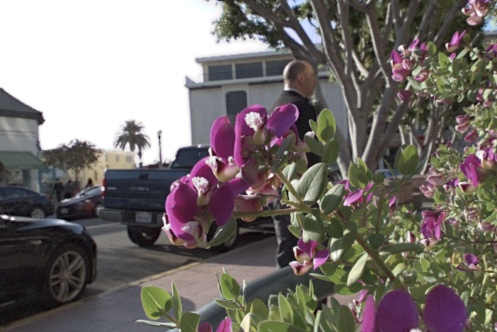}}~
{\includegraphics[scale=0.2,clip]{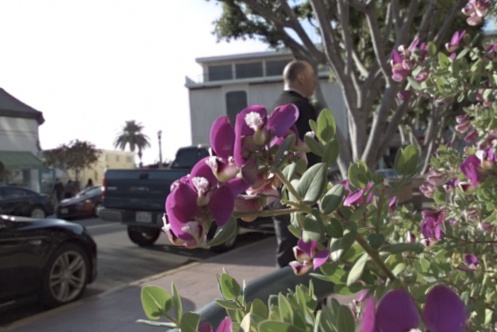}}~
{\includegraphics[scale=0.2,clip]{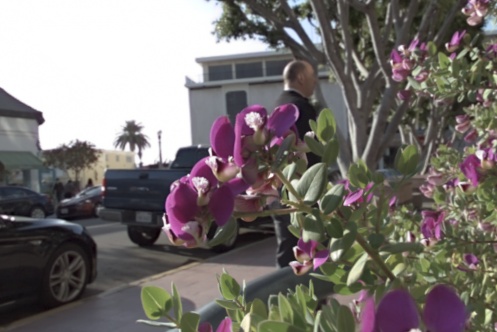}}~
{\includegraphics[scale=0.2,clip]{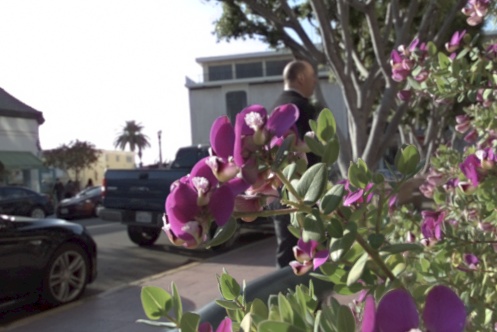}}\\
{\includegraphics[scale=0.2,clip]{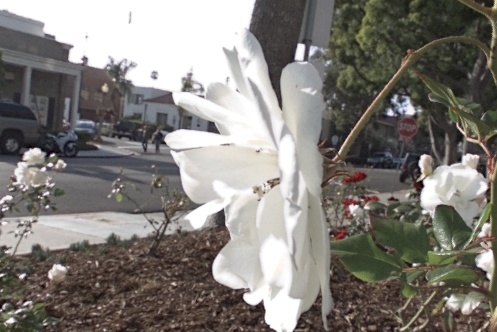}}~
{\includegraphics[scale=0.2,clip]{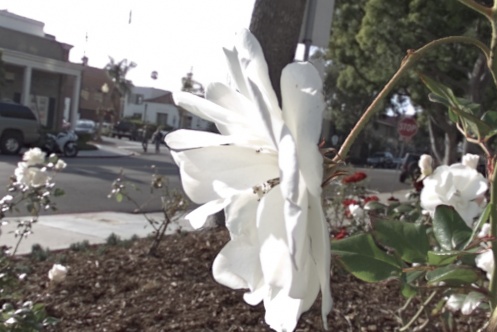}}~
{\includegraphics[scale=0.2,clip]{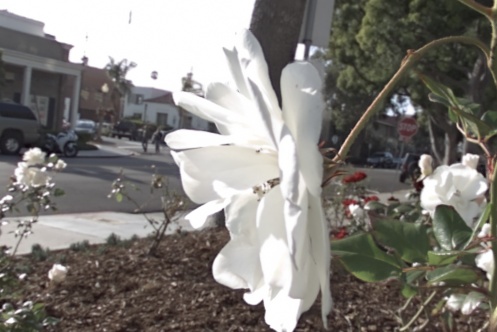}}~
{\includegraphics[scale=0.2,clip]{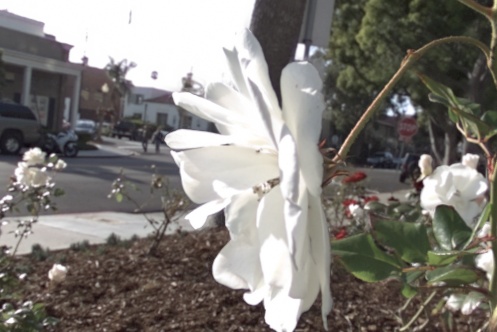}}\\
{\includegraphics[scale=0.2,clip]{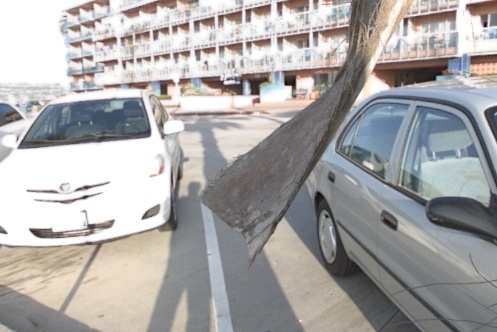}}~
{\includegraphics[scale=0.2,clip]{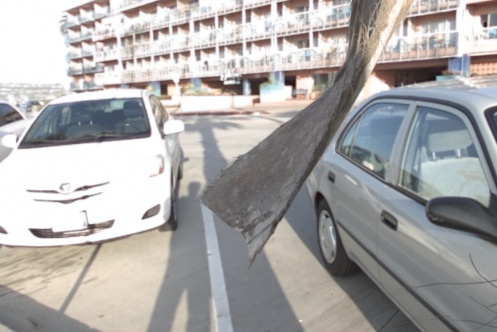}}~
{\includegraphics[scale=0.2,clip]{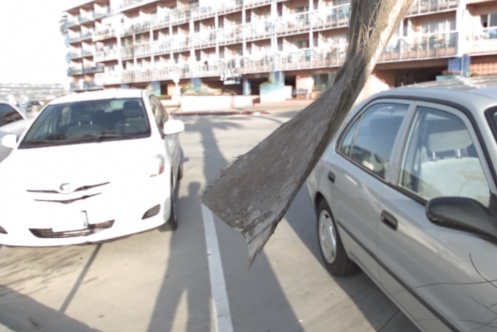}}~
{\includegraphics[scale=0.2,clip]{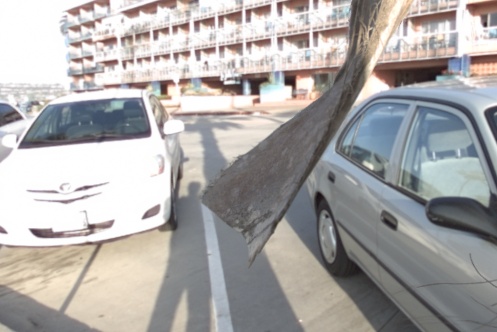}}\\
{\includegraphics[scale=0.2,clip]{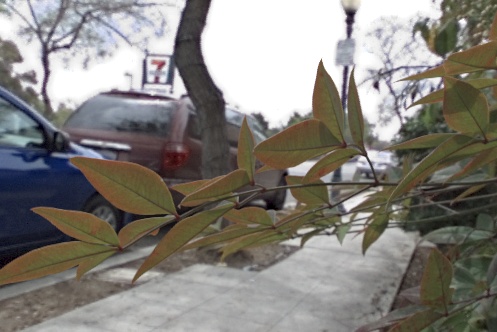}}~
{\includegraphics[scale=0.2,clip]{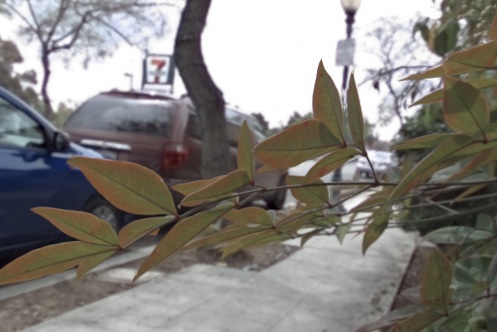}}~
{\includegraphics[scale=0.2,clip]{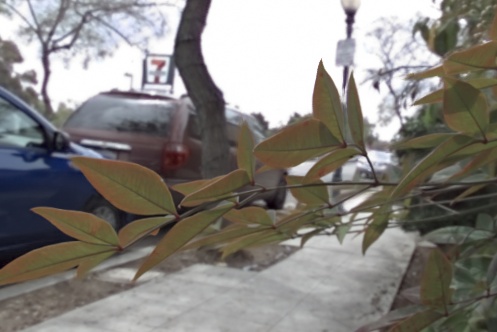}}~
{\includegraphics[scale=0.2,clip]{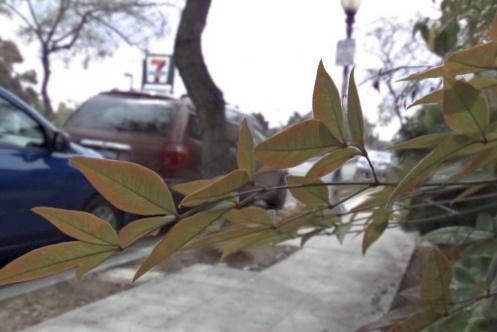}}\\
{\includegraphics[scale=0.2,clip]{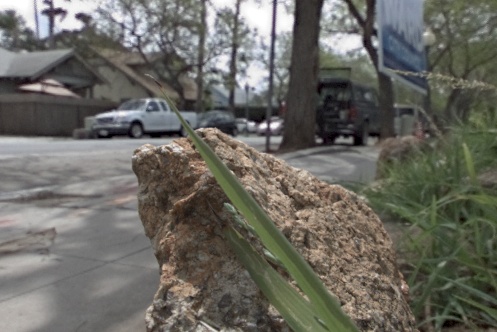}}~
{\includegraphics[scale=0.2,clip]{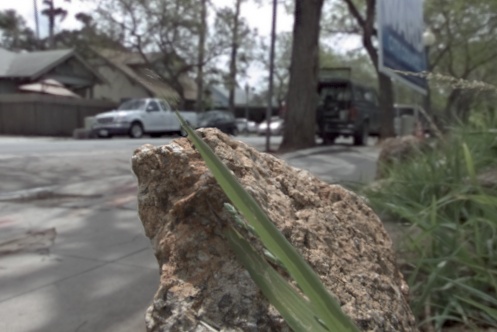}}~
{\includegraphics[scale=0.2,clip]{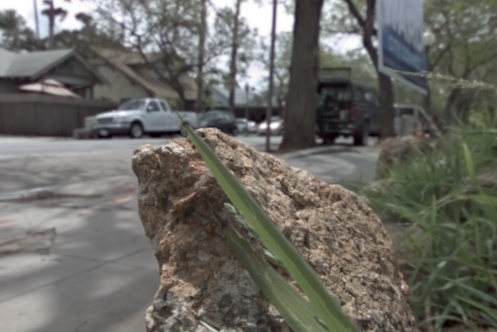}}~
{\includegraphics[scale=0.2,clip]{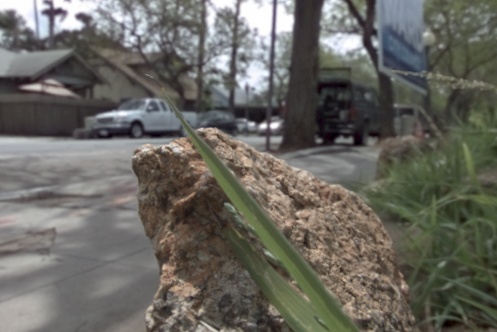}}\\
\caption{ We show view estimates using our method with 64 depth levels w/ mr (trained on Spaces and KITTI ) and Soft3D for qualitative analysis on light field dataset. }
\label{fig:light_field}
\end{figure*}

We show novel view estimates on all $5$ scenes from light field dataset of Kalantari \etal in Fig.~\ref{fig:light_field}. 

\end{document}